%% file: iclr2027_conference.tex
\PassOptionsToPackage{dvipsnames,table}{xcolor}
\documentclass{article} 
\usepackage{iclr2027_conference,times}
\iclrfinalcopy
\setcitestyle{citesep={,},yysep={,}}

\input{math_commands.tex}

\usepackage{hyperref}
\usepackage{caption}
\usepackage{url}
\usepackage{listings}
\usepackage{booktabs}
\usepackage{tablefootnote}
\makeatletter
\AfterEndEnvironment{table*}{\tfn@tablefootnoteprintout\gdef\tfn@fnt{0}}
\makeatother
\usepackage{multirow}
\usepackage{graphicx}
\usepackage{wrapfig}
\usepackage{caption}
\usepackage{xcolor}
\usepackage{algorithm}
\usepackage{makecell}
\usepackage{tcolorbox}
\usepackage{enumitem}

\definecolor{litSkillInk}{HTML}{273449}
\definecolor{litSkillBlue}{HTML}{244F78}
\definecolor{litSkillTeal}{HTML}{16746D}
\definecolor{litSkillPaper}{HTML}{F5F8FB}
\lstdefinestyle{litSkill}{
  basicstyle=\ttfamily\scriptsize\color{litSkillInk},
  columns=fullflexible,keepspaces=true,
  breaklines=true,breakatwhitespace=true,
  showstringspaces=false,frame=none,
  aboveskip=0pt,belowskip=0pt,
  morekeywords={name,description},
  keywordstyle=\color{litSkillTeal}\bfseries,
  morecomment=[l][\color{litSkillBlue}\bfseries]{\#},
  moredelim=[s][\color{litSkillBlue}\bfseries]{**}{**},
  moredelim=[s][\color{litSkillTeal}]{`}{`}
}
\tcbset{litSkillBox/.style={
  enhanced,breakable=false,
  colback=litSkillPaper,colframe=litSkillBlue!30,
  boxrule=0.4pt,arc=2pt,
  borderline west={1.5pt}{0pt}{litSkillBlue},
  left=7pt,right=7pt,top=5pt,bottom=5pt,
  before skip=8pt,after skip=8pt
}}

\tcbuselibrary{skins,breakable}

\title{Learning to Interact and Direct: Co-Evolving Video Director Agents}

\author{Yingzhao Jian, Zhongan Wang, Yuxuan Hou \& HeheFan* \\
College of Artificial Intelligence, Zhejiang University. *Corresponding author. \\
\texttt{hehefan@zju.edu.cn}
}

\begin{document}

\maketitle
\input{sections/0_abstract}

\input{sections/1_introduction}

\input{sections/2_related_work}

\input{sections/3_method}

\input{sections/4_experiments}

\input{sections/5_conclusion}




\bibliography{iclr2027_conference}
\bibliographystyle{iclr2027_conference}

\appendix

\newpage

\begin{center}
\fontsize{16}{20}\selectfont
\textbf{Learning to Interact and Direct: Co-Evolving Video Director Agents}

Appendix
\end{center}

\tableofcontents
\newpage

\input{sections/A1_supplementary_related_work}

\input{sections/A2_formal_representation}

\input{sections/A3_pipeline_implementation}

\input{sections/A4_dataset_construction}

\input{sections/A5_evaluation_protocol}

\input{sections/A6_comparison_methods}

\input{sections/A7_supplementary_experiments}

\end{document}

%% file: math_commands.tex
\usepackage{amsmath,amsfonts,bm}

\def\eqref#1{equation~\ref{#1}}

\def\1{\bm{1}}

\DeclareMathAlphabet{\mathsfit}{\encodingdefault}{\sfdefault}{m}{sl}
\SetMathAlphabet{\mathsfit}{bold}{\encodingdefault}{\sfdefault}{bx}{n}



%% file: sections/0_abstract.tex
\begin{abstract}
Video director agents orchestrate multistage production to realize users' creative visions. 
Emerging research on their self-evolution advances screenplay planning effectively.
However, the mutual reinforcement between directing and user interaction remains largely overlooked.
Improved interaction clarifies user intent and informs workflow refinement, while more reliable execution shifts collaboration from troubleshooting to creative decisions.
To this end, we introduce LIT (\textbf{L}earning to \textbf{I}nteract and Direc\textbf{T}), a framework comprises a director agent, a simulated user, and a meta engineer.
Simulated user assigns creative tasks to probe the director's weaknesses, while the meta engineer uses the task traces to revise the director’s interaction and production strategy. Although this approach can address discovered interaction and production issues, it still relies on passive attribution by the meta agent. This limits its ability to adapt to broader user interaction patterns and production techniques. To solve this problem, we equip LIT with a co-evolution paradigm. The meta agent proactively learns external knowledge, thereby improving its ability to discover problems and propose solutions. The meta agent, simulated user, and director together form an evolutionary closed loop, collaboratively enhances the system's directing and interaction capabilities. Across the Test and OOD Test splits, LIT improves production and Interaction scores by 19.1\% and 32.0\%, respectively, relative to the initial pipeline. Controlled ablations reveal the correlation between improvements in production and interaction, where the co-evolution mechanism further encourage directing skill and interaction mode exploration. 

\begin{center}
\textbf{Project Page:} \href{https://huggingface.co/LIT-review/LIT}{https://huggingface.co/LIT-review/LIT}
\end{center}




\end{abstract}

%% file: sections/1_introduction.tex
\section{Introduction}
\label{sec:introduction}

\begin{figure}[t]
    \centering
    \includegraphics[width=\linewidth]{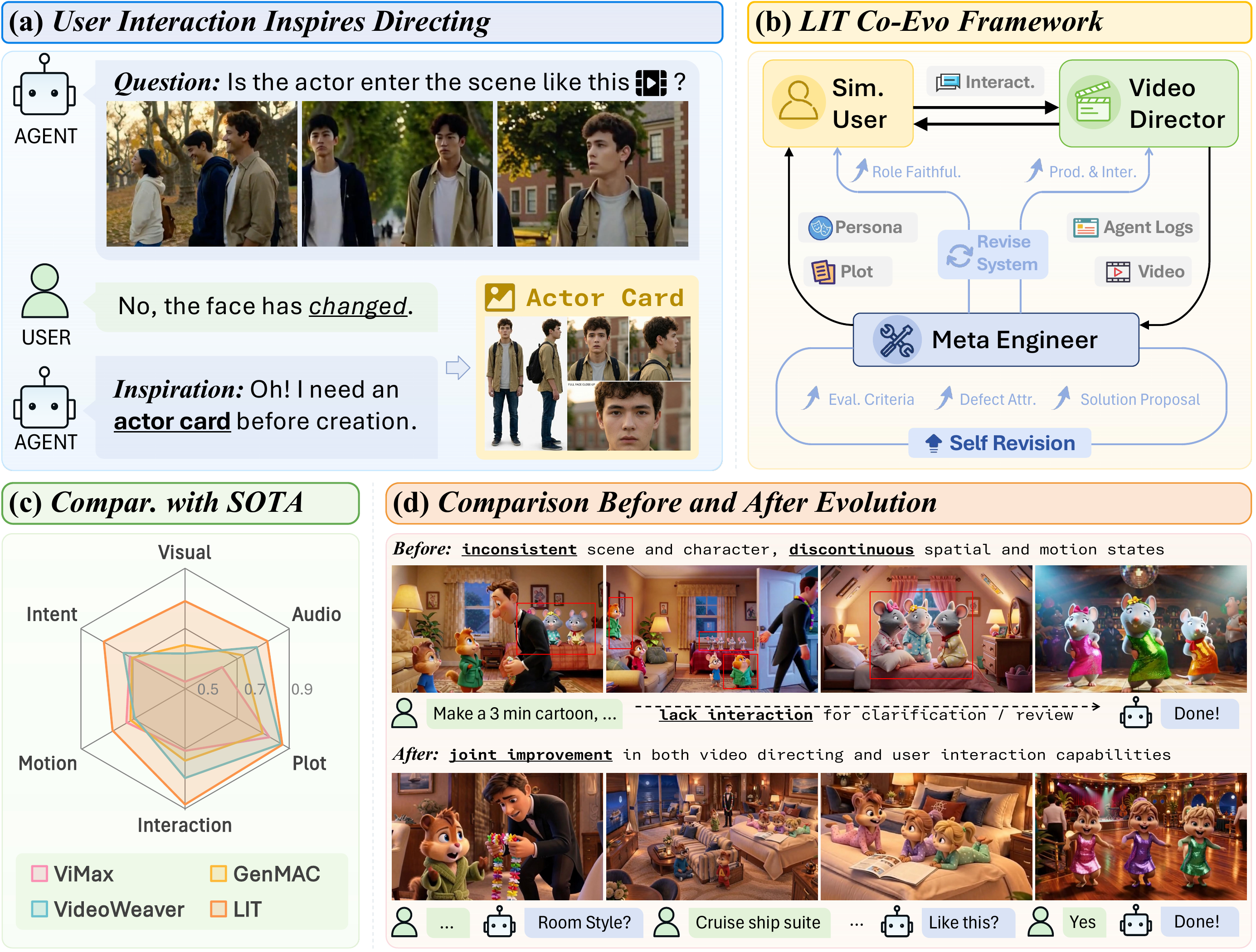}
    \caption{\textbf{Overview of LIT.} (a) User interaction provides inspiration
    for improving the director's workflow. Feedback about a change in a
    character's face leads the director to create an actor card before generation.
    (b) The meta engineer and the agent system co-evolve through simulated
    user interactions. The meta engineer revises the director and simulated
    user. It also updates its own evaluation criteria and revision strategy.
    (c) The radar chart compares LIT with existing methods on
    visual and audio consistency, motion, plot, interaction, and intent
    fulfillment. (d) After evolution, the director asks for clarification
    and feedback on drafts. The example also shows more consistent
    characters and scenes.}
    \label{fig:teaser}
    \vspace{-6mm}
\end{figure}

AI video directors broaden access to cinematic storytelling, enabling people to bring imagined worlds and narratives to the screen. They translate creative intent into coordinated decisions across scriptwriting, shot design, generation, and editing~\citep{filmagent2025}. To improve this orchestration beyond individual projects, recent research has begun to explore production-centered self-evolution. These approaches refine reusable skills, workflow configurations, and production policies from execution feedback, allowing lessons from earlier videos to improve subsequent planning and generation~\citep{videoweaver2026,cineforge2026}.

The user determines what a successful video should express, yet an initial request rarely captures every choice that shapes that expression. Directing therefore requires interaction: clarifying creative priorities, recognizing delegated decisions, and negotiating revisions as drafts emerge. User-guided creation supports this process through dialogue and feedback, which can also inform workflow refinement~\citep{aesopagent2024,animagents2025}. However, these approaches primarily use interaction to guide production. The interaction strategy itself is seldom treated as an evolving counterpart to the production workflow. This leaves their reciprocal improvement across projects largely unexplored. Poorly timed feedback can force revisions across completed assets and shots, while unreliable execution keeps users occupied with troubleshooting instead of creative decisions.

Our insight is that interaction and production are mutually reinforcing targets of evolution. Interaction reveals intent and workflow limitations that execution feedback alone may miss. Reliable production enables users to provide more informed creative feedback. In Fig.~\ref{fig:teaser}(a), the user points out that a character's face has changed. This feedback motivates a reusable workflow change: create an actor card before generation to maintain character consistency. Fig.~\ref{fig:teaser}(d) illustrates the joint improvement in production and interaction after evolution. The director clarifies scene preferences and requests feedback on drafts, while producing more consistent characters and scenes. Together, these examples illustrate our goal of improving both the production workflow and the interaction strategy across projects.

To this end, we introduce LIT (\textbf{L}earning to \textbf{I}nteract and Direc\textbf{T}), a self-improving video creation framework comprising a director agent, a simulated user, and a meta engineer.
When an evolution task begins, the simulated user first assigns a creative task to the director. This task is generated according to a user persona, and is designed to expose defects in the director's interaction and production strategies.
The director then receives the request and elicits further intent through dialogue and draft reviews. During this process, the simulated user may clarify preferences, delegate routine choices, or request revisions.
Finally, the meta engineer analyzes the resulting dialogue and production traces, thereby revising the director's interaction strategy and directing workflow. These revisions persist across projects, allowing each interaction to inform how future videos are made.

Although this process can solve problems exposed through outcome evaluation and defect attribution, it still depends largely on the meta engineer's ability to discover problems and explore solutions. Existing methods, however, rely on a fixed meta-engineer design, whose capacity for defect localization remains static and thus limits problem coverage. To address this, we equip LIT with a curiosity-driven co-evolution mechanism. During evolution, the meta engineer proactively learns external knowledge, improving its ability to discover problems and propose solutions. The meta engineer, the simulated user, and the director together form an evolutionary closed loop that collaboratively enhances the system's directing and interaction capabilities.

We evaluate LIT on held-out video creation tasks and user profiles. Across the Test and OOD Test splits, LIT improves production and Interaction scores by 19.1\% and 32.0\%, respectively, relative to the initial pipeline. These gains use a 2:1 weighting of the two splits to match their sample counts. In a user study, participants prefer LIT for intent fulfillment in 51.7\% of group assessments (31 of 60), compared with ViMax and VideoWeaver. Controlled ablations show that production and interaction learning reinforce each other, while Co-evolution further enhances this process.

In summary, our contributions are threefold:
\begin{itemize}[leftmargin=1.5em, labelsep=0.3em, itemsep=2pt]
    \item We identify interaction and production are mutually reinforcing during director evolution.
    \item We propose a co-evolution mechanism that evolves the meta engineer itself through proactive external knowledge learning, enhancing its ability to improve both production and interaction.
    \item LIT improves production and Interaction scores by 19.1\% and 32.0\%, respectively, relative to the initial pipeline across the Test and OOD Test splits.
\end{itemize}

%% file: sections/2_related_work.tex



%% file: sections/3_method.tex
\begin{figure}[t]
    \centering
    \includegraphics[width=\linewidth]{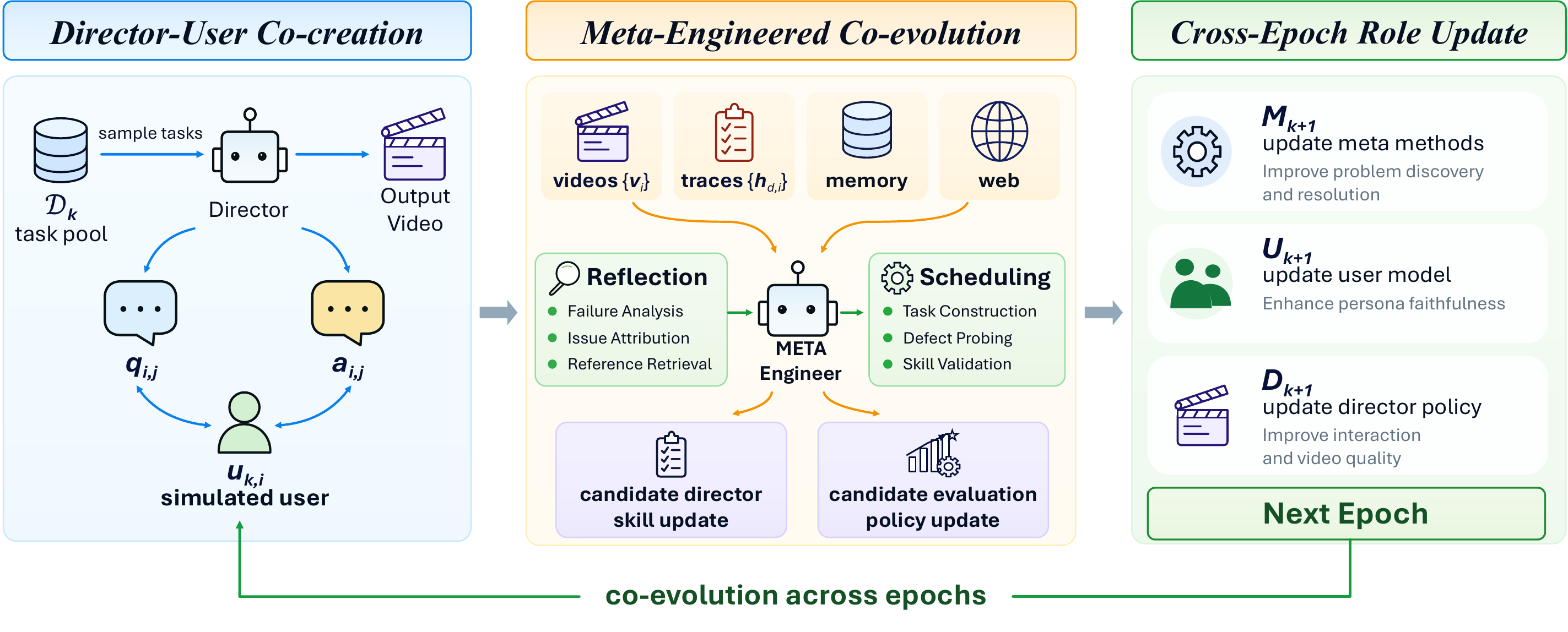}
    \vspace{-6mm}
    \caption{LIT learns from user interaction and video production
    across epochs. Left: The director asks the simulated user to clarify
    intent and uses the answers to guide production. Center: The meta engineer
    analyzes videos and production traces to identify problems. It uses memory
    and external references to propose skill and evaluation updates. It also
    schedules validation and designs tasks to expose weaknesses and test new
    methods. Right: Accepted revisions guide the meta engineer, simulated
    user, and director in the next epoch.}
    \label{fig:method}
    \vspace{-4mm}
\end{figure}

\section{Method}
\label{sec:method}

LIT is a harness framework for co-evolution between a meta engineer and an agent system across video creation projects, aimed at improving both directing
and user interaction.
In each project, a video director collaborates with a simulated user
to construct and execute a production pipeline.
A meta engineer analyzes the resulting dialogue, execution traces,
and artifacts to refine the director's production workflow
and interaction strategies.
It learns from successful and failed revisions to update its own
task selection and revision strategies.
An overview of the co-creation and co-evolution process is shown in
Fig.~\ref{fig:method}.

To support evolution and assess its outcomes, we construct a dataset
pairing scripts with user personas.
Each pair provides the simulated user with the creative context
and behavioral guidance for its interactions with the director.

\subsection{Director-User Interaction Dataset}

We construct a director--user interaction dataset by pairing video directing tasks wit user persona. Specifically, we extract directing task from video and plot dataset, and leverage LLM for persona synthesis. Each persona specifies the user's
production knowledge, communication style, and disclosure behavior, shaping
both the initial request and subsequent responses. During persona-consistent
verification, clarification and adversarial probes test these responses,
and reviewer feedback guides revisions to the persona and disclosure policy
(Fig.~\ref{fig:data}).

\noindent\begin{minipage}[t]{0.50\linewidth}
\vspace{0pt}
\paragraph{Example case.}
Consider the following illustrative exchange about the lighting of an
interior scene.

\noindent\textbf{Director:} ``What kind of lighting would you like in the room?''
\begin{itemize}[leftmargin=1.5em, labelsep=0.3em, itemsep=2pt]
\setlength{\itemsep}{2pt}
\item \textbf{User A (professional persona):}
``Use low-key lighting, with a soft side key and a subtle cool rim light.''
\item \textbf{User B (non-professional persona):}
``I'm not sure about lighting. Could you show me a few examples so I can choose?''
\end{itemize}
The first response specifies an actionable lighting setup. the second
requests visual alternatives, requiring the director to help the user
articulate a preference.
\par\medskip

The dataset contains 240 tasks, 284 reference-material entries, and 892
planned shots. Appendix~\ref{app:dataset} provides further details on
construction, task distributions, and persona settings.
\end{minipage}\hfill
\begin{minipage}[t]{0.43\linewidth}
\vspace{0pt}
\centering
\includegraphics[width=\linewidth]{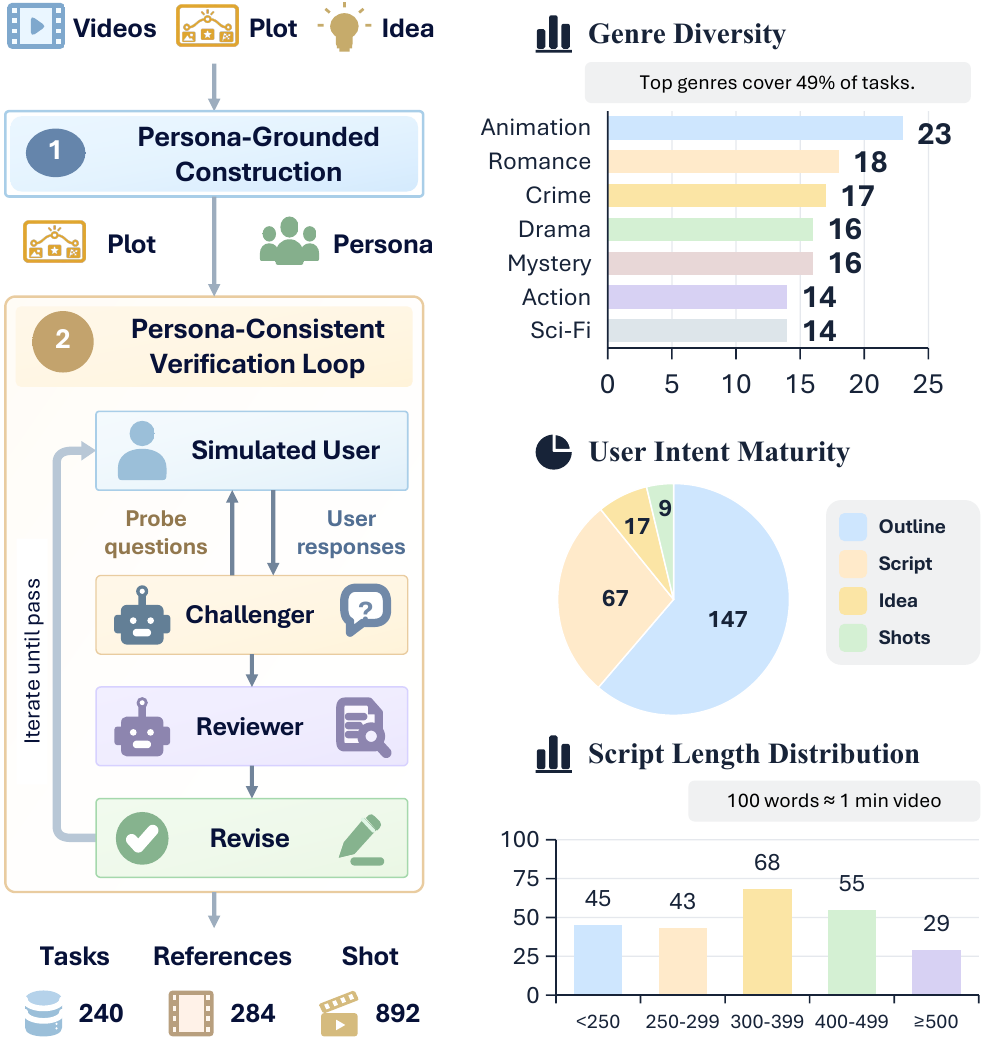}
\captionof{figure}{Dataset construction and task distributions, including
persona-consistent verification and structured annotation.}
\label{fig:data}
\end{minipage}
\par\medskip

\subsection{Simulated Director--User Interaction}
\label{sec:interaction}

LIT uses a skill-driven harness in which evolvable skills govern
most of the control flow, agent coordination, and evolution procedures.
These structures can be revised together with the agents'
production, interaction, and evolution strategies.

At evolution epoch $k$, we denote the editable resource bundles of the
director, simulated user, and meta engineer by $D_k$, $U_k$,
and $M_k$, respectively.
Each bundle contains governing skills, supporting tools, and assets.
A shared runtime $\mathcal{F}$ (e.g., Codex, Claude Code, or OpenClaw)
instantiates the meta engineer $m_k$, simulated user $u_{k,i}$, and director $d_k$ as:
\begin{equation}
m_k = \mathcal{F}[M_k], \quad
\{(s_i,p_i)\}_{i=1}^{B}
\gets m_k(\mathcal{D}_{\mathrm{evo}},\mathcal{H}_k),  \quad
u_{k,i} = \mathcal{F}[U_k; s_i,p_i], \quad
d_k = \mathcal{F}[D_k],
\end{equation}
where the meta engineer writes a script $s_i$ and persona $p_i$ for a simulated user, drawing on the evolution dataset $\mathcal{D}_{\mathrm{evo}}$. $\mathcal{H}_k$ is the evolution history, $i$ is the project index, and $B$ is the number of projects in a batch.
For clarity, we suppress the epoch index on project-specific
quantities and omit the agents' execution contexts.

For each project, the user first provides an initial production prompt $t_i$ to the director.
The director then develops and executes a production plan
specifying the workflow and the skills required at each stage.
During production, it may query the user about narrative or
design details, receiving a response $a_{i,j} = u_i(q_{i,j})$
to its $j$-th question.
These responses inform revisions to the plan or intermediate artifacts.
The complete interaction is expressed as:
\begin{equation}
(v_i,h_{d,i},h_{u,i}) \gets d_k(u_i), \qquad
h_{u,i} = \bigl(t_i,(q_{i,j},a_{i,j})_{j=1}^{J_i}\bigr),
\end{equation}
where $h_{d,i}$ and $h_{u,i}$ denote the production trajectory
and interaction history, respectively, and $J_i$ is the number
of question--answer exchanges.

After each batch, the meta engineer analyzes the generated videos,
production trajectories, and interaction histories to revise
the role bundles, thereby closing the evolution loop:
\begin{equation}
(D_{k+1},U_{k+1},M_{k+1})
\gets m_k\left(
\{(v_i,h_{d,i},h_{u,i})\}_{i=1}^{B},
\mathcal{H}_k
\right).
\end{equation}

\subsection{Meta Engineer Co-evolution}
\label{sec:meta_evolution}

The evolution procedure follows a hypothetico-deductive paradigm, which largely depends on problem discovery and solution exploration. However, existing methods rely on fixed self-improvement harness designs. The meta engineer's criteria and capacity for defect localization remain static, limiting the problem coverage. In addition, revision relies on reactive defect attribution, which constrains solution diversity.

Motivated by these observations, we introduce a curiosity-driven co-evolution harness for director–user interaction. The meta engineer proactively evolves its evaluation criteria and development capabilities through web data mining and reflection. The user assigns creation projects that expose defects, while encouraging the director to explore novel skills. In this process, a curriculum-learning schedule balances exploration and optimization, collaboratively raising the ceiling of evolution.

\textbf{Proactive reflection and exploration.}
After each revision, the meta engineer first reflects on historical evolution trajectories. It then searches the web for AI videos, audience comments, and technical analyses. By comparing these materials with previous iterations, it identifies under-explored topics and evaluation criteria, and uncovers potential solutions and alternatives. The meta engineer also proactively verifies individual production steps or its own evaluation criteria, thereby discovering improvements beyond reactive attribution.

\textbf{User-driven evolution scheduling.}
We incorporate curriculum learning along the evolution process. Based on pipeline maturity and historical evaluation results, the meta engineer designs interaction tasks with different difficulty levels and categories for the current evolution epoch. The user carries out the interaction tasks, which elicit defect exposure through targeted plot design, and encourage the trial of novel directing skills. Finally, the meta engineer proposes revision plans accordingly, drawing on its evolving solution library as a reference.

%% file: sections/4_experiments.tex
\begin{table*}[t]
\centering
\caption{Comparison on Test Split and OOD Test across six dimensions. The best and second-best results are denoted by \textbf{bold} and \underline{underline}, respectively. Ground Truth is excluded from ranking.}
\label{tab:main-comparison}
\label{tab:comparison-new-baselines}
\resizebox{\textwidth}{!}{%
\begin{tabular}{l *{12}{c}}
\toprule
\multirow{2}{*}{\textbf{Method}} 
& \multicolumn{6}{c}{\textbf{Test Split Scores} $\uparrow$} 
& \multicolumn{6}{c}{\textbf{OOD Test Scores} $\uparrow$} \\
\cmidrule(lr){2-7} \cmidrule(lr){8-13}
& Visual & Audio & Motion & Plot & Inter. & Intent
& Visual & Audio & Motion & Plot & Inter. & Intent \\
\midrule
\textit{Ground Truth}          
& \textit{0.8828} & \textit{0.8906} & \textit{0.9219} & \textit{1.0000} & -- & -- & \textit{0.8203} & \textit{0.8906} & \textit{0.9453} & \textit{1.0000} & -- & -- \\
\midrule
GPT-6-Astra\tablefootnote{\fontsize{8}{9}\selectfont\label{fn:gpt6astra}\url{https://openai.com/index/gpt-6-astra/}.}
& \underline{0.7995} & \underline{0.8151} & \textbf{0.8398} & 0.8730 & \underline{0.8625} & 0.7891 & \textbf{0.7500} & 0.7109 & \underline{0.6328} & 0.7695 & 0.6813 & 0.6188 \\
MovieAgent~\citep{movieagent2025}  
& 0.4557 & 0.6654 & 0.7695 & 0.8223 & -- & -- & 0.5990 & 0.5651 & 0.5469 & 0.6680 & -- & -- \\
VideoClaw\tablefootnote{\fontsize{8}{9}\selectfont\label{fn:videoclaw}VideoClaw repository: \url{https://github.com/HITsz-TMG/VideoClaw}.}
& 0.5156 & 0.7565 & 0.7617 & 0.8965 & 0.8398 & \underline{0.8227} & 0.6589 & 0.7005 & 0.6094 & \textbf{0.8086} & 0.7094 & 0.6375 \\
ViMax~\citep{huang2026vimax}
& 0.4909 & 0.6901 & 0.8027 & 0.8672 & 0.7695 & 0.7797 & 0.5859 & 0.5469 & 0.5703 & 0.7383 & 0.5828 & 0.5766 \\
\midrule
Before Evo. (Ours)
& 0.5091 & 0.7122 & 0.7891 & 0.8809 & 0.7367 & 0.7352 & 0.6901 & 0.5182 & 0.4961 & 0.7188 & 0.5406 & 0.5609 \\
GEPA~\citep{agrawal2026gepa} & 0.5443 & 0.7461 & 0.7793 & 0.8789 & 0.7609 & 0.7273 & 0.6979 & 0.6693 & 0.5078 & 0.7109 & 0.6906 & 0.5906 \\
GenMAC~\citep{genmac2026}        
& 0.6107 & 0.7409 & 0.8027 & 0.8496 & 0.7758 & 0.7484 & 0.7135 & 0.6823 & 0.5234 & 0.6953 & 0.6687 & 0.6141 \\
VideoWeaver~\citep{videoweaver2026} 
& 0.6523 & 0.7839 & 0.7734 & \underline{0.9082} & 0.8156 & 0.7688 & 0.7057 & \underline{0.7578} & 0.5352 & 0.7891 & \underline{0.7625} & \underline{0.6703} \\
\midrule
LIT (Ours)
            & \textbf{0.8125} & \textbf{0.8385} & \underline{0.8359} & \textbf{0.9121} & \textbf{0.9086} & \textbf{0.8359} & \underline{0.7474} & \textbf{0.7708} & \textbf{0.6641} & \underline{0.7969} & \textbf{0.8406} & \textbf{0.7641} \\
\bottomrule
\end{tabular}%
}
\end{table*}

\section{Experiments}
\label{sec:experiments}
\subsection{Experimental Setup}

We evaluate LIT's evolution trajectory and video directing performance on the constructed dataset. For the trajectory, we reports both the evaluation results and process metrics after each epoch, thereby analyzing the correlation between user interaction and production capabilities. 

Regarding directing performance, we compare with existing state-of-the-art video directing agents, including fixed and self-evolving variants. We further test the impact of each component, and verify the generalization across different base models. Finally, we conduct human evaluations and examine the alignment between machine and human judgments.

At test time, each compared method interacts with the privileged User
described in Appendix~\ref{app:evaluation}.
This User has access to the complete task specification and persona.
The Director receives only the public request and information disclosed
through the conversation according to the User's disclosure rules.

\subsubsection{Metrics}
\label{sec:metrics}

We evaluate generated videos along six aspects related to directing and user interaction. Specifically, we evaluate \textit{visual}, \textit{audio}, \textit{motion}, \textit{plot}, \textit{interaction}, and \textit{intent}. These aspects fall under four higher-level dimensions: \textbf{Consistency} (\textit{visual, audio}), \textbf{Dynamic} (\textit{motion, plot}), \textbf{Interaction}, and \textbf{Intent}. Each aspect is graded by an LLM against a predefined checklist. Details can be found in Appendix~\ref{app:evaluation}.

For user preference analysis, we compute Kendall's \(\tau\) between user and evaluator rankings. Overall, Consistency is assessed through Visual and Audio, Dynamic through Motion and Plot, while Interaction and Intent are each assessed directly.

\subsubsection{Implementation}
\label{sec:implementation}

We implement LIT based on Codex skills. We use Qwen3.8-27B model with W8A8 quantization. The reasoning effort is set to low, with a context length of 262K, and auto-compact at 220K. Video generation uses Minimax-H3 through ComfyUI, with 4-step turbo inference LoRA. Music is generated by MiniMax-Music3. Sound effects are generated by Woosh. Vocal separation uses Mel-RoFormer. All the above models are locally deployed on Ascend 910C NPU servers.

Image generation, ASR, audio understanding, TTS, and embedding and reranking are provided by commercial cloud APIs, specifically Nano-Banana-2-Lite, Qwen-Audio-3.0-ASR-Flash, Qwen3.5-Omni-Plus, Qwen3-TTS-Instruct-Flash, Qwen3-Embedding, and Qwen3-Rerank.

For evaluation, we uses 4 agent voting with Qwen3.8-27B xhigh reasoning effort. The multimodal perception suite for evaluation is consistent with the above. 
For the ground-truth samples in OOD-test, we use videos manually crafted with the same toolchain. Each handcrafted video is reviewed to ensure quality.

\subsection{Comparison between Video Director Agents}
\paragraph{Setting.}
We compare LIT with state-of-the-art video directing and self-evolution methods.
Within them, ViMax and VideoClaw\textsuperscript{\ref{fn:videoclaw}} are the latest directing pipeline. GenMAC and VideoWeaver performs in-project and cross-project evolution, respectively. We additionally included general-purpose methods, including GPT-6-Astra\textsuperscript{\ref{fn:gpt6astra}} and GEPA.

The evolution experiments use four epochs on the same task collection and infrastructure.
Development, validation, and production budgets differ across methods.
The comparison therefore measures performance under the method-specific
budgets described in Appendix~\ref{app:baselines}.
We report automatic and human evaluation results. Finally, we provide visualization of generated videos for qualitative assessment.

\textbf{Quantitative Results.}
As shown in Tab.~\ref{tab:main-comparison}, LIT achieves average scores of 85.7 and 76.4, respectively. On the interaction dimensions, LIT exceeds existing methods by 9.2\% (Interaction) and 6.7\% (Intent) on average, with the Test Split weighted 2 and the OOD Test weighted 1. These results demonstrate the effectiveness of a self-improving user interaction strategy.

For the gains summarized in the abstract and introduction, we use Before Evo. in Tab.~\ref{tab:comparison-new-baselines} as the baseline.
Production is the mean of Visual, Audio, Motion, and Plot scores.
We aggregate each score across the 32 Test and 16 OOD Test cases with weights of 2:1, then compute its relative increase over the baseline.
Production rises from 68.38 to 81.48, a 19.1\% relative increase.
Interaction rises from 67.13 to 88.59, a 32.0\% relative increase.

As shown in Tab.~\ref{tab:comparison-new-baselines}, LIT improves the pipeline by 17.9\% and 30.1\% on the Test and OOD-test Split, respectively. VideoWeaver, GenMAC, and GEPA improve by 7.8/19.7\%, 3.8/10.6\%, and 1.7/9.7\%, respectively. GEPA validate every modification by a full rollout of the pipeline, which leaves little budget for evolution in such a costly video generation setting. GenMAC evolves within a single project and lacks cross-project experience, whereas VideoWeaver keeps the meta agent fixed during evolution. In contrast, LIT explicitly models the user and co-evolves the interaction policy with the production workflow, enabling the learned capabilities to generalize across projects.

\begin{figure}[t]
    \centering
    \includegraphics[width=\linewidth]{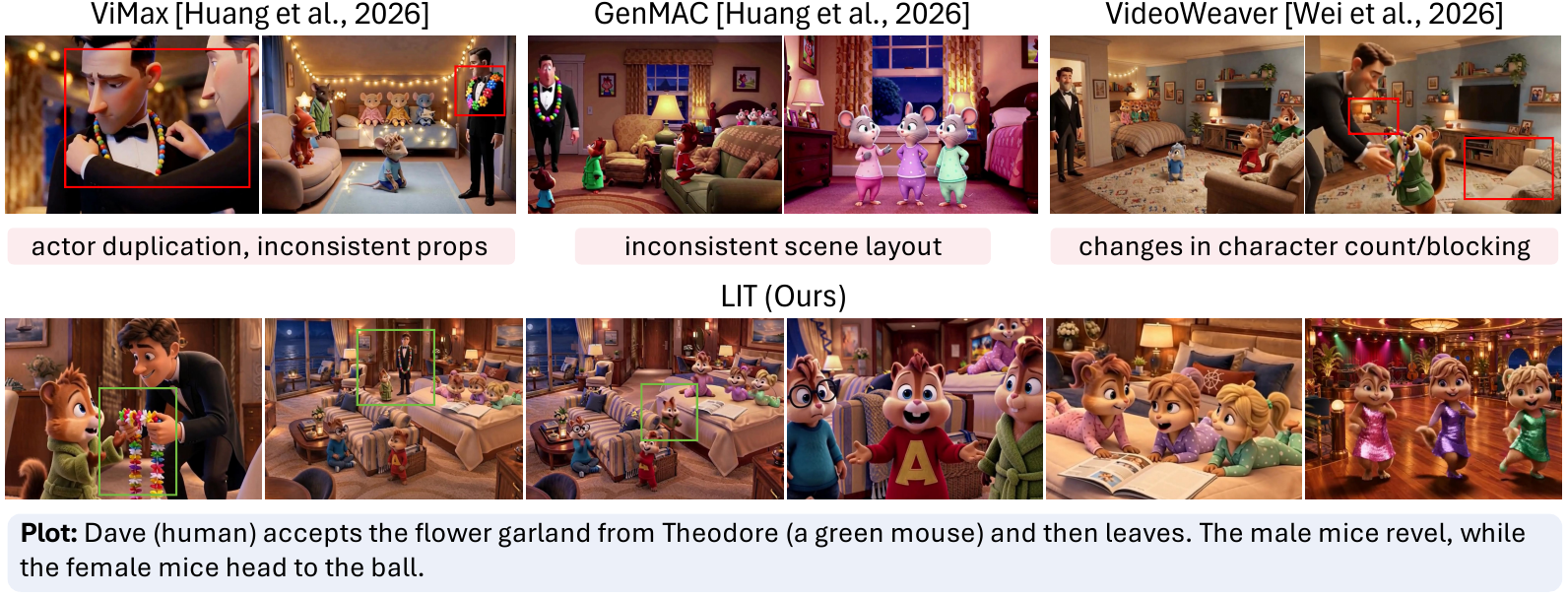}
    \caption{Qualitative comparison on the same plot. Baselines show
    duplicated characters, inconsistent props and layouts, or changes in
    character count and positioning. LIT shows better visual continuity
    in this example. Red boxes mark errors, and green boxes highlight
    consistent details.}
    \label{fig:qualitative-comparison}
\end{figure}

\begin{figure}[t]
    \centering
    \includegraphics[width=\linewidth]{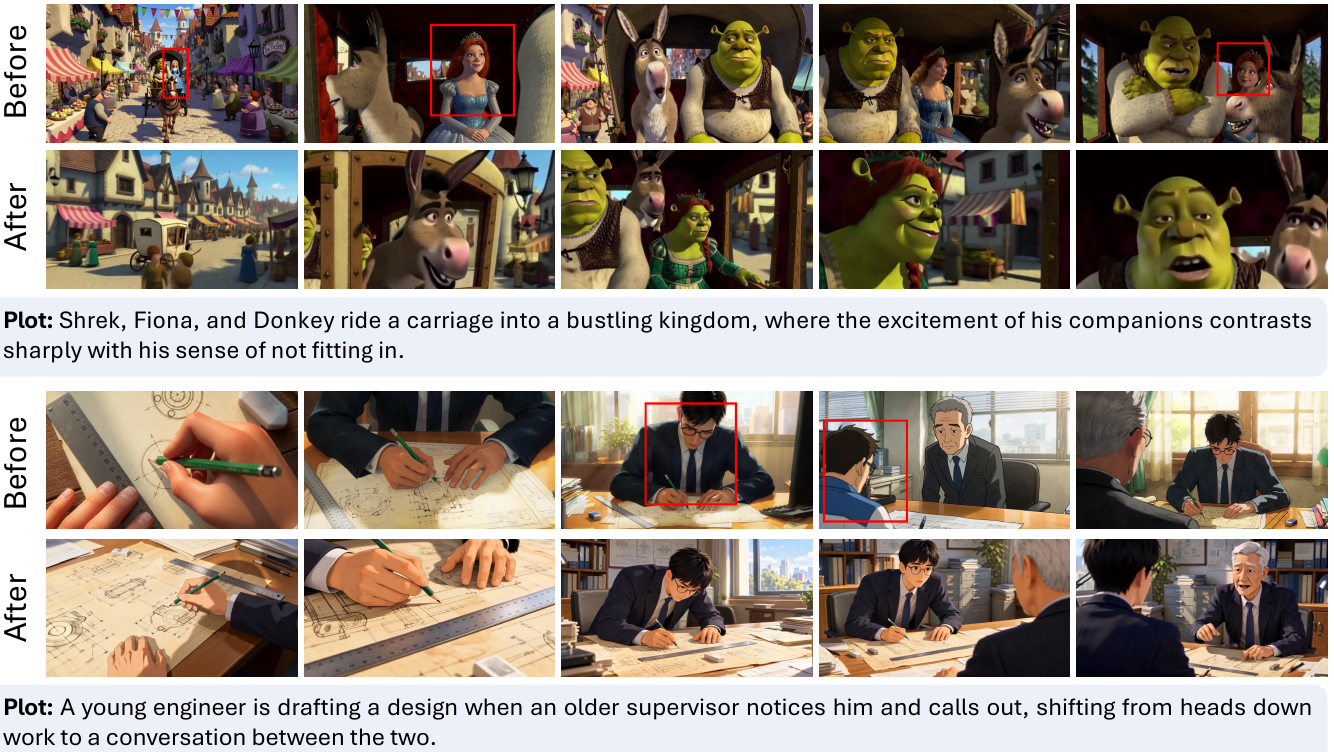}
    \caption{Visual continuity before and after evolution. The examples
    show more consistent character appearance in the carriage scene (top)
    and character positioning in the office scene (bottom) after evolution.
    Red boxes mark inconsistencies before evolution.}
    \label{fig:before-after}
    \vspace{-4mm}
\end{figure}

\paragraph{Qualitative Results.} ViMax exhibits actor duplication and inconsistent props, which are common problems for director agents without targeted revision. GenMAC performs self-improvement on a single project, but within a single project, it is difficult to discover and resolve scene consistency issues. VideoWeaver achieves scene consistency through cross-project attribution, but due to evaluation limitations, it does not learn character blocking consistency. LIT's co-evolution mechanism effectively improves the meta agent's problem discovery and resolution capabilities, achieving high-quality video direction (Figs.~\ref{fig:qualitative-comparison} and~\ref{fig:before-after}).

\paragraph{User Study.} We conduct a user study with \(15\) participants. Each method generates 16 videos. For each participant, we randomly sample 4 pairs of videos generated by different methods in a blind and randomized order, and the participant selects the preferred one along the corresponding evaluation dimension. We report the number of wins under user evaluation and machine evaluation, respectively. We further compute Kendall's \(\tau\) between human preferences and automatic metrics to validate the rationality of our evaluation metrics. 

\begin{figure*}[t]
\centering
\begin{minipage}[c]{0.46\textwidth}
    \centering
    \captionof{table}{User preference comparison across four dimensions. 
U denotes user and E denotes machine evaluator.
Human counts aggregate 60 group assessments. Evaluator counts cover 16 groups.
Ev., Co., Dy., It., and Ia. represents Evaluator, Consistency, Dynamic, Intent and Interaction, respectively.
The last row reports Kendall's $\tau$ between User and Evaluator rankings for each dimension. The best and second-best results are denoted by \textbf{bold} and \underline{underline}, respectively.}
    \label{tab:user-preference}
    \setlength{\tabcolsep}{3pt}
    \scriptsize
    \resizebox{\linewidth}{!}{%
    \begin{tabular}{l l cccc}
    \toprule
    \textbf{Method} & \textbf{Ev.} & \textbf{Co.} $\uparrow$ & \textbf{Dy.} $\uparrow$ & \textbf{It.} $\uparrow$ & \textbf{Ia.} $\uparrow$ \\
    \midrule
    \multirow{2}{*}{\shortstack[l]{ViMax\\\citep{huang2026vimax}}}        & U & 9 & 13 & 7 & 5 \\
                                                                          & E & 2 & \underline{5} & 0 & 1 \\
    \midrule
    \multirow{2}{*}{\shortstack[l]{VideoWeaver\\\citep{videoweaver2026}}} & U & \underline{14} & \underline{16} & \underline{22} & \underline{20} \\
                                                                          & E & \underline{5} & 4 & \underline{6} & \underline{4} \\
    \midrule
    \multirow{2}{*}{LIT (Ours)}                                           & U & \textbf{37} & \textbf{31} & \textbf{31} & \textbf{35} \\
                                                                          & E & \textbf{9} & \textbf{7} & \textbf{10} & \textbf{11} \\
    \midrule
    \multicolumn{2}{c}{Kendall's $\tau$} & 1 & 0.33 & 1 & 1 \\
    \bottomrule
    \end{tabular}%
    }
\end{minipage}
\hfill
\begin{minipage}[c]{0.50\textwidth}
    \centering
    \includegraphics[width=\linewidth]{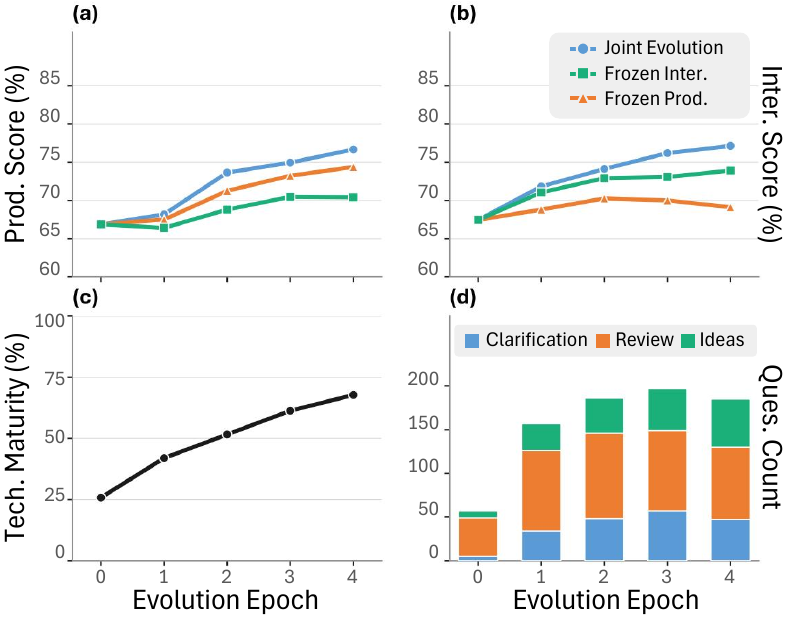}
        \captionof{figure}{Evolution over four epochs. (a, b) Joint evolution
    reaches higher production and interaction scores than either frozen
    variant. (c) Technical maturity measures coverage of directing practices.
    (d) Bars count clarification, review, and idea exchanges.
    Epoch 0 denotes the initial system.}
    \label{fig:evolution-process}
\end{minipage}
\end{figure*}

As shown in Tab.~\ref{tab:user-preference}, LIT achieves the most user preferences. The machine evaluation metrics achieve an average Kendall's \(\tau\) of 0.833 with human preferences, demonstrating a strong correlation.

\subsection{Evolution Process of User-Director Interaction}


\paragraph{Evolution Curve.}
We run the evolution process for four rounds (Fig.~\ref{fig:evolution-process}). Each round we measure the interaction and production capabilities of the director agent. Both capabilities are evaluated on the held-out and OOD test set. To analyze their collaborative improvement effect, we further conduct ablation studies, in which the production workflow and the interaction policy are frozen, respectively. We compare these ablation curves with those under joint evolution.

The results in Fig.~\ref{fig:evolution-process}(a) and (b) show a joint improvement in Interaction and Production capabilities, with their evolution effects mutually reinforcing each other. User Interaction raises the capability upper bound of Production. The lack of Production evolution limits the ability to fulfill User Intent, which leads to degraded performance in Interaction.

\paragraph{Process-Oriented Results.}
Due to the multi-step nature of the directing workflow, the agent's actual improvements at certain steps may not necessarily be reflected in the final evaluation results.
Therefore, we analyzed LIT's process metrics to provide a more comprehensive perspective. Specifically, we evaluated \textit{technical maturity}, \textit{user interaction modes}, and changes in \textit{meta criteria}.

\textit{Technical maturity} is evaluated using a predefined checklist of common directing practices extracted from open-source, handcrafted workflows. \textit{User interaction modes} are divided into three categories: requesting clarification, requesting review, and providing ideas. These categories reflect changes in the agent's mode of collaboration with the user. We also analyze changes in the \textit{meta engineer's evaluation criteria}, which guides the evolution of the director.

As shown in Fig.~\ref{fig:evolution-process}(c), the curve of Technical maturity exhibits consistent growth and is more stable than the trend of production quality. The changes in Evaluation Criteria are as follows:

\begin{tcolorbox}[
  colback=gray!5,
  colframe=gray!60,
  fonttitle=\bfseries,
  fontupper=\scriptsize,
  breakable
]

\textbf{P1:} Plan Fidelity,
\textbf{P2:} Reference Fidelity,
\textbf{P3:} Visual Quality,
\textbf{P4:} Audio Quality,
\textbf{P5:} Plot Coherence,
\textbf{P6:} Phys. Plausibility.

\textbf{I1:} Info Acquisition‌,
\textbf{I2:} Ques. Necessity‌,
\textbf{I3:} Ques. Timing,
\textbf{I4:} Intent Understanding,
\textbf{I5:} Intent Fulfill.,
\textbf{I6:} Ins. Following.

\vspace{1em}

$\rightarrow Epoch\ 1$: 
[\textcolor{RedOrange}{\textbf{Enhance}} \textbf{P1} + shot plan faithfulness]\quad
[\textcolor{RedOrange}{\textbf{Enhance}} \textbf{P4} + unison of snd. and img.]\quad
[\textcolor{ForestGreen}{\textbf{Add}} \textbf{P7}: Shot Coherence]

$\rightarrow  Epoch\ 2$: 
[\textcolor{ProcessBlue}{\textbf{Relegate‌}} \textbf{P2}]\quad
[\textcolor{RedOrange}{\textbf{Enhance}} \textbf{P3, P7} + actor num. \& pos. consistency]\quad
[\textcolor{RedOrange}{\textbf{Enhance}} \textbf{I2} + penalty on ques. num.]

$\rightarrow Epoch\ 3$: 
[\textcolor{ProcessBlue}{\textbf{Relegate‌}} \textbf{I3}]\quad
[\textcolor{RedOrange}{\textbf{Enhance}} \textbf{P5} + speech \& action density]\quad
[\textcolor{RedOrange}{\textbf{Enhance}} \textbf{P7} + motion coherence]

\end{tcolorbox}

The results indicate that the model's evolutionary focus shifts from fixing obvious defects to optimizing details. Moreover, based on the number and distribution of issues in Fig.~\ref{fig:evolution-process}(d), the model becomes more proactive in user interaction, evolving from initially requesting information to providing production ideas.

\subsection{Ablation Study}
We compare the performance of LIT on various VLM backbones, and conduct ablations on different components of our framework. Due to cost constraints, each epoch generates eight videos. On top of the evolved framework, we additionally tried 20-step inference to examine the impact of the video foundation model.

Based on the results in Tab.~\ref{tab:ablation}, LIT is effective across multiple VLM backbones. Models with stronger multimodal capabilities have stronger directing capabilities, but they also introduce more retries, leading to higher costs. 20-step inference improved Dynamic performance, indicating that video dynamics are strongly correlated with the foundation model.

Ablation studies show that the improvement of Interaction and Production capabilities has a synergistic effect, validating the effectiveness of LIT’s explicit modeling of users.
Co-Evo effectively improves the model’s performance by exploring defects and solution proactively, while keeping costs controllable.

\begin{table*}[t]
\centering
\caption{Ablation study on Test, OOD Test, and Cost across different methods (8 Samples per Evolution Epoch). The best and second-best results are denoted by \textbf{bold} and \underline{underline}, respectively. Ground Truth is excluded from ranking.}
\label{tab:ablation}
\resizebox{\textwidth}{!}{%
\begin{tabular}{l cccc cccc ccc}
\toprule
\multirow{2}{*}{\textbf{Method}}
& \multicolumn{4}{c}{\textbf{Test} $\uparrow$}
& \multicolumn{4}{c}{\textbf{OOD Test} $\uparrow$}
& \multicolumn{3}{c}{\textbf{Cost} $\downarrow$} \\
\cmidrule(lr){2-5} \cmidrule(lr){6-9} \cmidrule(lr){10-12}
& Consist. & Dynamic & Interact. & Intent
& Consist. & Dynamic & Interact. & Intent
& VLM & Image & Video \\
\midrule
\textit{Ground Truth}
& \textit{0.8867} & \textit{0.9610} & -- & -- & \textit{0.8555} & \textit{0.9727} & -- & -- & -- & -- & -- \\
\midrule
Before Evo. (Ours)
& 0.6107 & 0.8350 & 0.7367 & 0.7352 & 0.6042 & 0.6075 & 0.5406 & 0.5609 & -- & -- & -- \\
\midrule
LIT (Ours), GPT 5.6 luna
& \underline{0.7923} & \textbf{0.8506} & \underline{0.8281} & \underline{0.7836} & 0.7096 & \underline{0.6758} & 0.7531 & 0.6594 & \textbf{6.06M} & 385 & 1.01h \\
LIT (Ours), Gemini3.8 Flash
& \textbf{0.8177} & 0.8359 & \textbf{0.8734} & \textbf{0.8156} & \textbf{0.7370} & \textbf{0.6797} & \textbf{0.8031} & \textbf{0.6922} & 10.88M & 448 & 1.18h \\
LIT (Ours), Qwen3.8 27B
& 0.7826 & 0.8271 & 0.8219 & 0.7789 & 0.7188 & 0.6699 & \underline{0.7656} & \underline{0.6609} & 7.73M & 361 & 0.93h \\
\midrule
LIT (Ours), w/o Prod. Evo
& 0.6500 & 0.8390 & 0.8016 & 0.7453 & 0.6250 & 0.6219 & 0.7188 & 0.6219 & \underline{6.21M} & \textbf{299} & \underline{0.81h} \\
LIT (Ours), w/o Inte. Evo
& 0.7203 & 0.8453 & 0.7344 & 0.7641 & 0.6969 & 0.6531 & 0.5469 & 0.6031 & 7.18M & 331 & 0.94h \\
LIT (Ours), w/o Co-Evo
& 0.7018 & 0.8320 & 0.8016 & 0.7484 & 0.6744 & 0.6523 & 0.7469 & 0.6188 & 6.31M & \underline{307} & \textbf{0.71h} \\
\midrule
LIT (Ours), w/ 20 Denoising Steps
& 0.7812 & \underline{0.8457} & 0.8250 & 0.7734 & \underline{0.7279} & \textbf{0.6797} & 0.7625 & 0.6500 & -- & -- & -- \\
\bottomrule
\end{tabular}%
}
\end{table*}

\paragraph{User Study.} We conduct a user study with \(15\) participants. Each method generates 16 videos. For each participant, we randomly sample 4 pairs of videos generated by different methods in a blind and randomized order, and the participant selects the preferred one along the corresponding evaluation dimension. We report the number of wins under user evaluation and machine evaluation, respectively. We further compute Kendall's \(\tau\) between human preferences and automatic metrics to validate the rationality of our evaluation metrics. 

As shown in Tab.~\ref{tab:user-preference}, LIT achieves the most user preferences. The machine evaluation metrics achieve an average Kendall's \(\tau\) of 0.833 with human preferences, demonstrating a strong correlation.

%% file: sections/5_conclusion.tex

\section{Conclusion}
We presented LIT, a framework that jointly evolves the interaction
strategies and production workflows of AI video directors.
Through persona-grounded creative tasks, simulated user dialogue,
draft feedback, and execution traces, LIT converts project-level
experience into reusable improvements: the director learns to elicit
latent creative intent and coordinate more reliable production, while
the meta engineer diagnoses and revises both aspects of its behavior.
Crucially, LIT introduces curiosity-driven co-evolution, enabling the
meta engineer to proactively acquire external knowledge and refine its
own evaluation criteria and revision strategies. This expands its
ability to discover and resolve previously uncovered interaction and
production failures, rather than relying on a fixed capacity for
evolution. Experiments on held-out Test and OOD Test tasks show
improvements of 19.1\% in production and 32.0\% in interaction over
the initial pipeline, while user studies further demonstrate stronger
intent fulfillment. Ablation studies confirm that interaction and
production learning are mutually reinforcing and that co-evolution
provides additional gains. Overall, LIT suggests that video directors
can improve across projects not only by learning how to produce better
videos, but also by learning how to collaborate with users and how to
improve the process by which they evolve.

%% file: sections/A1_supplementary_related_work.tex
\section{Supplementary Related Work}

\subsection{AI Video Directors}
AI video directors translate users' creative intent into videos through structured scene planning~\citep{videodirectorgpt2024,videostudio2024} and role-based production coordination~\citep{filmagent2025,movieagent2025,animaker2025}.
Recent human-guided systems support stage-specific review~\citep{animagents2025} and creator-driven refinement~\citep{chief2026}, but rely on predefined interaction protocols.
LIT makes these protocols learnable, evolving interaction policies that adapt to user preferences and improve with experience across projects.

\subsection{Director Agent Self-Evolution}
Self-evolution aims to help video directors better realize users' creative visions.
Existing methods pursue this goal through prompt and layout refinement~\citep{genmac2026,vista2025} or creative configuration search~\citep{codirector2026}.
Others improve workflow instructions~\citep{aesopagent2024}, orchestration skills~\citep{videoweaver2026}, or stage-specific policies~\citep{cineforge2026} across projects.
However, these efforts evolve production while overlooking user interaction.
LIT explicitly models users, thereby driving the mutual improvement of directing and user intent alignment.

\subsection{LLM Agent v.s. Workflow}

\begin{figure}[t]
    \centering
    \includegraphics[width=\linewidth]{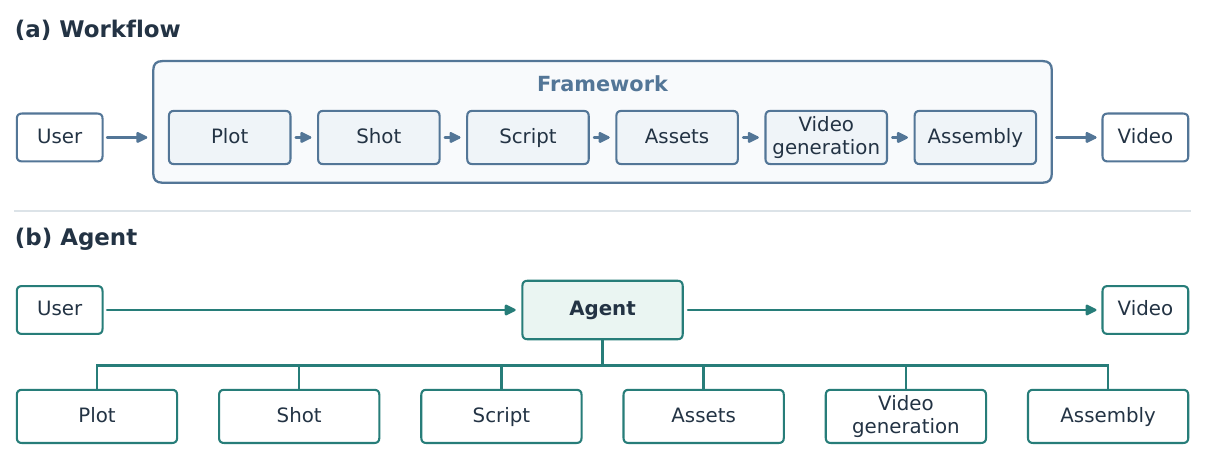}
    \caption{Workflow and agent control in video production. \textbf{Top:} A framework executes six production stages in a predefined sequence. \textbf{Bottom:} An agent accesses the same
  stages as independent capabilities, selecting its next action according to context and feedback. The shared connector indicates access, not execution order.}
    \label{fig:workflow_agent_comparison}
\end{figure}

An agent is an LLM-based system in which the model decides how to proceed and which tools to invoke.
A workflow is a system that coordinates LLM calls and tools through code-defined execution paths~\citep{anthropic2024agents}.
Their difference lies in control: agents determine the course of execution at runtime, whereas workflows prescribe its structure (Fig.~\ref{fig:workflow_agent_comparison}).
LIT is a video-director agent with an evolving workflow-skill harness, combining agent flexibility with the procedural guidance of a workflow.

\paragraph{Agent.}
Early tool-augmented agents established model-directed access to external information and actions~\citep{yao2023react,schick2023toolformer}.
Execution feedback then became a source of reusable knowledge, enabling agents to carry lessons beyond the current reasoning trace~\citep{shinn2023reflexion,wang2023voyager,zhao2024expel}.
This shift from isolated tool calls to sustained execution also motivates general-purpose runtimes.
Claude Code\footnote{\fontsize{8}{9}\url{https://github.com/anthropics/claude-code}} and Codex\footnote{\fontsize{8}{9}\url{https://github.com/openai/codex}} place model-directed action within a persistent working environment~\citep{anthropicclaudecode2025,openaicodex2026}.
This autonomy also introduces execution variability, making procedural guidance important for long-running tasks~\citep{anthropic2024agents,videoweaver2026}.

Video agents constrain long-horizon decisions through specialized production roles~\citep{filmagent2025,movieagent2025}.
Search and refinement further address unreliable individual generations~\citep{animaker2025,codirector2026}.
However, unconstrained agent execution can be unstable: locally plausible decisions may produce an inconsistent production process~\citep{anthropic2024agents,videoweaver2026}.
LIT encodes production workflows as skills that guide the director's execution.
These skills preserve procedural structure and accumulate improvements across projects, providing workflow-based constraints on autonomous decisions.

\paragraph{Workflow.}
Workflow development has moved from explicit orchestration toward automated design.
Component and graph frameworks expose the execution structure for composition, as in LangChain\footnote{\fontsize{8}{9}\url{https://github.com/langchain-ai/langchain}} and LangGraph\footnote{\fontsize{8}{9}\url{https://github.com/langchain-ai/langgraph}}~\citep{langchain2026,langgraph2026}.
With this structure specified, program-level optimization can revise component instructions using downstream feedback~\citep{khattab2024dspy,yuksekgonul2024textgrad,agrawal2026gepa}.
A broader search space also admits changes to the computation itself, which reduces dependence on a manually chosen agent or workflow architecture~\citep{hu2025adas,zhang2025aflow}.
The underlying trade-off is between predictable execution and adaptation to tasks that depart from the prescribed process~\citep{anthropic2024agents}.

In video production, explicit plans coordinate separate generation stages~\citep{videodirectorgpt2024,videostudio2024}.
Evolving workflows and reusable skills allow production experience to improve that sequence~\citep{aesopagent2024,videoweaver2026,cineforge2026}.
However, a fixed execution structure remains rigid when a project requires a different production path.
Improving instructions within that structure alone does not give the system discretion to reorganize it.
LIT represents workflow guidance as revisable skills within an agent architecture.
The director adapts their application to the current project, while skill evolution retains accumulated production expertise.
This combines flexible execution with progressively refined workflow guidance.

\subsection{User Intent Understanding of LLM Agents}
Intent understanding involves acquiring enough information to act and maintaining agreement during execution.
Conversational tool-use benchmarks connect dialogue to observable task outcomes, extending from agent-controlled execution to shared user--agent control~\citep{yao2024taubench,barres2025tau2bench}.
Within this setting, clarification is a decision problem: an unanswered question matters when it changes the appropriate action.
SAGE-Agent formalizes this connection through uncertainty over tool arguments and the expected value of clarification~\citep{suri2026clarification}.

In creative production, feedback on intermediate artifacts helps make an underspecified intention actionable.
AnimAgents organizes that feedback across pre-production stages~\citep{animagents2025}.
CHIEF extends the loop to generated videos, combining creator revisions with audience-conditioned critique~\citep{chief2026}.
User feedback alone cannot correct an interaction policy that fails to elicit useful creative direction.
Users may struggle to articulate preferences, while poorly realized previews make those preferences harder to judge.
LIT jointly improves interaction and production through simulated collaborations.
Dialogue reveals where guidance is ineffective, and production outcomes reveal whether the elicited intent was actionable.
The resulting updates connect better creative communication to more faithful execution.

\begin{tcolorbox}[colback=black!2,colframe=black!25,boxrule=0.4pt,
  arc=1pt,left=7pt,right=7pt,top=5pt,bottom=5pt]
\small
\textbf{Interaction as a learning problem:}

1. Which \textit{uncertainty} should be resolved with the user before making a production commitment?

2. How should the director provide \textit{guidance and inspiration} to the user?

3. What can the director give the user as \textit{feedback} to obtain valuable suggestions?

\end{tcolorbox}

\subsection{Long Video Generation and Evaluation}
Long-video generation requires coherence beyond a local generation window.
A short clip, such as a 15-second segment, may depict a single event. A longer production must preserve story state across successive events and shot transitions.
Duration extension and multi-shot storytelling therefore pose related but distinct problems~\citep{kim2024fifo,kara2025shotadapter}.
In agent-directed production, the relevant horizon also includes decisions made before rendering, whose consequences can persist throughout the film~\citep{movieagent2025,cineforge2026}.

\subsubsection{Long Video Generation}
Two main approaches address this extended horizon: model-level temporal generation and production-level orchestration.
FIFO-Diffusion extends pretrained generators through queued denoising, while ShotAdapter introduces shot-specific conditioning within a diffusion model~\citep{kim2024fifo,kara2025shotadapter}.
The orchestration approach instead decomposes a story into controllable generation tasks.
VideoDirectorGPT establishes an LLM-guided planning scheme, and VideoStudio uses shared entity information for consistent multi-scene synthesis~\citep{videodirectorgpt2024,videostudio2024}.
These methods connect local synthesis to longer-range visual structure.

MovieAgent coordinates production through hierarchical planning~\citep{movieagent2025}.
AniMaker and Co-Director use search to refine generation decisions~\citep{animaker2025,codirector2026}.
CHIEF incorporates creator feedback into video refinement~\citep{chief2026}, while production-evolution methods retain improvements across projects~\citep{videoweaver2026,cineforge2026}.
Visual continuity does not ensure fidelity to the intended story.
A mistaken interpretation established during planning can persist across an otherwise coherent sequence.
Production-only optimization leaves the interaction that produced this interpretation uncorrected.
LIT uses dialogue alongside production traces to revise both intent understanding and its realization throughout the workflow.

Tab.~\ref{tab:rw_video_capabilities} distinguishes generation capabilities from the mechanisms used to improve them.
User interaction concerns collaboration during creation. Meta co-evolution concerns joint improvement of the production system and its meta-level optimizer.
Defect attribution includes critic-based identification of the production stage responsible for an error.
\begin{table}[t]
\centering
\scriptsize
\setlength{\tabcolsep}{3pt}
\renewcommand{\arraystretch}{1.2}
\caption{Generation, interaction, and evolution mechanisms in video-production methods.
$\checkmark$: explicitly described, --: not established in the cited method.
Storyboard denotes planning without end-to-end video synthesis.}
\label{tab:rw_video_capabilities}
\resizebox{\linewidth}{!}{%
\begin{tabular}{@{}lcccccccc@{}}
\toprule
\textbf{Method} &
\shortstack{\textbf{Long-horizon}\\\textbf{Video}} &
\shortstack{\textbf{Self}\\\textbf{Evolution}} &
\shortstack{\textbf{Defect}\\\textbf{Attribution}} &
\shortstack{\textbf{Proactive}\\\textbf{Exploration}} &
\shortstack{\textbf{Prompt}\\\textbf{Update}} &
\shortstack{\textbf{Workflow}\\\textbf{Update}} &
\shortstack{\textbf{User}\\\textbf{Interaction}} &
\shortstack{\textbf{Meta}\\\textbf{Co-evolution}} \\
\midrule
 ViMax~\citep{huang2026vimax}
  & $\checkmark$ & -- & -- & -- & -- & -- & -- & -- \\ 
  CHIEF~\citep{chief2026}
  & $\checkmark$ & -- & -- & -- & $\checkmark$ & -- & $\checkmark$ & -- \\
\midrule
GenMAC~\citep{genmac2026}
  & -- & -- & $\checkmark$ & -- & $\checkmark$ & -- & -- & -- \\
Co-Director~\citep{codirector2026}
  & $\checkmark$ & -- & $\checkmark$ & -- & $\checkmark$ & -- & -- & -- \\
\midrule
AesopAgent~\citep{aesopagent2024}
  & $\checkmark$ & $\checkmark$ & -- & -- & $\checkmark$ & $\checkmark$ & -- & -- \\
VideoWeaver~\citep{videoweaver2026}
  & $\checkmark$ & $\checkmark$ & $\checkmark$ & -- & $\checkmark$ & $\checkmark$ & -- & -- \\
CineForge~\citep{cineforge2026}
  & $\checkmark$ & $\checkmark$ & $\checkmark$ & -- & $\checkmark$ & $\checkmark$ & -- & -- \\
\midrule
\textbf{LIT (Ours)}
  & $\checkmark$ & $\checkmark$ & $\checkmark$ & $\checkmark$ & $\checkmark$ & $\checkmark$ & $\checkmark$ & $\checkmark$ \\
\bottomrule
\end{tabular}%
}
\end{table}

\subsubsection{Long Video Evaluation}
Video evaluation has expanded from perceptual quality to the realization of structured events.
VBench measures distinct dimensions of visual quality and consistency, while VBench-2.0 adds tests of physical and semantic faithfulness~\citep{huang2024vbench,zheng2025vbench2}.
At the production level, FilmBench evaluates cinematic control, and CineScope measures episode-level narrative and directorial coherence~\citep{filmbench2026,cineforge2026}.
VideoWeaver further uses execution traces to assess the generation process~\citep{videoweaver2026}.
This progression provides evidence about both the completed film and the procedure that produced it.

Conversational benchmarks assess task outcomes together with user--agent coordination~\citep{yao2024taubench,barres2025tau2bench}.
In video production, judging against the disclosed prompt cannot reveal requirements that the director failed to elicit.
Output and execution scores alone also leave the source of creative misalignment ambiguous.
LIT evaluates the completed film against a private creative specification and examines the dialogue that informed production.
This distinguishes a requirement that was never elicited from one that was understood but poorly executed.
The distinction supports targeted improvement of the interaction policy or production workflow while keeping final-video quality independently measurable.

\subsection{LLM Agent Self-evolution}
Agent self-evolution has expanded from retaining experience to revising the system that uses it.
Early approaches preserve feedback as external knowledge, improving later decisions without changing model weights~\citep{shinn2023reflexion,wang2023voyager,zhao2024expel}.
Subsequent work treats the agent program as an optimization variable, allowing execution evidence to change its instructions or structure~\citep{agrawal2026gepa,hu2025adas,zhang2025dgm}.
A further direction adapts the learning conditions themselves through coupled curricula or an evolving improvement procedure~\citep{wang2019poet,xia2025agent0,ye2026mce}.

In video generation, feedback first supports refinement of the current output~\citep{genmac2026,vista2025}.
Cross-project methods preserve its benefits by updating reusable production procedures~\citep{aesopagent2024,videoweaver2026,cineforge2026}.
However, production-only updates cannot repair failures caused by an unchanged collaboration policy.
They can improve execution of the same misunderstood objective.
LIT uses interaction and production evidence together to revise the director's behavior across projects.
Its meta-engineer also learns from the outcomes of these revisions, allowing the improvement strategy to adapt as the director's weaknesses change.

\subsubsection{Preliminaries}
Let $\theta$ denote an agent's editable configuration and $\mu$ a distribution over tasks $x$.
The configuration induces an execution policy $\pi_\theta$ over trajectories $\tau$.
For a task-level score $R$, expected performance is
\begin{equation}
J(\theta,\mu)
=\mathbb{E}_{x\sim\mu,\;\tau\sim\pi_\theta(\cdot\mid x)}
\bigl[R(\tau,x)\bigr].
\label{eq:rw_evolution_objective}
\end{equation}
Prompt and agent-program optimization search over $\theta$ using observed performance~\citep{khattab2024dspy,agrawal2026gepa,zhang2025dgm}.
This search can operate on external resources while the underlying model weights remain fixed.

Let $\mathcal{T}_k$ contain the trajectories and feedback collected at iteration $k$.
An update procedure $\mathcal{U}$ with configuration $\omega_k$ produces
\begin{equation}
\theta_{k+1}=\mathcal{U}_{\omega_k}(\theta_k,\mathcal{T}_k),
\qquad
\mathcal{T}_k\sim\operatorname{Rollout}(\pi_{\theta_k},\mu_k).
\label{eq:rw_evolution_update}
\end{equation}
Adaptive curricula change $\mu_k$, thereby changing which weaknesses become observable~\citep{wang2019poet,wang2023voyager,xia2025agent0}.
Meta-level adaptation can also revise the strategy that constructs improved configurations~\citep{ye2026mce,videoweaver2026}.
In LIT, $\theta_k$ includes the director's interaction and production policies, while $\omega_k$ specifies the meta-engineer's improvement strategy.
Meta co-evolution makes both quantities editable. Adaptation of $\mu_k$ supplies learning opportunities for this coupled process.

\subsubsection{Curiosity-driven Proactive Evolution}
Proactive improvement depends on finding informative experience beyond the tasks already encountered.
Intrinsic-motivation methods use prediction-based novelty to guide exploration when external rewards provide little direction~\citep{pathak2017curiosity,burda2019rnd}.
Adaptive curricula move this choice from individual states to whole tasks, matching new challenges to the learner's current capabilities~\citep{wang2019poet,wang2023voyager,xia2025agent0}.
The resulting progression changes what the agent practices, as well as how it learns from each attempt.

Video search largely explores alternative realizations of an existing request.
VISTA revises generation prompts, while Co-Director searches creative configurations before local refinement~\citep{vista2025,codirector2026}.
Persistent improvement can then use completed productions to identify recurring weaknesses~\citep{videoweaver2026,cineforge2026}.
Reactive improvement cannot address a weakness that existing tasks never expose or the evaluator fails to recognize.
Repeatedly optimizing the same observed defects can therefore leave entire capabilities unexplored.
LIT's meta-engineer examines external productions and technical analyses to identify untested capabilities and neglected evaluation criteria.
It probes candidate improvements through directed experiments, allowing evolution to investigate problems absent from the existing failure record.

\subsubsection{Meta Co-Evolution}
Meta co-evolution concerns the coupling between an improving system and its improvement mechanism.
Agent-design methods search over candidate systems using execution feedback~\citep{hu2025adas,zhang2025aflow,zhang2025dgm}.
Adaptive curricula expand this search by changing the challenges presented to the learner~\citep{wang2019poet,xia2025agent0}.
An additional degree of adaptation arises when the strategy that constructs improved systems is itself revised~\citep{ye2026mce}.
Progress then depends on learning effective revisions as well as retaining their results.

Within video production, VideoWeaver evolves composition skills and the creator skills used to construct them~\citep{videoweaver2026}.
CineForge instead formalizes how recurrent defects become validated production-policy updates~\citep{cineforge2026}.
Updating production resources does not by itself improve the optimizer's diagnosis and revision strategy.
A fixed improvement procedure can retain the same blind spots even as the production system changes.
LIT exposes the meta-engineer's governing skills to revision alongside the director's skills.
The consequences of earlier interventions inform later problem selection and solution development.
The improvement mechanism can therefore adapt its diagnosis and revision strategy as the production system develops.

%% file: sections/A2_formal_representation.tex
\section{Formal Representation of Evolution Process}
\label{app:formal-evolution}

\subsection{Optimization Target}
\label{app:formal-target}

LIT optimizes directing and interaction quality. Task requirements define the
target risk, whereas the meta engineer's current evaluation criteria define
the empirical objective available during evolution.

\paragraph{Definition.}
Let $\theta\in\Theta$ denote a director policy specified by prompts, skills,
tools, and workflows. A task $t=(z,u)$ comprises the complete creative intent
$z$ and user behavior $u$.
The trajectory $\tau=\mathsf{Rollout}(\theta,t;\xi)$ contains the dialogue,
production trace, and final video. $\xi$ collects execution randomness.
The distribution $P$ describes the target population of user commissions.
ID and OOD performance are analyzed under their respective distributions.
The analysis applies separately to directing and interaction quality.
We omit the dimension index until needed.

Let $\ell^\star(t,\tau)\in[0,1]$ denote the true task loss. The optimization
target is the expected risk
\begin{equation}
\label{eq:lit-true-risk}
\mathcal R_P^\star(\theta)
=
\mathbb E_{t\sim P,\xi}
\left[\ell^\star(t,\mathsf{Rollout}(\theta,t;\xi))\right],
\qquad
\inf_{\theta\in\Theta}\mathcal R_P^\star(\theta).
\end{equation}
At epoch $k$, the evaluation program $G_k$ assigns a computable loss
$\ell_{G_k}\in[0,1]$. For the sampled tasks and executions,
\begin{equation}
\label{eq:lit-empirical-risk}
\begin{aligned}
\widehat{\mathcal R}_k(\theta;G)
&=
\frac{1}{n_k}\sum_{i=1}^{n_k}
\ell_G(t_{k,i},\mathsf{Rollout}(\theta,t_{k,i};\xi_{k,i})),\\
\mathcal R_G(\theta)
&=
\mathbb E_{t\sim P,\xi}
\left[\ell_G(t,\mathsf{Rollout}(\theta,t;\xi))\right].
\end{aligned}
\end{equation}
The empirical risk guides director revisions, while the meta engineer
updates the evaluation criteria and the procedures used to measure them.
Final testing estimates Eq.~\ref{eq:lit-true-risk} on held-out tasks using
independent criteria.

\subsection{Co-evolution Process}
\label{app:formal-process}

We use the policy--evaluation game of adversarial apprenticeship learning
to describe co-evolution~\citep{swamy2021moments}: the meta engineer searches
for evaluations that reveal deficiencies, and the director changes its
behavior accordingly. In LIT, $D_k$ and $U_k$ specify the director and
simulated-user policies, respectively. The meta-policy $M_k$ governs
task selection, evaluation, and the proposal and selection of revisions.

\paragraph{Definition.}
Let $E\in\Theta$ be a fixed, executable reference director, such as a
historical version. External videos, comments, and technical analyses
inform criterion design. Let $\mathcal G$ be the admissible evaluation
class, including the zero function. Define
\begin{equation}
\label{eq:lit-evaluation-game}
\begin{aligned}
L(\theta,G)&=\mathcal R_G(\theta)-\mathcal R_G(E),\\
D(\theta)&=\sup_{G\in\mathcal G}L(\theta,G),
\qquad
\inf_{\theta\in\Theta}D(\theta)=0.
\end{aligned}
\end{equation}
The last equality follows from $D(E)=0$ and $D(\theta)\ge0$.
The empirical game is
$\widehat L_k(\theta,G)=
\widehat{\mathcal R}_k(\theta;G)-\widehat{\mathcal R}_k(E;G)$.
Estimating the reference term requires execution evidence on comparable
tasks. Define
\begin{equation}
\label{eq:lit-evolution-errors}
\begin{aligned}
\epsilon_k
&=\sup_{G\in\mathcal G}\widehat L_k(\theta_k,G)
-\widehat L_k(\theta_k,G_k),\\
e_k
&=\sup_{\theta\in\Theta,G\in\mathcal G}
|\widehat L_k(\theta,G)-L(\theta,G)|,\\
B_T
&=\left[
\sum_{k=1}^{T}\widehat L_k(\theta_k,G_k)
-\inf_{\theta\in\Theta}\sum_{k=1}^{T}\widehat L_k(\theta,G_k)
\right]_+.
\end{aligned}
\end{equation}
Here $\epsilon_k$ is the gap between the selected evaluation and the
evaluation in $\mathcal G$ that reveals the largest discrepancy.
The uniform estimation error $e_k$ accounts for task sampling, reference
estimation, and scoring, while $B_T$ is the cumulative regret relative to
the best fixed director. External observations provide evidence of new
defects. Reflection and candidate testing use computation to identify
and assess repairs.

For the delivered version
$\theta_{\mathrm{out}}\in\{\theta_1,\ldots,\theta_T\}$, let
$s_T=D(\theta_{\mathrm{out}})-\min_{k\le T}D(\theta_k)$ be the selection
error, and define
\begin{equation}
\label{eq:lit-joint-error}
\eta_T
=
\frac{B_T}{T}
+\frac1T\sum_{k=1}^{T}(\epsilon_k+2e_k)+s_T.
\end{equation}

\subsection{Capability Upper and Lower Bounds}
\label{app:formal-capability}

We prove that LIT has a \textbf{capability upper bound at least as high as} that of a pipeline whose meta-agent keeps its evaluation criteria fixed throughout evolution. Its guaranteed \textbf{minimum performance is also higher} when the gains from correcting evaluation blind spots exceed the remaining errors.

\paragraph{Definition.}
Under the same initial resources, environment access, and computation
budget $b$, let $\mathcal A_0(b)$ and $\mathcal A_1(b)$ be the director
policies reachable by the baseline and LIT. In the nested design considered
here, LIT retains every baseline path at its original execution cost.

The baseline optimizes the fixed scoring loss
$f_0(\theta)=\mathcal R_{G_0}(\theta)$. For any baseline tolerance
$\eta_0\ge0$, define
\begin{equation}
\label{eq:lit-output-sets}
\begin{aligned}
\mathcal S_0
&=\left\{\theta\in\mathcal A_0(b):
f_0(\theta)-\inf_{\vartheta\in\mathcal A_0(b)}f_0(\vartheta)
\le\eta_0\right\},\\
\mathcal S_1
&=\{\theta\in\mathcal A_1(b):D(\theta)\le\eta_T\},\\
\overline Q_i
&=1-\inf_{\theta\in\mathcal A_i(b)}\mathcal R_P^\star(\theta),
\qquad
\underline Q_i
=1-\sup_{\theta\in\mathcal S_i}\mathcal R_P^\star(\theta).
\end{aligned}
\end{equation}
Thus, $\overline Q_i$ is the best reachable quality, while
$\underline Q_i$ is the minimum quality among outputs satisfying the
method's optimization requirement.
Assume that the fixed score has an optimal but low-quality policy:
\begin{equation}
\label{eq:lit-baseline-blind-spot}
\theta_b\in\arg\min_{\theta\in\mathcal A_0(b)}f_0(\theta),
\qquad
\mathcal R_P^\star(\theta_b)-\mathcal R_P^\star(E)
\ge\Delta>0.
\end{equation}
Let the representation error of the evolved evaluation class and the
total residual error be
\begin{equation}
\label{eq:lit-representation-error}
\delta
=
\inf_{G\in\mathcal G}\sup_{\theta\in\Theta}
|\mathcal R_P^\star(\theta)-\mathcal R_G(\theta)|,
\qquad
\rho_T=2\delta+\eta_T.
\end{equation}

\paragraph{Proposition.}
Under the preceding definitions and assumptions,
\begin{equation}
\label{eq:lit-capability-bounds}
\overline Q_1\ge\overline Q_0,
\qquad
\underline Q_1-\underline Q_0
\ge\Delta-\rho_T>0
\quad\text{if }\rho_T<\Delta.
\end{equation}

\paragraph{Proof.}
Preserving baseline paths gives
$\mathcal A_0(b)\subseteq\mathcal A_1(b)$.
Taking infima of the true risk proves the upper-bound claim.

For co-evolution, Eq.~\ref{eq:lit-evolution-errors} and the reference
policy's zero game loss give
\begin{equation}
\label{eq:lit-game-error-bound}
\begin{aligned}
\frac1T\sum_{k=1}^{T}D(\theta_k)
&\le
\frac1T\sum_{k=1}^{T}
[\widehat L_k(\theta_k,G_k)+\epsilon_k+e_k]\\
&\le
\frac{B_T}{T}
+\frac1T\sum_{k=1}^{T}(\epsilon_k+2e_k).
\end{aligned}
\end{equation}
Including $s_T$ yields $D(\theta_{\mathrm{out}})\le\eta_T$,
so the delivered policy belongs to $\mathcal S_1$.
An evaluation program arbitrarily close to the infimum in
Eq.~\ref{eq:lit-representation-error} approximates both the director's and
the reference's true losses, giving
\begin{equation}
\label{eq:lit-risk-transfer}
\mathcal R_P^\star(\theta)-\mathcal R_P^\star(E)
\le D(\theta)+2\delta.
\end{equation}
Consequently,
$\underline Q_1\ge1-\mathcal R_P^\star(E)-\rho_T$.
Since $\theta_b$ minimizes the baseline's fixed loss,
$\theta_b\in\mathcal S_0$ for every $\eta_0\ge0$, and
$\underline Q_0\le1-\mathcal R_P^\star(E)-\Delta$.
Subtracting proves Eq.~\ref{eq:lit-capability-bounds}.
\hfill$\square$

\paragraph{Conclusion.}
LIT has a capability upper bound at least as high as that of the
fixed-evaluation baseline because it preserves all baseline evolution
paths. Its capability lower bound also improves with more discriminative
evaluation criteria (provided $\Delta>\rho_T$).
This result motivates jointly improving evaluation
and director behavior: identify overlooked defects, repair them, and
control the errors introduced by these revisions.

%% file: sections/A3_pipeline_implementation.tex
\section{Pipeline Implementation}
\label{app:implementation}
LIT improves a directing system by changing both how it produces an
artifact and how it recognizes a useful improvement. Within a project,
the User discloses intent through interaction, and the Director turns
that information into a workflow. Across projects, Meta uses the
resulting evidence to revise the workflow and its evaluation criteria.
When the current criteria overlook a quality property, Meta can discover
it through external references and adapt tasks to exercise the methods
developed from that discovery. The Controller connects these two loops by executing the
selected role versions.

The implementation makes a production stage the link between execution
and evolution. Skills determine how the stage runs, while its input and
output provide a boundary at which Meta can investigate and test a
change. Infrastructure supplies these executable units and access to
external knowledge. The Director composes them into a project, and the
User supplies the interaction through which its requirements become
known. The Meta section explains how evidence from this process leads
to changes in each role and in Meta's own methods.

\subsection{Infrastructure}
Role workflows need a common way to invoke capabilities and retain the
artifacts those calls produce. We expose the deployed models through
skills, so a workflow can change its directing method while continuing
to use the same execution interfaces. The skill hierarchy allows workflows to compose these operations into
reusable procedures. To improve a procedure, Meta can consult the
reference database and develop a revision that the Controller makes
available in the next project.

\subsubsection{Deployment Details}

\begin{table*}[t]
\centering
\begin{minipage}[t]{0.49\textwidth}
\vspace{0pt}
\setlength{\tabcolsep}{3pt}
\caption{Inference settings for Qwen3.8.}
\label{tab:lit-agent-deployment}
\scriptsize
\begin{tabular}{@{}p{0.45\linewidth}p{0.50\linewidth}@{}}
\toprule
\textbf{Setting} & \textbf{Value}\\
\midrule
Quantization / compute dtype & W8A8 / BF16\\
vLLM / vLLM-Ascend & 0.26.0 / 0.26.0rc1\\
Hosts / replicas per host & 2 Ascend / 4\\
Tensor / pipeline parallelism & 4 / 1 per replica\\
Context / compaction threshold & 262,144 / 220,000 tokens\\
Evolution reasoning effort & low\\
Evaluations per video & Evolution: 1, final test: 4 at xhigh\\
MTP method / draft length & \texttt{qwen3\_5\_mtp} / 1 token\\
Draft execution & Eager\\
Image / video admission limit & 256 images / 16 videos per request\\
Image pixel-area range & 65,536--16,777,216\\
Video processor sampling & 2 fps, 4--768 frames, limited by source length\\
Video spatiotemporal pixel budget & 4,096--25,165,824\\
Video-inspection adapter & Timestamped images at 2 fps\\
Temperature / top-$p$ / top-$k$ defaults & 1.0 / 0.95 / 20\\
Completion-token limit & Remaining context, subject to request cap\\
Batch tokens / concurrent sequences & 16,384 / 8 per replica\\
Device-memory utilization target & 0.90\\
KV-cache dtype / prefix caching & \texttt{auto} / enabled\\
Chunked prefill & Enabled\\
Reasoning / tool-call parser & \texttt{qwen3} / \texttt{qwen3\_coder}\\
Agent endpoint & First host's four-replica pool\\
Replica routing / session affinity & Least connections / no pinning\\
Session history / prefix cache & Harness-managed / replica-local\\
Harness concurrent-task limit & 16\\
\bottomrule
\end{tabular}
\end{minipage}\hfill
\begin{minipage}[t]{0.49\textwidth}
\vspace{0pt}
\setlength{\tabcolsep}{3pt}
\caption{Video generation settings for MiniMax-H3.}
\label{tab:lit-h3-deployment}
\scriptsize
\begin{tabular}{@{}p{0.45\linewidth}p{0.50\linewidth}@{}}
\toprule
\textbf{Setting} & \textbf{Value}\\
\midrule
Hardware & 4 MetaX C550 hosts\\
ComfyUI / PyTorch & 0.34.5 / \texttt{2.8.0+metax3.7.0.7}\\
Checkpoints & Ref2VA / FL2VA\\
Turbo LoRA / strength & Matching four-step LoRA / 1.0\\
Ref2VA / FL2VA workers & 22 / 8\\
Worker distribution & Two hosts: $5+2$ each, two hosts: $6+2$ each\\
Sampler / schedule / steps & Euler / simple / 4\\
Seed & 42\\
Video sigma shift & Ref2VA: 12, FL2VA: 6\\
Audio sigma shift & 3\\
Frame rate & 24 fps\\
Production resolution & 480P\\
Adapter canvas support & 480P / 768P, dimensions rounded to multiples of 32\\
Duration & Integer 4--15 s, default 5 s\\
Default aspect ratio & 16:9\\
Diffusion / text-encoder dtype & BF16 / BF16\\
Video / audio VAE dtype & FP16 / FP32\\
Attention backend & PyTorch\\
Conditioning routes & Separate reference and first/last-frame pools\\
Reference / endpoint-frame resizing & Aspect-preserving / letterboxed to output canvas\\
Worker routing & Random within compatible pool\\
Polling / download affinity & Same worker as submission\\
Polling interval / HTTP timeout & 5 s / 30 s\\
Job deadline / cooldown & 1,800 s / 1,200 s\\
Worker storage & Separate inputs, outputs, and queues\\
Completed-result reuse & Fingerprint of parameters and references\\
\bottomrule
\end{tabular}
\end{minipage}
\end{table*}

\paragraph{Qwen3.8-27B.}
All evolution roles use Qwen3.8-27B\footnote{\fontsize{8}{9}
\url{https://huggingface.co/Qwen/Qwen3.8-27B}.}
with the configuration in Tab.~\ref{tab:lit-agent-deployment}.
Inference uses vLLM\footnote{\fontsize{8}{9}\url{https://github.com/vllm-project/vllm}.}
with the Ascend backend,\footnote{\fontsize{8}{9}
\url{https://github.com/vllm-project/vllm-ascend}.}
accessed through Codex's app-server interface for persistent role
sessions and tool execution.\footnote{\fontsize{8}{9}
\url{https://developers.openai.com/codex/app-server}.}

\paragraph{MiniMax-H3.}
We use MiniMax-H3\footnote{\fontsize{8}{9}
\url{https://huggingface.co/MiniMaxAI/MiniMax-H3}.}
through ComfyUI\footnote{\fontsize{8}{9}\url{https://github.com/Comfy-Org/ComfyUI}.}
with matching Turbo LoRAs\footnote{\fontsize{8}{9}
\url{https://huggingface.co/lightx2v/Minimax-h3-Turbo}.}
and the configuration in Tab.~\ref{tab:lit-h3-deployment}.

\paragraph{Other Models.}
Supporting models let a directing workflow prepare a reference or repair
one part of a shot without regenerating all of its content.
Tab.~\ref{tab:lit-media-settings} lists the nine supporting capability
interfaces and their configurations. The following description explains
how generation, inspection, and retrieval support a repair.
Nano-Banana-2-Lite supplies image references.\footnote{\fontsize{8}{9}
\url{https://ai.google.dev/gemini-api/docs/image-generation}.}
For sound, MiniMax-Music3 generates music from a description and separate
lyrics,\footnote{\fontsize{8}{9}\url{https://github.com/MiniMax-AI/MiniMax-Music3}.}
while Woosh generates effects either from text or conditioned on the
shot's timing.\footnote{\fontsize{8}{9}\url{https://github.com/SonyResearch/Woosh}.}
When existing audio needs revision, Mel-Band RoFormer separates the
vocal and background stems,\footnote{\fontsize{8}{9}
\url{https://github.com/lucidrains/BS-RoFormer}.}
and Qwen3-TTS-Instruct-Flash supplies speech with the chosen delivery.\footnote{\fontsize{8}{9}
\url{https://github.com/QwenLM/Qwen3-TTS}.}

The same workflow needs observations with which to decide what to
repair. To distinguish speech-content errors from acoustic problems,
we use Qwen-Audio-3.0-ASR-Flash for timestamped transcription and
Qwen3.5-Omni-Plus for assessment of the sound itself.\footnote{\fontsize{8}{9}
\url{https://arxiv.org/abs/2604.15804}.}
Reference selection uses Qwen3-VL-Embedding to retrieve candidates,
then Qwen3-VL-Rerank or Qwen3-Rerank to compare multimodal or text
candidates, respectively.\footnote{\fontsize{8}{9}
\url{https://github.com/QwenLM/Qwen3-Embedding}.}

Task-specific adapters keep this model selection separate from service
routing. Each local audio capability has two replicas, with inference
serialized within a replica. The client distributes requests between
them unless an endpoint is explicitly selected. API-backed tools use
their configured endpoint. Each media result retains its request
arguments and reference links, allowing a later revision to recover
the inputs responsible for that artifact.

\begin{table*}[t]
\centering
\caption{Configurations and interfaces of supporting media tools.}
\label{tab:lit-media-settings}
\scriptsize
\begin{tabular}{p{0.19\linewidth}p{0.73\linewidth}}
\toprule
\textbf{Capability} & \textbf{Configuration and interface}\\
\midrule
Image generation & Nano-Banana-2-Lite, default 1K, 16:9.
The request includes the prompt and optional reference images.
The adapter downloads the result and retains its request record.\\
Music & BF16 inference, seed 7, default duration 10 s.
The token budget is the requested duration multiplied by 25.
The server accepts 25--9,000 output tokens and returns PCM16 WAV.\\
Text-to-sound & Woosh-DFlow, 5 s, seed 31001, Euler, 4 steps,
CFG 4.5, renoising schedule $(0,0.5,0.5,0.3)$, 48 kHz WAV.\\
Video-to-sound & Woosh-VFlow-8s, first 8 s of the reference video,
Synchformer at 24 fps, seed 32001, CFG 4.5,
integration tolerances $10^{-3}$, 48 kHz WAV.\\
Audio separation & MelBand-RoFormer-Kim-Vocals.
Inputs are converted to 44.1 kHz stereo. The output contains separate
vocal and background PCM16 WAV files.\\
Transcription & Qwen-Audio-3.0-ASR-Flash-Filetrans, channel 0,
optional language hints, asynchronous polling every 2 s with a
900 s deadline. Utterance timestamps are exported in seconds.\\
Audio understanding & Qwen3.5-Omni-Plus, streaming text response.
Each base64-encoded audio input must be below 10 MiB.
The client timeout is 900 s.\\
Speech synthesis & Qwen3-TTS-Instruct-Flash, default Chinese and
voice Cherry, with optional delivery instructions.
Voice cloning uses Qwen3-TTS-VC-2026-01-22. Voice design uses
Qwen3-TTS-VD-2026-01-26. A selected voice identifier can be reused
across utterances.\\
Asset retrieval & The current adapter uses Qwen3-VL-Embedding,
1,024-dimensional cosine vectors and fusion disabled.
The default retrieval count is 10, and the synchronization batch size is 64.
Text candidates use Qwen3-Rerank. Multimodal candidates use
Qwen3-VL-Rerank.\\
\bottomrule
\end{tabular}
\end{table*}

\subsubsection{Skill Hierarchy}
A reusable method must specify both an operation and the context in
which to use it. LIT uses three skill types: atomic skills
provide callable operations, workflow skills determine their
dependencies, and library skills select the applicable method. Meta
can therefore change a procedure at the level where the defect arises
and make the revised procedure available to future projects.

\paragraph{Atomic Skill.}
An atomic skill defines one reusable decision or operation that a
workflow can invoke, whether the operation is performed through reasoning,
a media tool, or an agent session. For example, treating H3 prompt
construction as an atomic skill allows Meta to revise that procedure
independently of the generation call that consumes its output.
Tab.~\ref{tab:lit-atomic-skills} groups further examples by their role
in execution and identifies the optional extensions.

\begin{table}[t]
  \centering
  \setlength{\tabcolsep}{3pt}
    \caption{Atomic skills and optional extensions for video production.}
  \label{tab:lit-atomic-skills}
  \scriptsize
  \begin{tabular}{@{}p{0.25\linewidth}p{0.67\linewidth}@{}}
    \toprule
    \textbf{Role} & \textbf{Examples}\\
    \midrule
    Production operations & Image/video generation, frame extraction,
    transcription, audio separation, clip assembly.\\
    Directing procedures & Shot planning, H3 prompt construction,
    reference-asset selection, spatial-anchor instructions.\\
    Runtime operations & Agent delegation, session continuation,
    task-result submission.\\
    Optional visual analysis & DINOv3 features, SAM 3 segmentation,
    RelateAnything relations, Orient Anything V2 orientation,
    Depth Anything 3 geometry.\\
    Optional 3D workbench & Panorama conditioning, SHARP Gaussian
    reconstruction, depth alignment, splat-to-voxel conversion.\\
    \bottomrule
  \end{tabular}
\end{table}

The five optional visual-analysis adapters in
Tab.~\ref{tab:lit-atomic-skills} use the original interfaces of
DINOv3,\footnote{\fontsize{8}{9}\url{https://github.com/facebookresearch/dinov3}.}
SAM 3,\footnote{\fontsize{8}{9}\url{https://github.com/facebookresearch/sam3}.}
RelateAnything,\footnote{\fontsize{8}{9}\url{https://github.com/Maelic/RelateAnything}.}
Orient Anything V2,\footnote{\fontsize{8}{9}
\url{https://github.com/SpatialVision/Orient-Anything-V2}.}
and Depth Anything 3.\footnote{\fontsize{8}{9}
\url{https://github.com/ByteDance-Seed/Depth-Anything-3}.}
The workflow selects an adapter according to the decision it needs to
make. Appearance comparison uses visual features, with segmentation
available to isolate the subject. When a decision concerns spatial
consistency, relation, orientation, or depth estimates provide the
relevant evidence.

For methods requiring an editable spatial reference, the 3D workbench
connects panorama conditioning to Gaussian reconstruction through
SHARP.\footnote{\fontsize{8}{9}\url{https://github.com/apple-aiml-research/ml-sharp}.}
DA$^2$ provides panorama depth for alignment,\footnote{\fontsize{8}{9}
\url{https://huggingface.co/haodongli/DA-2}.}
after which \texttt{splat-transform} processing and geometric voxel
conversion produce an editable representation.\footnote{\fontsize{8}{9}
\url{https://github.com/playcanvas/splat-transform}.}
These stages remain separate operations, so a workflow can use the
spatial representation it needs without invoking the entire path.

\paragraph{Workflow Skill.}
A workflow skill turns callable operations into a procedure by specifying
what must be available before the next step and how its result affects
the subsequent decision. It can call a smaller workflow just as a
function calls a subroutine. Within the Controller's outer workflow,
the Director executes its role workflow by delegating to storyboard
and shot-generation workflows. Natural-language branches and loops
allow these calls to respond to the current evidence.

A child receives an upstream artifact and an expected deliverable.
Its parent uses the returned artifact to continue or revisit the
responsible step. For example, a storyboard establishes the references
needed for asset preparation, and those assets condition segment
generation. If one segment fails while its upstream plan remains valid,
execution can re-enter that shot's procedure. The same input--output
boundary lets Meta test a change without repeating plot development.

\paragraph{Library Skill.}
A workflow needs to find a suitable procedure before it can reuse it.
A library skill provides this selection rule by mapping an applicable
situation to a skill or resource path. Leaf entries hold the executable
instructions at the granularity needed by the task. Selection proceeds
through the hierarchy, for example from a narrative-production entry
to a genre library and then to the relevant storyboard or sound procedure.
Only the selected instructions need to enter the agent's context.

The parent directory determines when a leaf method is selected, while
the leaf provides its execution instructions. Incorporating an improvement
therefore requires updating both the instructions and the routing entry
that makes the method available in the appropriate situations.

\begin{wraptable}{r}{0.30\textwidth}
  \centering
  \setlength{\tabcolsep}{3pt}
    \caption{Composition of the local reference database.}
  \label{tab:lit-reference-inventory}
  \scriptsize
  \begin{tabular}{@{}lr@{}}
    \toprule
    \textbf{Reference type} & \textbf{Records} \\
    \midrule
    Documents and prompt articles & 355 \\
    Skills & 21 \\
    Workflows & 85 \\
    Code references & 12 \\
    Contiguous storyboard scenes & 263 \\
    Manual-feedback records & 17 \\
    Audience-feedback bundles & 40 \\
    Machine-evaluation records & 81 \\
    \midrule
    Indexed images & 5,999 \\
    Indexed videos & 2,910 \\
    Indexed audio files & 219 \\
    \bottomrule
  \end{tabular}
\end{wraptable}

\subsubsection{Mock Web Database}
A defect can remain invisible when Meta relies only on its own scores.
The Internet database provides external production knowledge against
which to compare a workflow or its output. We use a mock web database to emulate the Internet, thereby preventing ground-truth videos for test cases from leaking. It is stored separately from
task cases. Meta first searches typed records and cached summaries, then
reads the original source to decide whether its method or example
applies. Meilisearch indexes titles and content for this retrieval.\footnote{\fontsize{8}{9}
\url{https://github.com/meilisearch/meilisearch}.}
Tab.~\ref{tab:lit-reference-inventory} reports the reference inventory.

\paragraph{Documents and Blogs.}
Documents support the step from an observed problem to an explanation
or candidate technique. Collection queries combine a production problem
with a method or tool, such as multi-shot continuity with reference
conditioning. Meta uses technical papers and model documentation to
understand how a technique works, then consults creator tutorials and
production discussions to determine how to apply it.

Relevant pages retain their source links and associated media so Meta can return from
a search summary to the supporting material. During evolution, the
investigation target determines the query, and the underlying source
determines whether the proposed method fits the workflow.

\paragraph{Multi-media.}
Media references make a quality claim testable. Film excerpts provide
examples of established production conventions, while AI-video
showcases and making-of accounts connect a technique to generated
footage and its reported failures. Audience comments help locate events
that viewers found problematic. Collection therefore retains the
associated clip or time interval and the context needed to interpret
the observation.

An account of a technique can inform a solution proposal. If the source
also identifies a localized failure, Meta can use the corresponding
media as a candidate test example and label the target property from
what is observable. Keeping machine-generated scores under a separate
source type preserves the distinction between an observation and a
model's judgment of it.

\paragraph{Skills, Storyboards and Code Bases.}
Once a candidate technique is identified, Meta needs enough procedural
detail to apply it. Capability-directed searches of repository hosts and
workflow-sharing communities retrieve the relevant skill, workflow, or
source implementation. For example, a camera-control investigation can
retrieve a workflow that implements the required movement.
Screenplay--storyboard pairs provide the corresponding
reference for directing decisions. PROSE scenes are indexed in contiguous
order,\footnote{\fontsize{8}{9}\url{https://github.com/creDreams/PROSE}.}
so a retrieved shot remains connected to the story progression it serves.

\subsubsection{Closed-loop Controller}
The Controller carries learning decisions into execution. Meta adapts
database cases to the behavior it wants to investigate and assigns the
briefs to User instances. The Controller launches their interactions
with the selected Director, then returns the records to Meta. This
separates the choice of what to learn from the mechanics of running
the project.

Starting from $D_0$, four evolution epochs produce the Director versions
$D_1,\ldots,D_4$. The evaluator is updated after each of the first three
epochs, and each update takes effect in the following epoch. Thus the
four epochs run with $(D_0,G_0)$, $(D_1,G_1)$, $(D_2,G_2)$, and
$(D_3,G_3)$, respectively, with $D_4$ as the final Director.
Throughout epoch $k+1$, $G_k$ evaluates both production results and
local Director revisions. Deferring evaluator changes to the next epoch
ensures that the old and revised Director methods are compared under
the same checking rule.

Meta can develop independent changes at their affected stage boundaries
within an epoch. Each proposal retains the source problem and edited
skill together with its local comparison. The Controller publishes the
selected role bundles for the next epoch, without rerunning the entire
task batch after each edit or imposing an aggregate-score improvement
gate. The retained histories of Director, User, and Meta bundles include
a separate evaluator version within Meta's method. Shared large
assets remain in global storage, allowing method versions to refer to
them without making another copy.

The fixed-evaluation branch remains executable with its original
workflow and costs, without mandatory additional exploration. This is
the retained set of baseline paths used by the capability-upper-bound
argument in Sec.~\ref{app:formal-capability}.

\subsection{Director Agent}
The Director must turn a partially disclosed request into connected
production decisions. The workflow builder plans how to carry out the
project using a storyboard that connects each decision to its expected
output and the later work that depends on it.

\subsubsection{Workflow Builder}
The builder starts from the disclosed request and selects applicable
methods from the skill library. Its default dependency order is plot
development, shot planning, asset preparation, segment generation, and
assembly. Each stage receives the available upstream work and must
return the artifact required by the next stage. The builder also
identifies where a user decision can change that dependency chain.

Interaction can then revise the project at the responsible point.
For example, a clarification that changes a plot event requires reconsidering its
shots, whereas a failed generation under an accepted shot plan calls
for a local prompt, reference, or generation repair. The persistent
directing session retains the earlier work while the builder decides
which dependent decisions to revisit. Thus the workflow evolves with
the disclosed intent without discarding unaffected production.

\subsubsection{Storyboard Builder}
To carry a directing decision through production, the storyboard links
the intended event to the media that realizes it:
\begin{center}
\setlength{\fboxsep}{6pt}%
\colorbox{gray!15}{%
  $\displaystyle
  \text{plot}\longrightarrow\text{shot}
  \longrightarrow\text{asset}\longleftrightarrow\text{segment}.
  $%
}
\end{center}
Each plot node specifies a subject's action and the resulting scene
change, which one or more shot nodes translate into visible events.
The shots refer to the assets needed for generation, and their resulting
audiovisual takes are stored in segment nodes. Assembly then orders the
accepted segments into the delivery.

These links preserve continuity as production proceeds. Several shots
can share an asset, so changing that asset identifies the segments that
need reconsideration. In the reverse direction, an accepted segment can
supply a reusable observation as an asset for a later shot. For example,
an extracted frame carries an observed appearance or state forward instead
of reconstructing it from text alone. Meta also uses these dependencies
to isolate a local revision. Holding the upstream inputs fixed allows
it to compare the artifacts produced by the old and revised procedure
at the affected boundary.

\subsection{User Agent}
\subsubsection{Persona Faithfulness}
The User makes interaction necessary by holding the private script and
persona while the Director receives only the public request. Its task
is to reveal intent as that client would, so a question changes the
Director's information only through the permitted disclosure behavior.
Invented facts would change the task, whereas inappropriate refusals
would prevent the Director from obtaining information the persona
should disclose. The following skill maintains
this distinction when answering questions and reviewing intermediate work.

\begin{tcolorbox}[litSkillBox]
\begin{lstlisting}[style=litSkill]
---
name: simulated-user
description: Respond as the client specified by the private task.
---
# Simulated User
## Goal
Preserve the client's intent and persona throughout the interaction.
## Required Inputs
1. **Private task**: `oracle` and the persona's disclosure policy.
2. **Public context**: initial request, dialogue, and question budget.
## Core Workflow
1. **Open the task**: send the initial request unchanged via `pair_message`.
2. **Answer selectively**: disclose the facts needed for this question.
   - Correct false premises. Leave unspecified creative choices open.
   - Preserve disclosed facts and the assigned persona.
3. **Review the delivery**: give feedback against the client's intent.
   - Distinguish a requested revision from an original requirement.
...
## Final Response
Send natural-language feedback. Submit the accepted result to the scheduler.
\end{lstlisting}
\end{tcolorbox}

\subsubsection{Directing Project Scheduling}
\paragraph{Task Difficulty Scheduling.}
The next project should expose behavior for which Meta needs evidence.
Meta therefore starts from a criterion or recently revised skill and
selects a case that activates it. It adjusts shot and character counts
to set the production complexity, and chooses a persona whose disclosure
behavior creates the intended interaction challenge. Depending on the
evolution stage, the adapted case revisits a known weakness or exercises
a newly available method.

\paragraph{Defect and Skill Injection.}
Meta embeds the target in the task requirements so that completing the
project calls for the intended behavior. For example, a scene requiring
continuity across shots can depend on a private fact that the Director
must clarify under the persona's disclosure policy.
Meta preserves that opportunity in the public
request and assigns the adapted private case to the User. The Controller
executes the interaction, allowing Meta to observe whether the target
behavior actually occurs.

\begin{tcolorbox}[litSkillBox]
\begin{lstlisting}[style=litSkill]
---
name: schedule-directing-project
description: Adapt a task to exercise a target criterion or directing skill.
---
# Schedule a Directing Project
## Goal
Obtain execution evidence for a weak criterion or a revised skill.
## Required Inputs
1. **Target**: findings from `skill-optimizer` or `evaluator-optimizer`.
2. **Task source**: database case, persona, and evolution history.
## Core Workflow
1. **Select the target**: identify the condition that activates it.
2. **Adapt the case**: adjust plot, shot count, characters, and persona.
   - Keep the private task coherent and specify the expected behavior.
   - Preserve the intended clarification opportunity in the public request.
3. **Assign the task**: pass the case to `simulated-user` via the Controller.
...
## Final Response
Return the adapted case, source case, and behavior to observe.
\end{lstlisting}
\end{tcolorbox}

\subsection{Meta Agent}
Meta must decide both what to improve and whether a proposed change
actually helps. Its result evaluation supplies defects visible under
the active criteria. Director revision also searches external references
for improvements those criteria may miss. User revision addresses the
response behavior that determines what the Director can learn, while
self revision changes Meta's ability to identify and investigate defects.

\subsubsection{Local Development and Acceptance}

\begin{table*}[t]
\centering
\vspace{0pt}
\centering
\setlength{\tabcolsep}{3pt}
\caption{Design parameters for local development and validation.}
\label{tab:lit-local-design}
\scriptsize
\begin{tabular}{p{0.19\linewidth}p{0.48\linewidth}p{0.23\linewidth}}
\hline
Parameter & Categories & Consequence\\
\hline
Difficulty $d$ & 
\makecell[tl]{
1: direct deterministic check\\
2: perceptual decision within one stage\\
3: context-dependent or cross-stage judgment
}
& $8d$ development and $4d$ validation cases\\
\hline
Importance $h$ & 
\makecell[tl]{
1: local optional quality\\
2: recurring quality requirement\\
3: task-critical failure}
& $\alpha_h=0.10,\ 0.05,\ 0.01$\\
\hline
Scope $s$ & Number of distinct affected input--output boundaries
whose improvement is claimed
& Separate tests with $\alpha=\alpha_h/s$\\
\hline
\end{tabular}
\end{table*}

\begin{wraptable}{r}{0.30\textwidth}
\centering
\setlength{\tabcolsep}{5pt}
\caption{Required independent local test cases for 80\% power at a single
boundary. Columns vary the confidence level. $u$ and $v$ are planning
probabilities, not measured results.}
\label{tab:lit-local-sample-size}
\scriptsize
\begin{tabular}{r|r|rrr}
\hline
$u$ & $v$ & 90\% & 95\% & 99\%\\
\hline
0.50 & 0.00 & 10 & 12 & 17\\
0.30 & 0.05 & 29 & 37 & 56\\
0.15 & 0.05 & 104 & 134 & 208\\
\hline
\end{tabular}
\end{wraptable}

Across these routes, Meta compares the old and revised method at the
input--output boundary affected by the proposal. Before editing, Meta
defines the expected outcome and fixes how correctness will be judged.
The sample-construction library described in
Sec.~\ref{app:lit-sample-construction} supplies the corresponding inputs
and counterexamples. Its construction and labeling rules belong to
Meta's research methods and serve all three revision routes.

The test budget follows the claim being made.
Difficulty determines the breadth of development examples, importance
sets the statistical confidence required for acceptance, and scope
identifies the boundaries at which improvement is claimed.
Meta selects and justifies these categories before development using
Tab.~\ref{tab:lit-local-design}. This confines repeated experimentation
to the affected behavior while retaining an explicit comparison with
the method it would replace.

A candidate needs different evidence for development and acceptance.
Development cases expose the mechanism being repaired, and validation
cases select among the resulting alternatives. Once a candidate is
frozen, source-disjoint held-out cases test whether it improves on the
old method. Including both failures to correct and successful behavior
to preserve makes regressions part of the same comparison. Development
and validation counts set the budget for finding a candidate. The held-out
count is instead chosen to detect the anticipated improvement with the
statistical power specified below.

Before candidate development, Meta defines what constitutes a correct
output for the role and behavior being revised, as detailed below.
Both methods are then judged by this binary rule on independent test
sources drawn from the specified local task distribution.
Let $b$ count cases where the old method is wrong and the new method is
correct, and let $c$ count cases where a previously correct output becomes
incorrect. Their population probabilities are
$p_{01}$ and $p_{10}$, so
\begin{equation}
\label{eq:lit-local-accuracy-gain}
A_{\mathrm{new}}-A_{\mathrm{old}}=p_{01}-p_{10}.
\end{equation}
For $m=b+c$, a one-sided exact paired test gives
\begin{equation}
\label{eq:lit-local-paired-test}
p_{\mathrm{test}}(b,c)
=2^{-m}\sum_{r=b}^{m}\binom{m}{r},
\qquad p_{\mathrm{test}}(0,0)=1.
\end{equation}
Under no improvement, the conditional probability that the new method
wins a disagreement is at most $1/2$. The binomial upper tail therefore
controls a false declaration of improvement at level $\alpha$ when
$p_{\mathrm{test}}\le\alpha$.\footnote{\fontsize{8}{9}Exact binomial testing:
\url{https://docs.scipy.org/doc/scipy/reference/generated/scipy.stats.binomtest.html}.}
Acceptance therefore concerns the behavior changed at this boundary.
It supplies evidence for incorporating that local method without
requiring the entire pipeline's aggregate score to improve.

To choose the held-out sample size, let $u$ and $v$ be anticipated
correction and regression probabilities estimated during development.
For a fixed $n$, its detection probability under that planning alternative
is
\begin{equation}
\label{eq:lit-local-power}
\begin{aligned}
\mathcal P_n(u,v;\alpha)
={}&\sum_{b=0}^{n}\sum_{c=0}^{n-b}
\mathbf 1\{p_{\mathrm{test}}(b,c)\le\alpha\}\\
&\times\frac{n!}{b!\,c!\,(n-b-c)!}
u^b v^c(1-u-v)^{n-b-c},\\
n_{\mathrm{test}}
={}&\min\{n:\mathcal P_n(u,v;\alpha)\ge0.80\}.
\end{aligned}
\end{equation}
The sample size gives at least an 80\% probability of detecting the
improvement anticipated from the pilot estimates, using a test at
confidence level $1-\alpha$. It is fixed before inspecting any held-out
outcomes, which alone determine the final acceptance decision.

For a perceptual repair of recurring importance at one boundary,
$d=2$, $h=2$, and $s=1$ allocate 16 development cases and 8 validation
cases. If the pilot suggests correction probability $u=0.30$ and
regression probability $v=0.05$, the held-out comparison uses 37 cases
for at least 80\% power at $\alpha=0.05$. These are inputs to the affected
stage, such as shot descriptions or short clips, rather than complete
productions. Smaller anticipated effects require more cases, as shown
in Tab.~\ref{tab:lit-local-sample-size}.

On that fixed test, eight corrections and one regression give
$p_{\mathrm{test}}=10/512\approx0.0195$, supporting acceptance.
Five corrections with no regressions also satisfy the 95\% confidence
criterion, regardless of the additional ties. If the candidate fails
the held-out comparison, Meta returns it to development and uses fresh
held-out evidence for its next assessment. The confidence statement
belongs to each prespecified local comparison, with the scope
adjustment in Tab.~\ref{tab:lit-local-design}.

\subsubsection{Result Evaluation}
\paragraph{Initial Evaluation Criteria.}
The initial criteria give Meta a starting objective for identifying
production defects, as formalized in Sec.~\ref{app:formal-target}.
Each judgment relates an observed result to the applicable requirement
and retains the evidence needed to investigate a discrepancy.
This evidence lets Meta revisit a judgment when it revises either the
criterion or the checking procedure.

Before applying a check, the evaluator reconstructs the production
timeline from media metadata and detected shots. It aligns the full-audio
transcript with an overview grid sampled every 2 seconds and a grid of
shot-first frames, then consults the dialogue and execution records to
locate the relevant production decision.

For example, to investigate whether a generated shot realizes its
disclosed plan, the evaluator compares three frames from the shot with
that plan. If the discrepancy concerns its transition to the adjacent
shot, the cut-continuity procedure inspects 1.5 seconds on either side
of the cut at 4 fps. A suspected discrepancy in the shot's soundtrack
is examined with the audio-understanding tool. These observations
localize the failure so that Meta can trace it to the responsible decision.

Attribution then requires establishing what the Director understood
when making that decision. The evaluator forks the frozen production
session, asks it to summarize the intended result from the information
it received, and compares the summary with the private specification.
The timeline establishes whether the relevant information arrived before
the decision, while execution records establish whether it was used.
For example, a reference-dependent failure requires checking whether
the reference was actually passed to generation. The resulting assessment
links the observed defect to the earliest failure supported by the trace.

\subsubsection{Director Revision}
Director revision changes the method that turns a disclosed requirement
into a production artifact. A revision can begin with an evaluated
failure or an externally discovered technique. In either case, Meta
must identify the production decision to change before comparing the
candidate with the current method under the epoch's active evaluator.

\paragraph{Reactive Defect Attribution.}
Starting from a reported defect, Meta traces the artifact backward
through the inputs available to each producing step. The relevant
failure is the first point where available requirements cease to be
reflected in the output. Distinguishing a fact never disclosed from one
received but ignored determines whether the repair belongs to
interaction or production. The local test in the following skill then
checks the proposed explanation with upstream inputs held fixed before
the repair is generalized to the workflow builder.

\begin{tcolorbox}[litSkillBox]
\begin{lstlisting}[style=litSkill]
---
name: attribute-directing-defect
description: Trace a reported failure to its responsible directing rule.
---
# Attribute a Directing Defect
## Goal
Localize the repair before updating the reusable directing method.
## Required Inputs
1. **Method**: task-specific `video-skill` and `video-skill-creator`.
2. **Evidence**: disclosed requirements, execution trace, and evaluation.
## Core Workflow
**Disclosure boundary**: the Director receives only disclosed information.
1. **Trace the failure**: locate the first failed input--output boundary.
   - Distinguish undisclosed requirements from ignored disclosed facts.
2. **Test the cause**: change the suspect decision with upstream inputs fixed.
3. **Backpropagate the repair**: follow `skill-optimizer`.
   - Repair `video-skill` first. Generalize through `video-skill-creator`.
...
## Domain Reference
Read `video-skill-creator/references/video_generation_expertise.md`.
## Final Response
Report the failed boundary, causal evidence, and local repair test.
\end{lstlisting}
\end{tcolorbox}

\paragraph{Proactive Defect and Skill Exploration.}
To discover improvements beyond the evaluator's coverage, Meta selects
a production function and compares its method with one to three external
sources. It then constructs a stage-level input on which the reference
behavior or alternative technique would make an observable difference.
Executing the current method on that input reveals whether the proposed
difference identifies a defect to repair or a promising method to develop.
The resulting candidate joins the same proposal process as a reactively
diagnosed failure.

\begin{tcolorbox}[litSkillBox]
\begin{lstlisting}[style=litSkill]
---
name: explore-directing-method
description: Discover directing defects and techniques from external evidence.
---
# Explore a Directing Method
## Goal
Investigate improvements beyond failures reported by the evaluator.
## Required Inputs
1. **Method**: `video-skill` and `video-skill-creator`.
2. **Search target**: production function and relevant task evidence.
## Core Workflow
1. **Retrieve references**: use `reference_search` and `reference_read`.
   - Compare one to three sources with the target workflow.
2. **Construct a test**: use `evaluator-optimizer/references/sample-construction.md`.
   - Keep reference answers outside the executor's input.
3. **Investigate**: run the target stage and compare against the reference.
   - Pass supported defects or techniques to `skill-optimizer`.
...
## Domain Reference
Read `video-skill-creator/references/video_generation_expertise.md`.
## Final Response
Report the tested hypothesis, sources, and applicable improvement.
\end{lstlisting}
\end{tcolorbox}

\paragraph{Solution Proposal and Validation.}
Meta turns a diagnosed failure into a proposal by identifying the
responsible decision and retrieving a solution that can act there.
The proposal specifies which instruction or operation will change and
what different output that change should produce. Testing alternatives
on the same upstream inputs lets Meta compare that expected effect
rather than unrelated differences between projects.

For this route, correctness means satisfying the target production
requirement under the active checking rule. The comparison is performed
at the boundary where the proposed change is expected to affect the
output. For example, a shot repair is tested on the generated shot,
extending the comparison to assembly only when the claimed effect
depends on assembly.
The shared development budgets and paired acceptance rule above apply
at that selected boundary.

An accepted task-specific repair must then become a method the builder
can reuse. Meta first modifies the executable task workflow and transfers
the general rule to its creator and library entry. The following skill
uses validation accuracy to choose the candidate, resolving ties by
tool cost and then edit size. It merges overlapping rules through
\texttt{pair-wise-skill-merge} and repeats testing if the transfer changes
the tested behavior. In this way, local evidence determines both the adopted
repair and the conditions under which future projects should select it.

\begin{tcolorbox}[litSkillBox]
\begin{lstlisting}[style=litSkill]
---
name: develop-local-revision
description: Develop and accept a change at its affected workflow boundary.
---
# Develop a Local Revision
## Goal
Accept a local improvement and transfer its reusable rule.
## Required Inputs
1. **Proposal**: diagnosis and candidate solution from `skill-optimizer`.
2. **Test contract**: affected boundary, correctness rule, and (d, h, s).
## Core Workflow
1. **Prepare fixtures**: use `evaluator-optimizer/references/sample-construction.md`.
   - Split sources before augmentation. Include previously correct cases.
2. **Develop candidates**: revise the target skill on development fixtures.
3. **Select and freeze**: choose by validation accuracy, then cost and edit size.
   - Freeze the candidate, judging rule, and held-out sample count.
4. **Accept locally**: apply the paired test to old/new held-out outcomes.
   - Use the epoch's evaluator for Director changes. Retest on fresh sources.
5. **Integrate**: follow `skill-optimizer` to update the task and creator rules.
   - Use `pair-wise-skill-merge` for overlapping changes. Retest changed behavior.
...
## Final Response
Return the skill change, applicability, and acceptance evidence.
\end{lstlisting}
\end{tcolorbox}

\subsubsection{User Revision}
A User method shapes the evidence available to the Director. Its revision
therefore aims to preserve the assigned client's behavior while making
that behavior more faithful to the private task. The comparison operates
on a reply to a fixed conversation, rather than on a generated video.

\paragraph{Dialogue Analysis.}
Meta starts from the Director's actual question and asks how this persona
should respond under the private script. This determines whether a
missing answer was a faithful refusal or a response defect. A correct
reply must satisfy the expected response specification in both the facts
it discloses and the behavior it adopts.

A dialogue fixture fixes the conversation prefix and the Director
message, so old and revised User instructions face the same request.
Contrasting fixtures change the question or persona while retaining
the underlying fact. For example, the same private fact may require
an answer under one disclosure policy and a refusal under another.
The comparison therefore tests fidelity to the assigned persona rather
than rewarding additional disclosure.

\paragraph{Persona Faithfulness Enhancement.}
The diagnosed response defect identifies the instruction to revise in
\texttt{simulated-user}. Meta develops that instruction on the contrasting
contexts, then applies the shared local comparison to held-out contexts.
The fixed response specification supplies the correctness label.
The skill below makes the accepted change reusable while preserving
the private script and persona as the authority for each reply.

\begin{tcolorbox}[litSkillBox]
\begin{lstlisting}[style=litSkill]
---
name: refine-user-faithfulness
description: Repair replies that conflict with the assigned client.
---
# Refine User Faithfulness
## Goal
Improve persona and script fidelity in `simulated-user`.
## Required Inputs
1. **Private specification**: script, persona, and disclosure policy.
2. **Dialogue fixture**: fixed conversation prefix and Director message.
## Core Workflow
1. **Specify the response**: determine the permitted facts and response behavior.
2. **Develop the rule**: contrast answer, refusal, clarification, and correction.
   - Revise `simulated-user` using `develop-local-revision`.
3. **Assess fidelity**: compare old/new replies on held-out dialogue contexts.
   - A correct reply satisfies both disclosure and persona requirements.
...
## Final Response
Return the instruction change and response-level comparison.
\end{lstlisting}
\end{tcolorbox}

\subsubsection{Self Revision}
Meta's own evaluation determines which defects receive attention.
Self revision therefore investigates whether a score reflects actual
quality or a weakness in the check, and whether external examples reveal
a property absent from the criteria. The resulting distinction is
tested on labeled samples before it guides Director development.
The sample-construction methods developed for this purpose also provide
the labeled examples and stage-level fixtures used across revision routes.

\paragraph{Evaluation History Analysis.}
Meta examines the scores of cases to which a criterion applies, scheduling
targeted tasks if too few such cases are available. Scores concentrated
at an endpoint or showing little variation motivate further investigation
because they can reflect either a weak check or consistently good or bad
production.

For normalized scores $s_i\in[0,1]$, define endpoint
concentration $E$ and score spread $V$ by
\begin{equation}
\label{eq:lit-score-audit}
E=\max\left\{\frac1N\sum_i\mathbf1[s_i\le0.1],
             \frac1N\sum_i\mathbf1[s_i\ge0.9]\right\},
\qquad
V=\sqrt{\frac1N\sum_i(s_i-\bar s)^2}.
\end{equation}
The audit design uses $N\ge16$
applicable cases and either $E\ge0.9$ or $V\le0.05$ to trigger a
discriminating sample test. Labeled positive and negative examples then
separate measurement failure from production quality. If defective
examples pass, Meta strengthens the check or adds the missing criterion.
If valid examples fail, it revises the decision rule responsible for
those rejections.

If the check distinguishes both classes, production scores determine
the next investment. Persistently low scores direct repair toward the
Director. Persistently high scores support treating the property as
mastered on the observed tasks and developing a cheaper checking
procedure. The thresholds thus allocate investigation, while the sample
comparison determines which method should change.

\paragraph{Criteria Exploration.}
History can expose a weak existing check, but an omitted property
requires another source of evidence. Meta searches external production
references for defects and assessment methods relevant to the task,
then compares them with the coverage of its criteria. For a supported
gap, it defines applicability and the distinction between a passing and
failing example before developing the inspection procedure. The following
skill connects that discovery to labeled examples and the shared local
development loop.

\begin{tcolorbox}[litSkillBox]
\begin{lstlisting}[style=litSkill]
---
name: refine-evaluation-criterion
description: Repair a weak check or develop a missing quality criterion.
---
# Refine an Evaluation Criterion
## Goal
Improve defect discrimination within `evaluator-optimizer`.
## Required Inputs
1. **Criterion**: applicability, checking procedure, and score history.
2. **References**: external methods and labeled examples.
## Core Workflow
1. **Find the gap**: inspect extreme scores and search external references.
   - Distinguish missed properties, weak checks, and untriggered criteria.
2. **Build examples**: follow `evaluator-optimizer/references/sample-construction.md`.
3. **Develop the check**: use `develop-local-revision` at the criterion boundary.
   - Compare corrected and newly introduced classification errors.
4. **Allocate effort**: simplify reliable checks while retaining their criteria.
...
## Final Response
Return the criterion, checking procedure, held-out results, and cost.
\end{lstlisting}
\end{tcolorbox}

\paragraph{Construction of Positive and Negative Samples.}
\label{app:lit-sample-construction}
To distinguish a better check from a more permissive one, Meta needs
examples whose target property is established independently of the
candidate check. The construction library maps that property to an
extraction or transformation method. Tab.~\ref{tab:lit-sample-methods}
gives six worked construction examples, specifying the reference and
contrasting cases for each target boundary.

Each revision route uses these constructions to hold its test inputs
fixed. An evaluator revision is compared on the same artifacts, so the
observed difference concerns classification. A Director revision instead
receives the same stage inputs, allowing Meta to compare the behavior
produced by the two workflows. For User revisions, dialogue fixtures
provide the fixed request against which each reply is judged. The
construction methods are shared, but the required output follows from
the role being improved.

\begin{table*}[t]
\centering
\caption{Examples of positive and contrasting samples for local development.}
\label{tab:lit-sample-methods}
\scriptsize
\begin{tabular}{@{}p{0.18\linewidth}p{0.34\linewidth}p{0.40\linewidth}@{}}
\toprule
\textbf{Target boundary} & \textbf{Positive/reference construction} & \textbf{Negative or contrasting construction}\\
\midrule
Plot to shot plan & Extract a contiguous screenplay--storyboard
sequence. Annotate the event each shot makes visible.
& Remove the shot needed to establish a specified event, or change its
depicted action so that it fails the fixed event requirement.\\\midrule
Reference selection & Build a shot fixture with the required subject
and scene references. Annotate which generation input must use each.
& Omit or replace the required reference in the generation arguments.
Retain the rest of the fixture.\\\midrule
Character or scene continuity & Extract adjacent shots with consistent
identity and spatial relations. Annotate the relation being tested.
& Regenerate a short shot with an omitted scene reference or an
altered character/layout reference. Label the observed violation.\\\midrule
Audio synchronization & Extract a clip with aligned speech and image.
Record the aligned interval.
& Separate the voice from background audio, shift the voice, and remix
it with the unchanged background.\\\midrule
Sound continuity & Extract a continuous scene with a coherent
soundscape across its cuts.
& Join shots with independently generated background tracks to create
a localized soundscape discontinuity.\\\midrule
User response & Derive the permitted answer and response behavior
from a private script, persona, and fixed conversation prefix.
& Change the question or disclosure condition to require a different
response. Retain the underlying fact to isolate the behavioral distinction.\\
\bottomrule
\end{tabular}
\end{table*}

The annotation skill records the property and applicability condition,
then establishes its label from the relevant evidence. A media label
belongs to a specified interval or relation, not to the entire film.
When a construction changes the source, its record retains the
transformation and the resulting observation. A generated counterexample
is labeled from what actually appears, since the intended perturbation
may fail to produce a defect.

Augmentation either varies content irrelevant to the judgment or
deliberately changes the target property. This creates useful contrasts
for development, but variants of one source remain related. Sources are
therefore split into development, validation, and held-out pools before
transformation, and all derived variants stay in their source's pool.
One held-out source contributes one prespecified comparison to the
independent-case count.

The probability analysis in Eq.~\ref{eq:lit-local-power} assumes correct
labels and independent held-out sources representative of the specified
target distribution. That distribution also fixes the meaning of the
accuracy comparison. Balanced positive and negative sampling measures
balanced discrimination. To assess performance on the deployment mixture,
the samples must instead follow that mixture's class proportions. These construction choices determine what
the shared acceptance test can establish.

\paragraph{Criteria Validation and Refinement.}
A criterion revision uses one checking rule as its test boundary.
Correctness means agreement with the sample label, and the old and
revised procedures classify the same examples. A newly introduced
criterion is compared with the preceding behavior, which passed the
previously unchecked property.

An improvement can come from detecting a defect that the old check
missed or from accepting a valid example that it incorrectly rejected.
Conversely, a newly missed defect or false alarm counts as a regression. Thus $b$ and $c$ enter
Eq.~\ref{eq:lit-local-paired-test} as corrected and newly introduced
classification errors. Miss and false-alarm probabilities are also
estimated separately on their respective classes, so an aggregate
improvement retains its class-specific interpretation.

Once a criterion is reliable, Meta can develop a less expensive way to
apply it while keeping the requirement active. For example, a detailed
frame-by-frame coordinate comparison can become a brief position
continuity inspection during frame sampling. That revision concerns
the cost of obtaining the judgment and is compared at the same criterion
boundary.

These local revisions connect the editable criteria $G_k$ to the
Director method $\theta_k$ in Sec.~\ref{app:formal-process}. When external
examples reveal a defect overlooked by the old evaluation, Meta develops
a criterion that can detect it. This criterion then guides Director
development toward a repair. At each changed boundary, the paired
comparison assesses whether the corrections outweigh newly introduced
errors. The resulting sequence implements the objective in
Sec.~\ref{app:formal-capability} by using better defect discrimination
to guide the Director toward its attainable capability bound.
Publication uses the resulting local evidence without an aggregate
pipeline-score gate.

%% file: sections/A4_dataset_construction.tex
\section{Dataset Construction}
\label{app:dataset}

Each case couples a video-creation task with a simulated user: the task
defines the creative context, while the persona determines how the user
expresses intent, answers questions, and delegates production decisions.
This section describes the case schema, construction procedure, and
persona-conditioned interaction examples.

\subsection{Data Structure}
\label{app:data-structure}
Each case separates the director's public input from the simulated user's
private context and evaluator-only annotations. Tab.~\ref{tab:case-fields}
summarizes these fields and their roles in the interaction task.

\begin{table}[t]
\centering

\caption{Public input, private user context, and evaluation records in a
dataset case. The director receives the opening request, references, and
interaction settings. The simulated user's private context contains the task,
persona, and disclosure rules. Separate fields record evaluation requirements
and construction history. Private fields are not passed directly to the director.}
\label{tab:case-fields}
\scriptsize
\setlength{\tabcolsep}{6pt}
\renewcommand{\arraystretch}{1.12}
\begin{tabular}{@{}>{\raggedright\arraybackslash}p{0.46\linewidth}>{\raggedright\arraybackslash}p{\dimexpr0.54\linewidth-12pt\relax}@{}}
\toprule
\textbf{Field (JSON path)} & \textbf{Content} \\
\midrule
\rowcolor{ProcessBlue!7}
\multicolumn{2}{@{}l}{\textcolor{ProcessBlue!75!black}{\textbf{Public input \& interaction protocol}}} \\
\textit{benchmark\_case.reference\_materials} & Reference assets and descriptions, when available. \\
\textit{benchmark\_case.interaction} & Clarification budget and interaction mode. \\
\midrule
\rowcolor{RedOrange!6}
\multicolumn{2}{@{}l}{\textcolor{RedOrange!75!black}{\textbf{Private task \& persona}}} \\
\textit{oracle.script} & Story, characters, shot plan, and audiovisual design. \\
\textit{user\_profile.description} & Identity, knowledge, style, preferences, and patience. \\
\textit{user\_profile.process\_expertise} & Method-level expertise and explicit workflow requirements. \\
\textit{user\_profile.disclosure\_plan} & What to state, reveal on request, withhold, or leave undecided. \\
\midrule
\rowcolor{ForestGreen!6}
\multicolumn{2}{@{}l}{\textcolor{ForestGreen!75!black}{\textbf{Evaluation}}} \\
\textit{evaluation\_points} & Additional requirements for this case during evaluation. \\
\bottomrule
\end{tabular}
\end{table}

These fields serve three complementary roles.

\paragraph{Public brief and interaction protocol.}
The \texttt{benchmark\_case} records the initial request, available
references, and interaction settings, including the clarification budget.
The opening request can range from an idea seed to a shot specification.
Reference entries identify assets and their intended content. They are
not counts of distinct source videos.

\paragraph{Private task and user context.}
The \texttt{oracle} organizes the story, characters, and production design.
The \texttt{user\_profile} specifies identity, knowledge boundaries,
communication style, preferences, and material readiness. Its disclosure
plan distinguishes what is volunteered, revealed when asked, withheld, or
unspecified. The director obtains task information through the public brief
and actual user responses, not by reading these private records.

\paragraph{Assessment and traceability.}
Evaluation points, where present, record case-specific output requirements.
Behavioral test points describe expected interaction behavior. Provenance and generation
records track the source material, transformations, and review history.
These annotations are kept separate from the director's input.

\paragraph{Two interaction goals.}
In \emph{specification recovery}, the director elicits requirements already
known to the user. In \emph{co-development}, the user may need alternatives
before making a decision. The source oracle is a reference, not a hidden
mandatory answer. The intent state distinguishes these goals, so an
unspecified creative choice need not be treated as a missed requirement.

\subsection{Data Sources}
The 240-case collection comprises 178 FilmBench-derived
cases~\citep{filmbench2026}, 46 MovieChat-1K-derived
cases~\citep{song2024moviechat}, four cases from supplied plot material,
and 12 authored OOD cases. Source descriptions are organized into story
and production fields, then paired with personas and disclosure policies.
The MovieChat-derived cases include 46 local source clips, ranging from
30.16 to 119.66 seconds (mean: 67.8 seconds). These durations describe source
material, not generated outputs or target durations for the full collection.

\subsection{Construction and Verification}
As shown in Fig.~\ref{fig:data}, construction first organizes the source
into a script and design, then assigns a persona and an initial disclosure
state. The opening request must agree with both the task and the persona's
knowledge: a user can specify a desired result without knowing how to
produce it.

The persona-consistent verification loop uses ordinary clarification
questions and adversarial probes, including false premises and requests
for protected annotations. Reviewers check factual consistency, recoverability
within the question budget, persona fidelity, and disclosure boundaries.
Feedback guides revision of the profile and disclosure policy. This is an
interaction-level check, not a test of video-generation feasibility.
The export records passing reviews for 228 cases and no such review for
the 12 authored OOD cases. These records apply to the reviewed versions.
Later edits require renewed validation. Construction probes are not
held-out robustness tests.

\subsection{Data Distribution}
\label{app:data-distribution}
The dataset contains 240 tasks, 284 reference-material entries, and 892
planned shots. Initial development, evaluator development, evolution
epochs 1--4, and test each contain 32 cases. OOD contains 16.
The OOD cases comprise ten outlines and six idea seeds. This is a task-level
partition: source-video identifiers can recur across splits.

\begin{table}[ht]
\centering
\small
\caption{Distributions of intent maturity and reference availability across 240 tasks.}
\label{tab:dataset-distributions}
\setlength{\tabcolsep}{9pt}
\renewcommand{\arraystretch}{1.15}
\begin{tabular}{@{}lr@{\hspace{28pt}}lr@{}}
\toprule
\textbf{Intent maturity} & \textbf{Tasks} & \textbf{Reference assets} & \textbf{Tasks} \\
\midrule
Idea seed & 17 & None & 22 \\
Story outline & 147 & One & 168 \\
Supplied script & 67 & Two & 39 \\
Specified shots & 9 & Three or more & 11 \\
\midrule
Total & 240 & Total & 240 \\
\bottomrule
\end{tabular}
\end{table}

The collection is diverse but not uniformly balanced (Tab.~\ref{tab:dataset-distributions}). Four-shot plans
account for 162 tasks. The \texttt{professional\_user} flag is true for
11 profiles and false for 229. It marks explicit method-aware behavior
and does not classify users solely by occupation. Planned shots describe
the task specification, not detected shots or measured video duration.

\subsection{Persona Settings and Response Examples}
\label{app:persona-examples}
Personas vary along identity, production knowledge, communication style,
preferences, and patience. We distinguish \emph{intent maturity} (how
developed the request is) from \emph{process expertise} (which production
decisions the user can specify). A non-professional user may have a complete
story but delegate the workflow. A professional user may specify a method
while leaving other choices to the director. Both must acknowledge unknowns
and avoid inventing new requirements.
Fig.~\ref{fig:persona-responses} illustrates this distinction using a
shared creative brief and the same director question.

\begin{figure}[!htb]
\centering
\begin{minipage}{\linewidth}
\small
\textbf{Shared task.} Create a short dialogue scene: preserve the supplied
lines, assign them to the correct characters, and keep speech and on-screen
performance synchronized. The director asks the same production question
of two users with different knowledge boundaries.

\begin{tcolorbox}[litSkillBox,
  colback=ProcessBlue!3,colframe=ProcessBlue!35,
  borderline west={1.5pt}{0pt}{ProcessBlue},
  title={Example 1: Non-professional user},
  colbacktitle=ProcessBlue!10,coltitle=litSkillInk,fonttitle=\bfseries\small]
\begin{lstlisting}[style=litSkill,morekeywords={persona,knowledge,style,director,user}]
persona: A community librarian making a short film for friends.
knowledge: Can describe the desired scene but does not know how
           dialogue recording and shot timing are coordinated.
style: Everyday language, delegates technical decisions.

director: "How would you like us to handle the dialogue
           and shot timing?"
user: "I'm not sure how that part works. Please keep the lines
       as written and make sure each person says the right ones.
       I want their mouths and voices to match, without cutting
       off the end of a line. Please choose how to do it."
\end{lstlisting}
\end{tcolorbox}

\begin{tcolorbox}[litSkillBox,
  colback=ForestGreen!3,colframe=ForestGreen!35,
  borderline west={1.5pt}{0pt}{ForestGreen},
  title={Example 2: Professional user},
  colbacktitle=ForestGreen!10,coltitle=litSkillInk,fonttitle=\bfseries\small]
\begin{lstlisting}[style=litSkill,morekeywords={persona,knowledge,style,director,user}]
persona: An assistant editor at a small post-production studio.
knowledge: Understands shot planning and has an explicit
           audio-first dialogue requirement in the profile.
style: Concise production terminology, concrete instructions.

director: "How would you like us to handle the dialogue
           and shot timing?"
user: "Keep the lines unchanged. Separate them by speaker and
       finalize the dialogue audio first. Then time the shots
       and performances to those recordings so no line is cut
       off or assigned to the wrong character. I have no other
       tool or workflow requirements."
\end{lstlisting}
\end{tcolorbox}

\textbf{Implication for the director.} The first reply specifies an outcome
and delegates implementation. The second adds a supported workflow
constraint. The director should select the method in the first case and
preserve the audio-first requirement in the second, without assuming
either user can answer unspecified technical questions.
\end{minipage}
\caption{Illustrative responses from users with different levels of production expertise.}
\label{fig:persona-responses}
\end{figure}

\clearpage

%% file: sections/A5_evaluation_protocol.tex
\section{Evaluation Protocol}
\label{app:evaluation}

We evaluate production quality, interaction and intent alignment, and the
evolution of working methods. The first two evaluations comprise the
four dimensions in Sec.~\ref{sec:metrics}: Consistency, Dynamic,
Interaction, and Intent. They contain six reported aspects and twenty
scoring items. Process analysis examines which methods the agents can
use and whether they use them during production.

\textbf{Consistency} is divided into two aspects: \textit{visual} aspect evaluates scene layout, character appearance, and blocking, while \textit{audio} aspect assesses voice, sound effects, and audio-visual consistency.

\textbf{Dynamic} is evaluated at different granularities. Fine-grained \textit{motion} includes physical plausibility and motion coherence. Coarse-grained \textit{plot} includes plot coherence and storyboard design quality.

\textbf{Interaction} evaluates director's timely access to necessary information, question necessity, adoption of obtained information, question timing, and feedback provision.

\textbf{Intent} measures alignment between director system and user understanding, expected output achievement, known requirement achievement, reference fidelity, and case-specific scoring items.

Each scoring item is graded on a three-level scale: 0 (non-compliant), 0.5 (flawed), and 1 (compliant). Case-specific scoring evaluates the checkpoints attached to each project in the test set. The score is computed as the proportion of satisfied checkpoints, with a maximum total of 3 points. Each dimension’s score is ultimately scaled to 100.


\paragraph{Privileged user and evaluator.}
For test-time interaction, the privileged User receives the complete task specification, a user
persona, and rules for when to reveal information. It starts with the
specified initial request and provides further information during the
conversation according to those rules. Its replies remain consistent
with earlier disclosures, correct the Director's mistaken assumptions,
and leave unspecified creative choices to the Director. When the User
explicitly changes a requirement, the new requirement applies to
subsequent work.
The privileged evaluator receives the complete task, case checkpoints,
released references, dialogue, production records, and delivered media.
It determines what the Director knew when making each production
decision. Production consistency is scored against the script that the
Director finalized for production. Intent scores measure whether the
result meets the User's complete requirements, including those not
disclosed to the Director. We calibrate the privileged rubric on the
evaluator development set and use the same rubric for all compared
methods and evolution epochs. Meta can change its own evaluation
criteria but cannot change this rubric.

\paragraph{Evidence and independent judgments.}
Each case is evaluated in four independent Qwen3.8-27B sessions with
xhigh reasoning effort. Evaluators receive the same source materials and
score without reading one another's judgments. The evidence package
contains media metadata, timestamped transcripts, detected shots, an
overview grid sampled every two seconds, and the first, middle, and
last frames of each shot.
Evaluators inspect the two seconds before and after each cut at
0.5-second intervals and sample more densely when a suspected defect
lasts less than this interval. They inspect images directly and use
audio-understanding tools to examine the soundtrack. Transcripts
provide the spoken words and their timing. Every score deduction must
name the scoring item, describe the defect, and cite a frame, time
interval, or execution record. If required evidence cannot be obtained,
the evaluation is marked as failed and must be completed before its
scores enter the reported average.

\paragraph{Scoring and aggregation.}
Except for case-specific checkpoints, each item receives $1$ when its
requirements are met, $0.5$ for a minor defect, and $0$ for a clear
violation. For items scored by the fraction of successful events, we
convert that fraction $r$ into a score using
\begin{equation}
\label{eq:eval-discretization}
q(r)=
\begin{cases}
1,&r\ge0.75,\\
0.5,&0.25\le r<0.75,\\
0,&r<0.25.
\end{cases}
\end{equation}
If a production item contains several checks, its score is the lowest
score among the applicable checks. A defect that meets an explicit
zero-score rule receives $0$, even if another rule would allow $0.5$.
Case-specific checkpoints are scored by the fraction satisfied, without
converting that fraction into the three score levels.

We denote evaluator $r$'s score for item $j$ on case $n$ by $s_{nj}^{(r)}$.
We first average the four evaluations of each item. We then average
over all cases and items in aspect $g$ and multiply by 100:
\begin{equation}
\bar s_{nj}=\frac14\sum_{r=1}^{4}s_{nj}^{(r)},\qquad
S_g=\frac{100}{N|\mathcal I_g|}
       \sum_{n=1}^{N}\sum_{j\in\mathcal I_g}\bar s_{nj}.
\label{eq:eval-aggregation}
\end{equation}
Here $N$ is the number of cases, and $\mathcal I_g$ is the set of items
in aspect $g$. Visual and Audio each contain three items, Motion and
Plot each contain two, and Interaction and Intent each contain five.
Cases and items within an aspect have equal weight.
For the four dimensions reported in the main text, we use
$S_{\mathrm{Consistency}}=(S_{\mathrm{Visual}}+S_{\mathrm{Audio}})/2$
and $S_{\mathrm{Dynamic}}=(S_{\mathrm{Motion}}+S_{\mathrm{Plot}})/2$,
with Interaction and Intent unchanged. Scores are reported on a
0--100 scale. Values on a 0--1 scale are these scores divided by 100.
Ground-truth videos without an agent interaction receive only production
scores.

\subsection{Production Quality}
\label{app:eval-production}

\subsubsection{Visual Consistency}

\paragraph{Scene consistency.}
Scene boundaries are identified from the frozen script. Fixed objects
are compared across adjacent shots and against the scene's first clear view.
\par\smallskip
\begingroup
\scriptsize
\setlength{\tabcolsep}{4pt}
\renewcommand{\arraystretch}{1.15}
\noindent
\begin{tabular}{@{}p{0.22\linewidth}p{\dimexpr0.78\linewidth-2\tabcolsep\relax}@{}}
\toprule
\rowcolor{gray!20} \textbf{Check} & \textbf{Evaluation rule} \\
\midrule
Properties & The comparison covers fixed-object attributes (presence, color, shape, text,
material, count), architecture and furniture layout, spatial relations
(left/right, front/back, height, adjacency), and environmental conditions
(season, weather, time of day). \\
\hline
Grades & A scene change absent from the script, a missing scripted transition,
a changed major structure or layout, or changes to multiple fixed
objects receive $0$. A slight change to one minor decoration or texture
receives $0.5$. Stable scenes and valid transitions receive $1$. \\
\hline
Valid changes & A change in camera position is valid if the objects still
occupy consistent locations in the scene. A scene-wide color shift is
valid only if several fixed objects change consistently with the same
lighting or color-grading adjustment.
Content shown on a screen within the scene may change. \\
\hline
Evidence & The evaluation records which scene each shot belongs to,
where scenes change, and which objects were compared across shots.
SAM3 helps locate suspected changes. Conflicting detections are
resolved by inspecting the source footage. \\
\bottomrule
\end{tabular}
\par
\endgroup
\smallskip

\paragraph{Character consistency.}
Visible people are identified before being matched to scripted
characters. Each character is tracked through occlusions until an
explicit exit or the end of the scene.
\par\smallskip
\begingroup
\scriptsize
\setlength{\tabcolsep}{4pt}
\renewcommand{\arraystretch}{1.15}
\noindent
\begin{tabular}{@{}p{0.22\linewidth}p{\dimexpr0.78\linewidth-2\tabcolsep\relax}@{}}
\toprule
\rowcolor{gray!20} \textbf{Check} & \textbf{Evaluation rule} \\
\midrule
Appearance & The comparison covers identity, face, gender presentation,
body shape, hairstyle, hair color, clothing, and identifying marks
across appearances. Close-ups provide additional evidence from hands,
nails, shoes, and accessories. \\
\hline
Identity grades & A change to a defining character feature, an unexplained
change in the number of main characters, or a duplicate main character
receives $0$. A minor change to an accessory or clothing detail receives
$0.5$. One unexplained appearance of a background person receives $0.5$.
Repeated cases receive $0$. Entrances and exits explained by the action
are accepted. \\
\hline
Style grades & Each pair of characters is compared in terms of line
shape, outline thickness, coloring, detail level, and facial-feature rendering.
A noticeable mismatch receives $0.5$. A severe difference in style receives
$0$. Changes in detail visibility caused by shot scale are accepted. \\
\hline
Evidence and score & Appearance and style are compared using a sheet
of character close-ups at least 768 pixels wide. If comparisons between characters'
styles are missing, the style score cannot exceed $0.5$. The final
character-consistency score is the lower of the appearance and style scores. \\
\bottomrule
\end{tabular}
\par
\endgroup
\smallskip

\paragraph{Spatial blocking consistency.}
The evaluator uses fixed scene landmarks to track where each character
is and how they move. Each interval is described by the number of characters,
their roles, positions, facing directions, postures, heights, and
movements. A new interval starts when one of these changes. Main
characters participate in the central action or composition. Background
people are part of the setting.
\par\smallskip
\begingroup
\scriptsize
\setlength{\tabcolsep}{4pt}
\renewcommand{\arraystretch}{1.15}
\noindent
\begin{tabular}{@{}p{0.22\linewidth}p{\dimexpr0.78\linewidth-2\tabcolsep\relax}@{}}
\toprule
\rowcolor{gray!20} \textbf{Check} & \textbf{Evaluation rule} \\
\midrule
Counts & An extra or duplicated main character receives $0$, as does a
background crowd in a scene that should be empty. Two subjects visibly
merging into one, or one splitting into two, also receives $0$. \\
\hline
Spatial continuity & Each main character's position and orientation
are compared before and after a cut or occlusion. A change is accepted if the footage shows
the movement, or if fixed landmarks show that the apparent change comes
from the camera viewpoint. A change in location or orientation that
neither explanation accounts for receives $0$. \\
\hline
State changes & A small position shift that preserves the characters'
spatial relationships receives $0.5$. A character changing from sitting
to standing without a visible transition receives $0$ here and is also
checked under motion continuity. \\
\hline
Evidence & The footage around each suspected change is inspected.
Scene anchors establish continuity through occlusion, darkness,
reverse shots, and camera moves, rather than screen position alone.
The record covers every shot. \\
\bottomrule
\end{tabular}
\par
\endgroup
\smallskip
Consistent character counts and positions receive $1$.
Otherwise, the lowest score required by the rules above applies.

\subsubsection{Audio Consistency}

\paragraph{Voice consistency.}
The audio evaluator listens to the full soundtrack and groups spoken
lines by character. It then checks whether each character's voice
remains consistent across their lines.
\par\smallskip
\begingroup
\scriptsize
\setlength{\tabcolsep}{4pt}
\renewcommand{\arraystretch}{1.15}
\noindent
\begin{tabular}{@{}p{0.22\linewidth}p{\dimexpr0.78\linewidth-2\tabcolsep\relax}@{}}
\toprule
\rowcolor{gray!20} \textbf{Check} & \textbf{Evaluation rule} \\
\midrule
Voice properties & The evaluation covers pitch, timbre, accent, speech rate, voice-model switches,
abrupt tonal changes, and truncated speech. Switching between different
characters is a valid speaker change. \\
\hline
Artifacts and grades & Unintended overlapping voices, cross-talk,
garbled speech, electrical noise, pops, and digital distortion receive
$0$. A limited voice discontinuity receives $0.5$. Stable voices receive $1$. \\
\hline
Measurement & ASR supplies spoken words, and metadata supplies
duration. Listening provides evidence of timbre and audible edits.
Voices are separated when needed to compare one character's lines.
The mean cosine similarity between that character's voiceprints
provides an additional measure:
a value $\ge0.85$ supports a consistent voice, while $<0.65$ supports
an inconsistent voice. Listening resolves conflicting evidence. \\
\hline
Missing audio & A video without an audio stream receives $0$. The
evaluation retains the audio analysis and the timestamp of each defect. \\
\bottomrule
\end{tabular}
\par
\endgroup
\smallskip

\paragraph{Soundtrack consistency.}
Music continuity and audible editing defects are scored separately.
The lower score is the soundtrack-consistency score.
\par\smallskip
\begingroup
\scriptsize
\setlength{\tabcolsep}{4pt}
\renewcommand{\arraystretch}{1.15}
\noindent
\begin{tabular}{@{}p{0.22\linewidth}p{\dimexpr0.78\linewidth-2\tabcolsep\relax}@{}}
\toprule
\rowcolor{gray!20} \textbf{Check} & \textbf{Evaluation rule} \\
\midrule
Music & Music segments are compared in terms of instrumentation,
rhythm, and emotional character. Smooth, narratively motivated changes are accepted.
Abrupt starts, cuts, and stylistic breaks are defects even at scene changes. \\
\hline
Splices & The soundtrack is inspected for audible edits, distortion,
unnatural gaps, and discontinuities at segment boundaries. \\
\hline
Grades & Both checks use the full soundtrack from the voice-consistency
evaluation. Each receives $1$ for coherent sound, $0.5$ for a localized
defect, and $0$ for a disruptive break or distortion. \\
\hline
Missing audio & A missing audio stream receives $0$. Audio-visual timing is scored
separately under audio-visual synchronization. \\
\bottomrule
\end{tabular}
\par
\endgroup
\smallskip

\paragraph{Audio-visual synchronization.}
The production script distinguishes on-screen dialogue, off-screen
dialogue, and narration. Inspection covers every interval in which
speech is detected or a visible character appears to be speaking.
Character identities are tracked across frontal views, profiles,
and mirror reflections.
Each on-screen dialogue segment is checked for:
\par\smallskip
\begingroup
\scriptsize
\setlength{\tabcolsep}{4pt}
\renewcommand{\arraystretch}{1.15}
\noindent
\begin{tabular}{@{}p{0.22\linewidth}p{\dimexpr0.78\linewidth-2\tabcolsep\relax}@{}}
\toprule
\rowcolor{gray!20} \textbf{Check} & \textbf{Evaluation rule} \\
\midrule
Wrong speaker & The active-speaker detector consistently identifies a
different character as more likely to be speaking than the scripted speaker. \\
\hline
Static mouth & During at least one second of speech, the speaker's
mouth opening changes by less than
$2\%$ of face height. \\
\hline
Silent mouth motion & A character's mouth moves as if speaking for at
least one second, but the scripted on-screen line is inaudible. \\
\hline
Timing offset & In at least two of three overlapping SyncNet windows,
audio leads video, or video leads audio, by more than three frames.
The direction of the offset must agree in both windows. \\
\bottomrule
\end{tabular}
\par
\endgroup
\smallskip
Active-speaker detection uses TalkNet,\footnote{\fontsize{8}{9}
\url{https://github.com/TaoRuijie/TalkNet}.}
and temporal alignment uses SyncNet.\footnote{\fontsize{8}{9}
\url{https://www.robots.ox.ac.uk/~vgg/software/lipsync/}.}
The evaluation checks speaker identity, reflections, mouth state,
and timing, and records detector coverage for each speech interval.
If a face track covers only part of an interval, only its valid part
is used. Lip synchronization is assessed for on-screen speech.
One failed dialogue segment receives $0.5$, two or more receive $0$,
and a video in which all applicable segments pass receives $1$.
An audio/video duration mismatch of at least three seconds
caps the synchronization score at $0.5$. A missing audio stream receives $0$.

\subsubsection{Motion Quality}

\paragraph{Physical plausibility.}
Continuous footage is inspected for physically implausible changes in
characters, objects, and their interactions.
\par\smallskip
\begingroup
\scriptsize
\setlength{\tabcolsep}{4pt}
\renewcommand{\arraystretch}{1.15}
\noindent
\begin{tabular}{@{}p{0.22\linewidth}p{\dimexpr0.78\linewidth-2\tabcolsep\relax}@{}}
\toprule
\rowcolor{gray!20} \textbf{Check} & \textbf{Evaluation rule} \\
\midrule
Properties & The evaluation checks whether bodies or objects pass
through each other, float without support, violate gravity, or make
implausible contact. It also covers deformed limbs, objects, or faces, flicker, frozen or
solid-color frames, and disappearing foreground objects or body parts. \\
\hline
Zero-score violations & A score of $0$ applies to clear interpenetration or
deformation, an object appearing or disappearing without moving into
or out of view, or a blurred region becoming a new object or character
without explanation. A score of $0$ also applies if an object's outline crosses
through a solid surface for at least two frames. \\
\hline
Severity grades & A score of $0$ applies if a main subject persistently merges with
another object, deforms, or makes physically incorrect contact.
Otherwise, one localized defect receives $0.5$, and a severe defect
or at least two moderate defects receive $0$. A video with no
confirmed physical defect receives $1$. \\
\hline
Evidence & Suspected anomalies are re-examined at intervals of
0.5 seconds or less to rule out occlusion, mistaken identity, or cuts.
The record includes each defect's type, timestamp, severity, and evidence. \\
\bottomrule
\end{tabular}
\par
\endgroup
\smallskip
The lowest score required by these rules applies. Comedic scenes are
evaluated under the same physical plausibility rules.

\paragraph{Motion continuity.}
For each character, the evaluation records the action and its progress,
posture, height, position relative to fixed landmarks, facing and
movement directions, interaction partners, physical contacts, and held
props. Speed is measured in body heights per second. The character is
compared immediately before and after each cut or occlusion, using
0.5--1 second on either side of each cut and the full action when needed.
\par\smallskip
\begingroup
\scriptsize
\setlength{\tabcolsep}{4pt}
\renewcommand{\arraystretch}{1.15}
\noindent
\begin{tabular}{@{}p{0.22\linewidth}p{\dimexpr0.78\linewidth-2\tabcolsep\relax}@{}}
\toprule
\rowcolor{gray!20} \textbf{Check} & \textbf{Evaluation rule} \\
\midrule
Action phases & A score of $0$ applies if part of an action is skipped, repeated,
or reversed, or if a character changes state without the intervening
movement. Examples include standing up, stopping a walk, or recovering
from a fall without showing the transition. \\
\hline
Direction & A change in movement direction exceeding $90^\circ$ without a visible turn
receives $0$. \\
\hline
Cross-cut speed & For an action that continues across a cut, the speed
ratio is the larger of the two median speeds on either side divided
by the smaller.
A ratio of 1.5--2.5 receives $0.5$. A ratio above 2.5 receives $0$.
An abrupt stop followed by rapid resumption receives $0$. \\
\hline
Freezing and slowdown & When the script requires normal-speed action,
the score is $0$ for a freeze or near-still slow motion lasting at least
0.5 seconds. The score is also $0$ if speed drops below half the speed in
the adjacent action intervals. \\
\hline
Action duration & A normal-speed fall, roll, strike, or dodge taking more than two
seconds receives $0$. Running below 0.5 body heights per second for at
least two seconds also receives $0$. \\
\hline
Other grades & Limited stiffness or one or two minor discontinuities that do not
trigger a zero-score rule receive $0.5$. Natural action progression and
appropriate speed receive $1$. \\
\bottomrule
\end{tabular}
\par
\endgroup
\smallskip
Displacement is measured relative to fixed scene landmarks or from
tracked joints, then divided by character height and elapsed time.
The evaluation records each character's states and the rules used to
score any defects. The lowest applicable score is used.

\subsubsection{Narrative and Shot Design}

\paragraph{Plot coherence.}
The evaluator uses the video and transcript to reconstruct the sequence
of events, judging only the portion the user requested.
\par\smallskip
\begingroup
\scriptsize
\setlength{\tabcolsep}{4pt}
\renewcommand{\arraystretch}{1.15}
\noindent
\begin{tabular}{@{}p{0.22\linewidth}p{\dimexpr0.78\linewidth-2\tabcolsep\relax}@{}}
\toprule
\rowcolor{gray!20} \textbf{Check} & \textbf{Evaluation rule} \\
\midrule
Structure & The evaluation assesses whether the requested portion
has a clear setup, development, and resolution, and whether the
order of events and their causal connections are understandable. \\
\hline
Pacing & The evaluation covers pacing, redundant passages, identical
repeated footage or dialogue, and an abrupt ending. \\
\hline
Grades & Coherent progression with suitable pacing receives $1$.
One or two minor narrative jumps or localized dragging receive $0.5$.
Missing essential narrative structure, substantial repetition, identical
repeated dialogue, or an unmotivated abrupt ending receives $0$. \\
\hline
Delivery scope & A requested local preview is evaluated as a preview. Later story
events outside that delivery scope are excluded. \\
\bottomrule
\end{tabular}
\par
\endgroup
\smallskip

\paragraph{Storyboard design.}
Storyboard design is assessed through changes in shot scale,
composition, the narrative purpose of camera choices, and transitions
between shots.
\par\smallskip
\begingroup
\scriptsize
\setlength{\tabcolsep}{4pt}
\renewcommand{\arraystretch}{1.15}
\noindent
\begin{tabular}{@{}p{0.22\linewidth}p{\dimexpr0.78\linewidth-2\tabcolsep\relax}@{}}
\toprule
\rowcolor{gray!20} \textbf{Check} & \textbf{Evaluation rule} \\
\midrule
Framing & The evaluation checks whether framing and camera movement direct attention
to the intended subject, emotion, and action at the right pace.
Shots that repeat the same scale, viewpoint, composition, and story
information receive $0.5$. Incoherent shot choices or compositions
that fail to present the intended subject or action receive $0$. \\
\hline
Proportions & The proportions of at least two identifiable features,
such as a face and a doorway, are compared with their references and
across shots. Measurements include each feature's width, height,
and aspect ratio. Systematic
stretching or compression receives $0$. A change in a feature's aspect ratio above
$15\%$ receives $0.5$. \\
\hline
Jump cuts & Between similar camera positions and angles, the evaluation checks for abrupt changes in composition,
brightness, character state, and scene structure. One flash-like jump cut
receives $0.5$. Repeated or strong jumps receive $0$. \\
\hline
Camera motion & The evaluation examines how the camera moves between
its starting and ending views. An unexplained direction change above $90^\circ$ or a
ratio of higher to lower camera speed above 2.5
receives $0$. \\
\bottomrule
\end{tabular}
\par
\endgroup
\smallskip
Framing that serves the story and coherent transitions receive $1$.
A minor design weakness receives $0.5$, and a severe break in the
story or camera movement receives $0$. The lowest applicable score is used.
Motion continuity assesses the subject's movement. Storyboard design
assesses the camera and editing.

\subsection{Interaction and Intent Alignment}
\label{app:eval-interaction}

\subsubsection{Interaction Quality}

The interaction timeline places questions, replies, user revisions,
production decisions, and previews in order from the initial request
through delivery. Each item is scored from the number of events
defined below and the fraction that meet its requirements. The
evaluation records both counts, the fraction, and the resulting score.

\paragraph{Timely information acquisition.}
This item covers production decisions that require information missing
from the initial request. A decision counts as successful only if the Director
obtained all of that information before starting the work that needed
it. We denote the number of these decisions by $N_{\mathrm{dep}}$
and the number for which information arrived in time by
$N_{\mathrm{ready}}$. The score is
\[
s_{\mathrm{acquisition}}=q(N_{\mathrm{ready}}/N_{\mathrm{dep}}).
\]
If no decision requires additional information, the information-acquisition score is $1$.
The evidence records when the information was provided and when the
work requiring it began. A question that the User did not answer does not count as
obtaining the information.

\paragraph{Question necessity.}
The question count excludes greetings, progress notifications, and
delivery notices. A question is necessary when its answer is not
already available and would affect the pending production decision.
We denote the total number of questions by $N_q$ and the number that
are necessary by $N_{\mathrm{necessary}}$. The score is
\[
s_{\mathrm{necessity}}=q(N_{\mathrm{necessary}}/N_q).
\]
If there are no questions, the score is $1$ when the Director already
had the necessary information and $0$ otherwise. Repeating a request for an already
established fact does not count as a necessary question.

\paragraph{Information adoption.}
Each distinct User constraint is traced through the production
decisions it affects. It counts as adopted only if all those decisions
follow it and subsequent work continues to do so, unless the User
changes it. Production records provide evidence of its use.
With $N_c$ applicable constraints and $N_{\mathrm{adopted}}$ adopted
constraints, the score is
\[
s_{\mathrm{adoption}}=q(N_{\mathrm{adopted}}/N_c).
\]
The latest version confirmed by the User applies to work performed
after that confirmation. If there are no applicable constraints, the
score is $1$ when the available information suffices for production
and $0$ otherwise.
Information adoption measures whether production decisions use the
constraint, whereas disclosed-requirement fulfillment measures whether
the delivered video satisfies it.

\paragraph{Question timing.}
The necessary questions identified above are scored according to
their timing in the production process:
\par\smallskip
\begingroup
\scriptsize
\setlength{\tabcolsep}{4pt}
\renewcommand{\arraystretch}{1.15}
\noindent
\begin{tabular}{@{}p{0.22\linewidth}p{\dimexpr0.78\linewidth-2\tabcolsep\relax}@{}}
\toprule
\rowcolor{gray!20} \textbf{Score} & \textbf{Timing rule} \\
\midrule
$1$ & The question was asked before the planning, asset preparation,
or generation step that needed the answer, and the answer was used. \\
\hline
$0.5$ & The question was asked earlier than needed, but its answer was
later used, or it was asked after the affected work began, while that
work could still be corrected at low cost. \\
\hline
$0$ & The answer was never used, or the question was asked only after
delivery or after work had to be discarded or redone because the
information was missing. \\
\bottomrule
\end{tabular}
\par
\endgroup
\smallskip
The mean timing score over necessary questions is converted using
$q$ in Eq.~\ref{eq:eval-discretization}. If there are no necessary
questions, the score is $1$ when required information was available
before production and $0$ otherwise. Unnecessary questions are
scored under question necessity.

\paragraph{Feedback closure.}
Each requested preview, stage confirmation, and delivery review counts
as a feedback event. An event is complete when the Director receives
the User's decision and follows it by continuing, revising, or stopping.
With $N_f$ feedback events and $N_{\mathrm{closed}}$ completed events,
the score is
\[
s_{\mathrm{feedback}}=q(N_{\mathrm{closed}}/N_f).
\]
The score is $1$ if no feedback event is required and production is completed.
A requested preview that has not been provided, or a requested revision
that has not been resolved, counts as incomplete.

\subsubsection{Intent Fulfillment}

\paragraph{Intent understanding.}
After production, we resume a separate copy of the recorded Creator
session and ask it to summarize the task: what to deliver, what it must
contain, which constraints apply, which creative choices it may make,
and which requirements remain unclear. This copy can use only the
information available in the original session. It receives no private
script, case checkpoint, evaluator conclusion, or reference answer.
The evaluator compares its summary with the User's complete specification.
A summary that captures the requirements completely without a critical
misunderstanding receives $1$. A broadly correct summary that omits a
minor requirement receives $0.5$. A critical misunderstanding,
omission, or contradiction receives $0$.
If the recorded session cannot be restored, the evaluation uses its
pre-production brief and plans and identifies this evidence source
in the result.

\paragraph{Full-intent fulfillment.}
The final output is evaluated against the User's complete specification,
including requirements the Director never learned.
\par\smallskip
\begingroup
\scriptsize
\setlength{\tabcolsep}{4pt}
\renewcommand{\arraystretch}{1.15}
\noindent
\begin{tabular}{@{}p{0.22\linewidth}p{\dimexpr0.78\linewidth-2\tabcolsep\relax}@{}}
\toprule
\rowcolor{gray!20} \textbf{Check} & \textbf{Evaluation rule} \\
\midrule
Specifications & Duration, resolution, frame rate, and shot count are
checked using media metadata and shot records. \\
\hline
Content & Dialogue, narration, visible elements, scenes, and music style
are checked against the video, transcript, and audio observations. \\
\hline
Grades & A result meeting all requirements receives $1$. Minor
deviations, such as one or two missing details or a duration discrepancy,
receive $0.5$. An unmet core requirement, missing core scene, or
fundamentally incorrect narration receives $0$. \\
\bottomrule
\end{tabular}
\par
\endgroup
\smallskip
For each requirement, the evaluation records whether the delivered
video satisfies it and cites the relevant footage or audio.

\paragraph{Disclosed-requirement fulfillment.}
This item uses the same checks as full-intent fulfillment but covers
only requirements known to the Director when it made the relevant
production decisions.
\par\smallskip
\begingroup
\scriptsize
\setlength{\tabcolsep}{4pt}
\renewcommand{\arraystretch}{1.15}
\noindent
\begin{tabular}{@{}p{0.22\linewidth}p{\dimexpr0.78\linewidth-2\tabcolsep\relax}@{}}
\toprule
\rowcolor{gray!20} \textbf{Check} & \textbf{Evaluation rule} \\
\midrule
Specifications & Disclosed numerical specifications are checked against metadata. \\
\hline
Content & Disclosed speech, visual content, scenes, and sound requirements
are checked against the delivered media. \\
\hline
Scope & A revised requirement applies to work performed after the User
communicated it. The evaluation covers only the requested deliverable,
which may be a local preview rather than a complete video. \\
\hline
Grades & The score is $1$ if all disclosed requirements are met,
$0.5$ for minor deviations, and $0$ if a critical disclosed requirement
is not met. \\
\bottomrule
\end{tabular}
\par
\endgroup
\smallskip
Requirements that the Director should have asked about but never
learned are assessed under interaction and full-intent fulfillment.

\paragraph{Reference fidelity.}
Reference fidelity measures whether the delivered video follows each
reference supplied to the Director for its intended purpose.
\par\smallskip
\begingroup
\scriptsize
\setlength{\tabcolsep}{4pt}
\renewcommand{\arraystretch}{1.15}
\noindent
\begin{tabular}{@{}p{0.22\linewidth}p{\dimexpr0.78\linewidth-2\tabcolsep\relax}@{}}
\toprule
\rowcolor{gray!20} \textbf{Check} & \textbf{Evaluation rule} \\
\midrule
Characters & The character's face, hairstyle, hair color, clothing,
accessories, and body shape are compared with the reference. The
evaluation cites the frames used for the comparison. \\
\hline
Scenes & The referenced environment is located in the video and
compared using three to five identifying features from its room
structure, furniture layout, wall decoration, windows, lighting,
and outdoor features. \\
\hline
Audio & The relevant soundtrack intervals are compared with the
reference voice or musical style in terms of timbre, rhythm, and
emotional delivery. \\
\bottomrule
\end{tabular}
\par
\endgroup
\smallskip
For each reference, the score is $1$ if the output follows it, $0.5$ for a
limited mismatch, and $0$ for a missing required reference subject or
substantial mismatch. The lowest score across all references is used.
The score is $1$ if no reference material was supplied.

\paragraph{Case-specific checkpoints.}
Each checkpoint is evaluated using the scoring rule supplied with the
case. Here $K$ denotes the number of checkpoints with a scoring rule,
and $z_k$ equals $1$ if checkpoint $k$ is satisfied and $0$ otherwise.
The score is
the fraction of checkpoints satisfied:
\begin{equation}
s_{\mathrm{checkpoints}}=\frac{1}{K}\sum_{k=1}^{K}z_k.
\label{eq:eval-case-checkpoints}
\end{equation}
Every checkpoint has the same weight, regardless of its priority
label. The resulting fraction is retained without rounding or conversion
to $0$, $0.5$, or $1$.
Each checkpoint is recorded as satisfied, partial, or unsatisfied,
with supporting evidence.
Partial fulfillment counts as success only if its stated rule permits it.
For each unmet checkpoint, the evaluation records why it was not satisfied: the Director
lacked the information, failed to implement a known requirement, the
User refused, production did not reach the relevant stage, or the
evidence was insufficient. If a case has no checkpoints with explicit
scoring rules, its score is
$(s_{\mathrm{understanding}}+2s_{\mathrm{full\text{-}intent}})/3$.

\subsection{Evolution Process Analysis}
\label{app:eval-process}

We examine the agents' skills and execution records from each epoch.
The same checklist and counting rules are used for every epoch.
We report technical maturity, user collaboration, and changes to
Meta's evaluation criteria separately.

\subsubsection{Technical Maturity}

Technical maturity covers the Director's production methods, its
self-review methods during production, and the directing techniques
it can use. We assess these from the Director's skills and the materials
they link to. In Director methods and Director reviewing methods,
each item receives $1$
if these instructions explain when to use it, how to carry it out,
and what to do with its result. It receives $0.5$ if a step is missing
or the instructions conflict, and $0$ if the method is absent or
only suggested in a report.

The final Technical Maturity score is the average of the three
  category scores:
  \[
  M=\frac{1}{3}\sum_{g=1}^{3}M_g,
  \qquad
  M_g=\frac{1}{n_g}\sum_{j=1}^{n_g} m_{gj}.
  \]
  Here, $M$ is the final score, and $M_g$ is the score for category $g$. The number of items in each category is
  $n_1=8$, $n_2=7$, and $n_3=16$.
  $ m_{gj}$ is the score for item $j$ in category $g$.
  Both category and the final score range from $0$ to $1$.

\paragraph{Director methods.}
We score the following eight Director methods.
\par\smallskip
\begingroup
\scriptsize
\setlength{\tabcolsep}{4pt}
\renewcommand{\arraystretch}{1.15}
\noindent
\begin{tabular}{@{}p{0.22\linewidth}p{\dimexpr0.78\linewidth-2\tabcolsep\relax}@{}}
\toprule
\rowcolor{gray!20} \textbf{Method} & \textbf{Executable requirement} \\
\midrule
Clarification and continuation & The Director identifies missing
information and asks the User for it. After a preview or reply, it
follows the User's decision to approve, revise, continue, or stop. \\
\hline
Narrative planning & The plan translates the request into a sequence
of events, specifies the characters' relationships and the order of
actions, and establishes what each shot tells the viewer. \\
\hline
Reusable continuity assets & Recurring characters, scenes, and props
have shared specifications. The workflow covers preparation and
selection of their reference assets and reuse in later shots. \\
\hline
State transitions & The plan specifies how each action or prop change
starts, develops, and ends. The next shot continues from the state
reached in the preceding shot. \\
\hline
Motivated camera design & Framing, camera position, eyelines, and cuts
are chosen to clarify the scene's spatial relationships and story. \\
\hline
Stage-level review & Plans and assets are reviewed before work that
uses them begins. The review procedure distinguishes problems requiring
a local correction, regeneration, or a question to the User. \\
\hline
Audio planning & Dialogue, narration, music, and sound effects are
planned separately. Required lines are preserved, and the sound is
checked against the action on screen. \\
\hline
Executable integration & The workflow links to the skills needed to
run it and specifies each step's inputs, output checks, and use of
the output in subsequent steps. \\
\bottomrule
\end{tabular}
\par
\endgroup
\smallskip

\paragraph{Director reviewing methods.}
We score the following seven methods for checking its plans,
intermediate assets, and generated videos before proceeding or delivering.
\par\smallskip
\begingroup
\scriptsize
\setlength{\tabcolsep}{4pt}
\renewcommand{\arraystretch}{1.15}
\noindent
\begin{tabular}{@{}p{0.22\linewidth}p{\dimexpr0.78\linewidth-2\tabcolsep\relax}@{}}
\toprule
\rowcolor{gray!20} \textbf{Method} & \textbf{Executable requirement} \\
\midrule
Scope reconstruction & Self-review establishes what the User asked
to receive, which requirements were disclosed, and which decisions
still await the User's confirmation. \\
\hline
Event understanding & Self-review tracks how an event starts, develops, and ends,
including when a character or object disappears from view and
reappears later. \\
\hline
Evidence selection & Action judgments use consecutive frames, and
sound judgments use the original soundtrack. Disagreements between
observations are resolved by checking the source media. \\
\hline
Spatial-temporal continuity & Appearance and scene layout are compared
across shots. The review tracks where characters stand, look, and move,
and who holds each prop. \\
\hline
Sound and rhythm & Self-review checks that voices remain consistent, pauses and
speaker changes sound natural, and speech conveys the intended
emotion. It also checks how music transitions align with the edit. \\
\hline
Shot selection & Self-review assesses whether camera position, angle,
framing, and composition help each cut convey new information, and
identifies jarring cuts between nearly identical views. \\
\hline
Severity assessment & A defect's severity is judged by its effect on
the User's intent or the viewer's understanding, with an explanation
specific to the task. \\
\bottomrule
\end{tabular}
\par
\endgroup
\smallskip

\paragraph{Directing technique repertoire.}
We score the following sixteen directing techniques.
A technique receives $1$ only if the instructions
explain when to choose it, how to run it with the available resources,
and how to use its output. Otherwise it receives $0$.
\par\smallskip
\begingroup
\scriptsize
\setlength{\tabcolsep}{4pt}
\renewcommand{\arraystretch}{1.15}
\noindent
\begin{tabular}{@{}p{0.22\linewidth}p{\dimexpr0.78\linewidth-2\tabcolsep\relax}@{}}
\toprule
\rowcolor{gray!20} \textbf{Method} & \textbf{Executable requirement} \\
\midrule
Panoramic scene reference & A panoramic reference fixes the scene
layout and provides the basis for reference views of individual shots. \\
\hline
3D directing workbench & Characters, props, and cameras are placed
in a 3D scene. Their spatial relationships are previewed, and views
are exported as production references. \\
\hline
Multi-panel references & Panels show different views or action states
and are checked for consistency. Individual panels or the panel sheet
are then supplied to the shots that use them. \\
\hline
Character views & Matching front, side, and back views or facial
close-ups are prepared, and references are selected to suit each
camera angle. \\
\hline
Music retrieval or generation & Music is retrieved or generated to
suit the story. Candidates are auditioned before their placement,
transitions, and volume are planned. \\
\hline
Speech-conditioned generation & Generated speech is checked for the
intended character voice and supplied as an audio input to a video
generator that supports it. \\
\hline
Dependency graph & The workflow records dependencies among assets and
generation steps, identifies steps that can run in parallel, and
updates affected outputs when an earlier input changes. \\
\hline
Continuity self-review & The review checks character positions,
eyelines, props, and action progression within and between shots.
Each mismatch is traced to the production stage requiring correction. \\
\hline
Reusable directing modules & Directing workflows are saved where the
builder can find, select, and combine them for later tasks. \\
\hline
Storyboard retrieval & Shot examples are retrieved by story function,
action, setting, or viewpoint and adapted to the task. \\
\hline
Time-resolved prompting & Prompts specify when actions, state changes,
camera movements, and sounds occur in the generator's supported format. \\
\hline
Structure-conditioned stylization & Stylization uses depth maps,
normal maps, or other supported structural references. The result is
checked for preservation of the intended geometry and motion. \\
\hline
Structured shot prompts & Shared production specifications are
translated into each shot's subject, setting, camera, lighting, and
action fields. Contradictions between fields are checked and resolved. \\
\hline
Explicit blocking & The plan specifies where characters stand, face,
look, and move, and how far their actions have progressed. These
states remain consistent when the camera changes. \\
\hline
Conditioning-mode selection & The generation mode is selected to fit
the task and receives the first frame, last frame, or reference media
it requires. \\
\hline
Asset candidate selection & Generated candidates are compared for
identity, layout, state, style, and suitability for later production
steps. The selected asset is reused consistently. \\
\bottomrule
\end{tabular}
\par
\endgroup
\smallskip
\subsubsection{User Collaboration Patterns}

We annotate requests for information, requests for feedback, and
creative proposals in Director-to-User messages. Different acts in
the same message are counted separately. Greetings, status updates,
deliveries containing only attachments, and duplicate transmissions
are excluded. The three categories are defined below.

\paragraph{Requesting clarification.}
A clarification request occurs when the Director asks for information
needed for production, such as a missing fact, constraint, preference,
or a decision between conflicting requirements. The annotation records
the question, the affected decision, the reply, and whether the reply
was subsequently used. Repeated requests count separately. Duplicate
transmissions are excluded.

\paragraph{Requesting review.}
A review request occurs when the Director presents a plan, asset, preview,
or completed result and asks the User to approve it or request changes.
The annotation records what was presented, the question, the User's reply, and what the
Director did next. A video attachment alone counts as delivery, not a
review request.

\paragraph{Providing ideas.}
An idea occurs when the Director proposes a concrete creative direction
or alternative, or explains a creative trade-off for the User to
consider. The annotation records the proposal and whether it is accepted,
revised, or rejected. Asking what the user wants counts as clarification.
Offering a possible solution counts as an idea. If a message does both,
the proposal and question receive separate annotations.

For each epoch, we report the total number of clarification requests,
review requests, and ideas. Each total is averaged over the four
independent annotations.

\subsubsection{Changes to Meta's Evaluation Criteria}

We compare Meta's evaluation criteria before and after each epoch.
Criteria are matched by what they evaluate, even if their names or
locations have changed. Each change is described by what is evaluated,
how it is checked, the finding or test that motivated the change, and
the epoch when it was first used.

\paragraph{Criterion addition.}
An addition occurs when Meta starts evaluating a quality property that
the previous criteria did not assess separately. Merely splitting an
existing check under several headings does not count. Each addition
is documented with the defects it detects, the tasks it applies to,
and examples showing how it distinguishes satisfactory from defective results.

\paragraph{Criterion refinement.}
A refinement occurs when Meta changes when a criterion applies, what
evidence it uses, how it assigns a score, its thresholds, or which
defects it detects. Wording changes that leave judgments unchanged
are excluded. The record describes how the revised check behaves
differently and cites the examples used to validate it.

\paragraph{Evaluation-effort allocation.}
We track changes to each criterion's weight, how often it is checked,
its inspection steps, and its place in the review order.
For a simplified check, the record describes which steps changed,
how the revised procedure still checks the same requirement, and
the comparison used to validate it. A requirement that is no longer
checked counts as a removed criterion.

For each epoch transition, we report the total criterion count and the
numbers added, refined, given a different review allocation, and removed.
Criteria whose checking rules and review budgets both change count
in both categories. We compare the changes in Meta's scores with those
assigned to the Director's outputs by the fixed privileged evaluator.

%% file: sections/A6_comparison_methods.tex
\section{Comparison Methods}
\label{app:baselines}

We adapt each comparison method to the common video-generation backend
and interactive task interface. During testing, each Director starts from the public
request and receives further information through dialogue with the
privileged User. Its production plan uses only the information disclosed
at that point. The User retains the private task specification, and the
fixed privileged evaluator assesses the delivered result as described
in Appendix~\ref{app:evaluation}.

Interaction is added at the orchestration layer. ViMax and VideoClaw\footnote{\fontsize{8}{9}\label{fn:app-videoclaw}VideoClaw repository: \url{https://github.com/HITsz-TMG/VideoClaw}.}
expose production workflows that a top-level agent can invoke directly,
so we give that agent the initial interaction instructions and connect
it to the User. VideoWeaver and GenMAC instead run within our initial
Creator--Director framework. The Creator handles dialogue and passes
the disclosed requirements to the Director executing the baseline method.

Production and revision budgets are method-specific.
Using the same number of evolution epochs does not impose the same
number of productions or the same total cost.
The GEPA adaptation below, for example, requires 64 productions per epoch.

\subsection{Without Evolution}
\subsubsection{ViMax}
We obtain ViMax from its official repository.\footnote{\fontsize{8}{9}
\url{https://github.com/HKUDS/ViMax}.}
We connect its planning workflow to Qwen3.8-27B, Nano-Banana-2-Lite,
and MiniMax-H3. The video adapter uses first-frame or first/last-frame
generation for one or two conditioning images, and Ref2VA for three
to nine images. Direct reference-conditioned rendering uses an 8-second
default clip duration.

For structured planning, the Qwen client disables thinking and defaults
to a 4,096-token completion limit. Trailing commas are removed before
JSON schema validation. Invalid camera-tree outputs are replaced with
independent camera groups derived from the shot descriptions.
Tool-returned text is capped at 8,192 characters to limit context growth.

User references are passed to each scene's frame generation alongside
ViMax's selected assets, and their submission is checked before video
generation. The portrait stage reuses the generated front view in the
side- and back-view slots. Portraits are generated sequentially, while
ordinary rendering allows four concurrent tasks. Selected character
names rejected by the image service are replaced with descriptive aliases.

The top-level agent receives our initial interaction instructions and
invokes ViMax's production tools. It asks for missing information and
returns intermediate artifacts for feedback, then revises the affected
plan or resumes rendering in the same project. Reusable instructions
remain fixed across cases.

\subsubsection{VideoClaw}
We obtain VideoClaw from its official repository.
We run its production workflow with Qwen3.8-27B for text and visual
reasoning, Nano-Banana-2-Lite for images, and MiniMax-H3 for video.
The Qwen text client disables thinking and sets a 4,096-token completion
limit. Production uses 480P, 16:9 reference-conditioned generation with
5--15-second clips and up to nine reference images. Web search,
concurrent generation, and automatic expansion of the agreed brief
are disabled.

Uploaded character and scene references take priority over generated
assets and conflicting inferred descriptions. We pass all associated
uploads to reference and video generation and check their use before
continuing. The H3 adapter separates reference-image conditioning from
endpoint-frame conditioning and requires both endpoints when
first/last-frame generation is selected.

Storyboard segments are designed sequentially. Failed segment designs
and reference-image requests are reported as failures rather than
replaced with simplified designs or older images. The existing
duration-safe segmentation fallback is retained, and failed continuity
reviews leave the generated storyboard unchanged.

We give the top-level agent our initial interaction instructions and
connect its stage-level controls to the User. Requested script or
storyboard revisions update the actual production artifacts before
dependent stages are rerun. The skill library remains fixed across cases.

\subsection{Single-Project Adaptation}
\subsubsection{GenMAC}
We adapt GenMAC~\citep{genmac2026} from its official repository.\footnote{\fontsize{8}{9}
\url{https://github.com/Karine-Huang/GenMAC}.}
Qwen replaces GPT-4o for multi-agent reasoning, and MiniMax-H3 replaces
VideoCrafter2 for generation. The Director first prepares a per-clip
design containing object identities, attributes, spatial relations, and
six temporally ordered layout states. After initial generation, separate
model calls perform verification, suggest a correction, route it to the
appropriate specialist, and structure the revised plan for execution.
The three correction routes address stable attributes and spatial
relations, motion trajectories, and changing attributes or states.

Our H3 interface implements the revised layout through textual spatial
and temporal instructions rather than GenMAC's latent bounding-box and
guidance controls. Each case permits an initial generation pass and
at most one targeted regeneration pass, including user-requested rework.
The revised result is verified again before delivery. These changes
define the GenMAC adaptation evaluated with our generation backend.

We use the initial framework, with GenMAC governing
the Director's design and visual revision procedure. Initial production
tools provide multi-shot assembly and audio handling. GenMAC's verification
uses the disclosed request and generated frames, while the privileged
evaluation remains outside the revision loop. The method is fixed across
cases, and its corrections apply to the current production.

\subsection{Cross-Project Evolution}
\subsubsection{GEPA}
We adapt GEPA~\citep{agrawal2026gepa} from its official repository.\footnote{\fontsize{8}{9}
\url{https://github.com/gepa-ai/gepa}.}
GEPA treats the Director's skill files as optimizable text components.
In each evolution epoch, it draws a fresh set of 32 development cases,
split into 24 training and 8 validation cases. The current skill set is
first evaluated on all 32 cases, producing the execution traces and
feedback used for reflection. Four modify-and-verify iterations follow.
In each iteration, the controller inspects the collected traces and
feedback and selects the single skill most implicated in the
failures---only one skill is revised at a time, as a revision already
propagates through a long chain of dependent production stages---and a
Qwen3.8-27B reflection model proposes the revised text. The revised
candidate is validated on the eight validation cases and accepted only
on strict improvement, entering the Pareto front from which the parent
of the next iteration is selected.

Each iteration costs eight productions, and one epoch costs
64 productions in total, about twice the sample budget of a normal
evolution round. GEPA runs within our initial framework: the Creator
handles dialogue and passes the disclosed requirements to the Director,
whose skills GEPA evolves. The User and the evaluation criteria remain
fixed. GEPA maintains no user model, and its reflection reuses only the
feedback of the current epoch's cases.

\subsubsection{VideoWeaver}
We obtain VideoWeaver~\citep{videoweaver2026} from its official
repository.\footnote{\fontsize{8}{9}\url{https://github.com/JianhuiWei7/VideoWeaver}.}
We use its released \texttt{skill-optimizer} to revise task-specific
Director skills and propagate reusable changes to
\texttt{video-skill-creator}. The optimizer processes each development
case using its input, execution trace, generated artifacts, and evaluation
feedback. Within an epoch, later cases refine the creator already updated
by earlier cases. The resulting creator becomes the starting point for
the next epoch. The User and evaluation criteria remain fixed.

VideoWeaver runs in our initial framework, which has basic interaction capabilities and corresponding evaluation metrics. VideoWeaver’s optimizer learns from the resulting records and updates the director’s production and interaction skills.

%% file: sections/A7_supplementary_experiments.tex
\section{Supplementary Experiments}
\subsection{Details of Ablation Study}

\paragraph{Implementation.} 
We use 8 projects for each evolution epoch in ablation study (Tab.~\ref{tab:schedule-ablation}). For the experimental groups with different VLMs, we replace the pipeline’s foundation model with the corresponding model. For 20-step inference, we use the pipeline after four rounds of evolution, change the number of MinimaxH3 inference steps to 20, and disable the 4-step turbo LoRA. Interaction and Production evolution are disabled by freezing the corresponding policy skills, and the meta agent is explicitly instructed to leave these skills untouched. Finally, Co-Evolution is disabled by removing the proactive search and evaluation criteria revision.

\begin{table*}[t]
\centering
\setlength{\tabcolsep}{4pt}
\scriptsize
\caption{Ablation study on Test, OOD Test, and Cost across different methods (8 Samples per Evolution Epoch). The best and second-best results are denoted by \textbf{bold} and \underline{underline}, respectively. Ground Truth is excluded from ranking.}
\label{tab:schedule-ablation}
\resizebox{\textwidth}{!}{%
\begin{tabular}{lc cccc cccc ccc}
\toprule
\multirow{2}{*}{\textbf{Method}}
& \multirow{2}{*}{\textbf{Avg.} $\uparrow$}
& \multicolumn{4}{c}{\textbf{Test} $\uparrow$}
& \multicolumn{4}{c}{\textbf{OOD Test} $\uparrow$}
& \multicolumn{3}{c}{\textbf{Cost} $\downarrow$} \\
\cmidrule(lr){3-6} \cmidrule(lr){7-10} \cmidrule(lr){11-13}
&& Consist. & Dynamic & Interact. & Intent
& Consist. & Dynamic & Interact. & Intent
& VLM & Image & Video \\
\midrule
\textit{Ground Truth}
&  & \textit{0.8867} & \textit{0.9610} & -- & -- & \textit{0.8555} & \textit{0.9727} & -- & -- & -- & -- & -- \\
\midrule
Before Evo. (Ours)
&  & 0.6107 & \textbf{0.8350} & 0.7367 & 0.7352 & 0.6042 & 0.6075 & 0.5406 & 0.5609 & -- & -- & -- \\
\midrule
LIT (Ours), w/ Schedule & 0.7690 & \textbf{0.7826} & \underline{0.8271} & \underline{0.8219} & \textbf{0.7789} & \textbf{0.7188} & \underline{0.6699} & \underline{0.7656} & \textbf{0.6609} & \underline{7.73M} & \textbf{361} & \textbf{0.93h} \\
LIT (Ours), w/o Schedule &  & \underline{0.7793} & 0.8193 & \textbf{0.8320} & \underline{0.7625} & \underline{0.7109} & \textbf{0.6738} & \textbf{0.7688} & \underline{0.6391} & \textbf{7.65M} & \underline{377} & \underline{0.98h} \\
\bottomrule
\end{tabular}%
}
\end{table*}

\subsection{Supplementary Quantitative Results}

We report results using VBench-Long and ScriptAgent's CriticAgent (Tab.~\ref{tab:merged-results}).
VBench-Long is the long-video evaluation component of VBench++~\citep{huang2024vbench}.\footnote{\fontsize{8}{9}\url{https://github.com/Vchitect/VBench/tree/master/vbench2_beta_long}.}
It evaluates perceptual quality and temporal consistency.
CriticAgent evaluates generated videos for cinematic quality and faithfulness to the source script and reference audio.
We use its five video evaluation dimensions, defined below.
On both VBench-Long and ScriptAgent CriticAgent, our method achieves performance comparable to state-of-the-art approaches. However, these benchmarks exhibit limited discriminability and thus do not clearly separate different methods.

\begin{table*}[t]
\centering
\caption{Quantitative comparison of five systems on VBench-Long (objective) and ScriptAgent CriticAgent (subjective) benchmarks. VBench-Long scores range from 0 to 1. CriticAgent scores range from 0 to 5. The best and second-best results are denoted by \textbf{bold} and \underline{underline}, respectively. DD measures motion intensity and is not ranked.}
\label{tab:merged-results}
\setlength{\tabcolsep}{6.5pt}
\scriptsize
\resizebox{\textwidth}{!}{%
\begin{tabular}{l cccccc ccccc}
\toprule
 & \multicolumn{6}{c}{\textbf{VBench-Long}} & \multicolumn{5}{c}{\textbf{ScriptAgent CriticAgent}} \\
\cmidrule(lr){2-7} \cmidrule(lr){8-12}
\textbf{Method} & SC $\uparrow$ & BC $\uparrow$ & MS $\uparrow$ & DD & AQ $\uparrow$ & IQ $\uparrow$ & CCA $\uparrow$ & KBL $\uparrow$ & VDF $\uparrow$ & EAM $\uparrow$ & NPT $\uparrow$ \\
\midrule
VideoClaw\textsuperscript{\ref{fn:app-videoclaw}} & 0.9082 & 0.9292 & 0.9904 & 0.4348 & 0.5579 & \textbf{0.7053} & 3.958 & 3.896 & 4.104 & \underline{4.042} & \textbf{4.271} \\
ViMax~\citep{huang2026vimax} & 0.9318 & 0.9484 & \underline{0.9925} & 0.2810 & 0.5895 & 0.5924 & \textbf{4.417} & \underline{4.208} & \textbf{4.542} & 3.625 & 3.938 \\
\midrule
GenMAC~\citep{genmac2026} & 0.9440 & \textbf{0.9598} & \textbf{0.9931} & 0.3510 & 0.5769 & 0.6066 & 3.250 & 4.062 & 3.104 & 3.083 & 3.500 \\
VideoWeaver~\citep{videoweaver2026} & \underline{0.9452} & \underline{0.9536} & 0.9924 & 0.4720 & \underline{0.5921} & 0.6475 & \underline{4.396} & 4.083 & 3.625 & \textbf{4.292} & 4.021 \\
\midrule
LIT (Ours)        & \textbf{0.9588} & 0.9518 & 0.9907 & 0.5074 & \textbf{0.5940} & \underline{0.6653} & 4.146 & \textbf{4.479} & \underline{4.396} & 3.875 & \underline{4.083} \\
\bottomrule
\end{tabular}
}
\end{table*}

\smallskip
\textbf{VBench-Long} provides objective, automatically computed scores from 0 to 1. Higher is better except for DD:
\begin{itemize}[leftmargin=1.5em, labelsep=0.3em, itemsep=2pt]
\setlength{\itemsep}{2pt}
\item \textbf{SC} (Subject Consistency) measures whether the appearance of subjects (people and objects) remains stable across frames without distortion.
\item \textbf{BC} (Background Consistency) measures whether the background scene stays stable across frames without noticeable jitter or abrupt changes.
\item \textbf{MS} (Motion Smoothness) measures whether actions and motion are coherent and natural, free of jumps or stuttering.
\item \textbf{DD} (Dynamic Degree) measures the amount of motion and change occurring in the video, reflecting the model's ability to ``make the scene move''. Higher DD is not necessarily better. DD should be interpreted together with the consistency and smoothness scores.
\item \textbf{AQ} (Aesthetic Quality) measures overall visual appeal in terms of composition, color, and lighting.
\item \textbf{IQ} (Imaging Quality) measures basic picture quality, including sharpness, distortion, and artifacts.
\end{itemize}

\textbf{ScriptAgent CriticAgent} reports subjective scores assigned by an AI critic (0--5, higher is better) along five dimensions:
\begin{itemize}[leftmargin=1.5em, labelsep=0.3em, itemsep=2pt]
\setlength{\itemsep}{2pt}
\item \textbf{CCA} (Cinematic Camera Articulation) measures the sophistication of camera work---shot types, framing transitions, and dynamic movements---in supporting the scripted narrative.
\item \textbf{KBL} (Kinetic Body Language \& Blocking) measures whether character motions, physical interactions, and spatial arrangements are specific, expressive, and consistent with the scripted actions.
\item \textbf{VDF} (Visual Descriptive Fidelity) measures how faithfully visual details (character appearance, clothing textures, scene layout, lighting) reproduce the descriptive cues in the script.
\item \textbf{EAM} (Emotional Arc \& Micro-Expressions) measures whether facial expressions, subtle gestures, and their temporal evolution reflect the emotional progression intended by the script and audio.
\item \textbf{NPT} (Narrative Pacing \& Timing) measures the alignment of shot timing, action beats, and pauses with the narrative structure and rhythm implied by the script and audio.
\end{itemize}

\subsection{Supplementary Qualitative Results}

\begin{figure}[t]
    \centering\includegraphics[width=\linewidth]{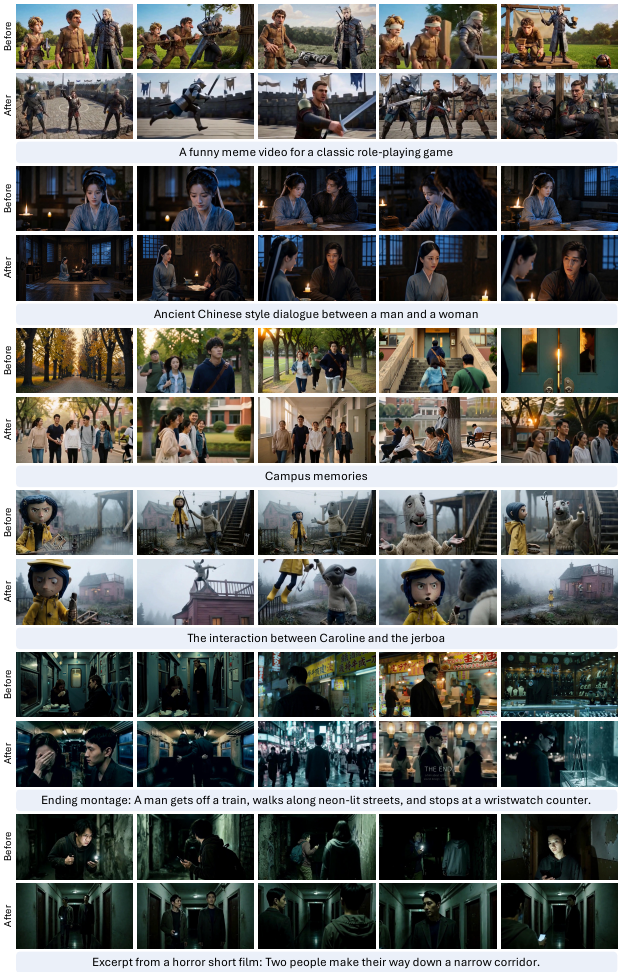}
    \caption{Character and scene consistency before and after evolution across six tasks.}
    \label{fig:fig9}
\end{figure}

\subsection{Case Study}

\begin{figure}[t]
    \centering
    \includegraphics[width=\linewidth]{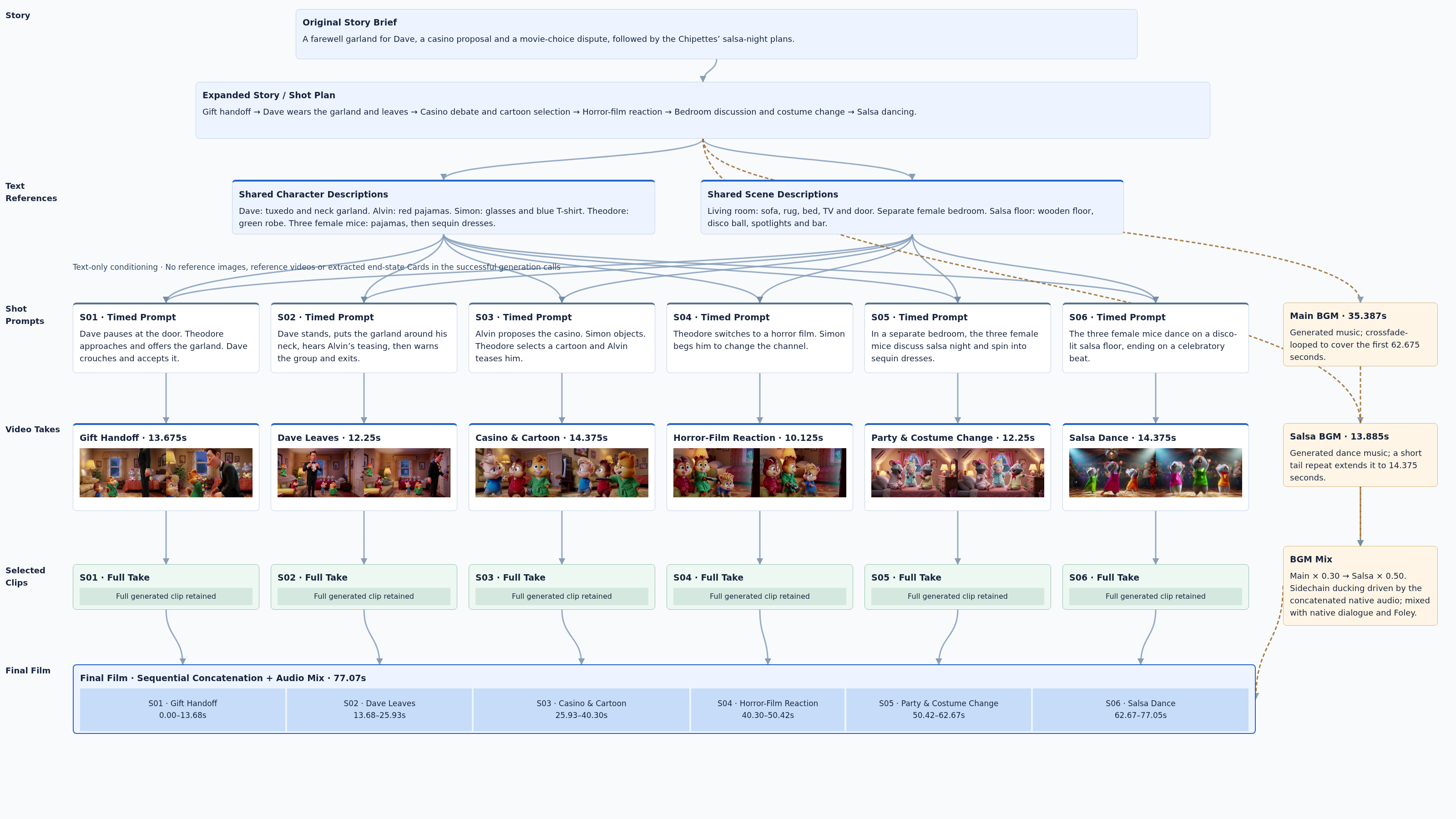}
    \caption{Production workflow before evolution. Shared text descriptions
    of characters and scenes guide timed shot prompts. Generated clips are
    retained in full and concatenated, then mixed with background music.
    The shown generation calls use no reference images or videos.}
    \label{fig:storyboard_old}
\end{figure}

\paragraph{Storyboard before and after evolution.} As shown in Fig.~\ref{fig:storyboard_old}, the initial pipeline first analyzes each shot, then directly uses the shot plan and the previous video as conditioning to generate the next video. Without a global anchor, this easily causes character and scene drift.

\begin{figure}[t]
    \centering
    \includegraphics[width=\linewidth]{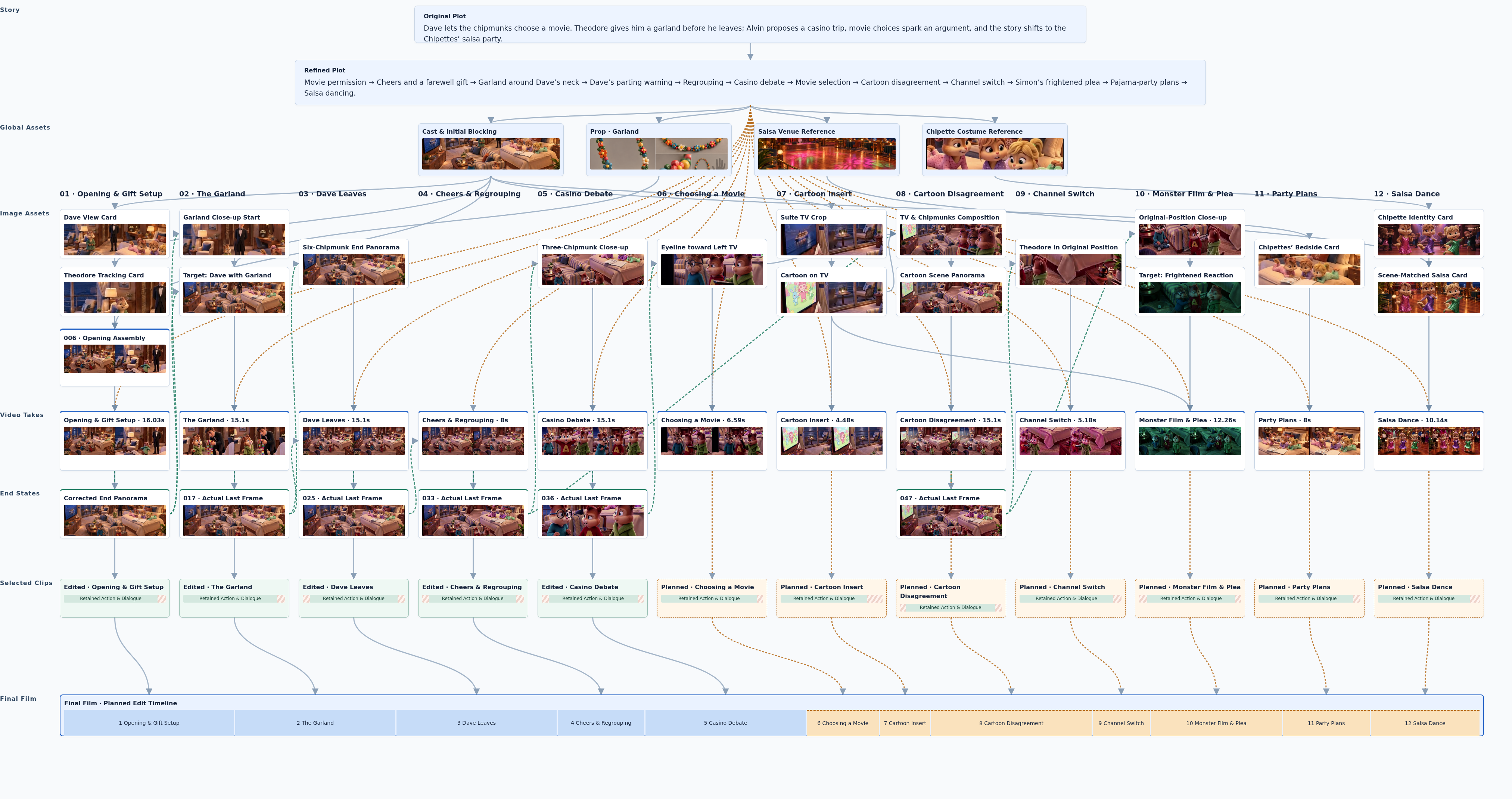}
    \caption{Production workflow after evolution. Shared character, scene,
    and prop references guide image preparation for each shot. Generated end
    frames provide visual references for later shots. The selection row
    distinguishes edited clips (green) from planned edits (orange), which
    feed into the final edit timeline.}
    \label{fig:storyboard_new}
\end{figure}

In Fig.~\ref{fig:storyboard_new}, the story is first expanded into performance beats. Character and scene references are used to set initial blocking and prepare shot-specific Cards. For continuous performance, pose, orientation, and prop states are extracted from the last generated frame as new references. For state changes, start and end images are used for guidance. When transitioning to the ball, identity and costume references are retained and the composition is rebuilt. Finally, footage is selected, trimmed, and spliced into a complete narrative, coordinating image generation, video generation, and editing through material dependencies.

\begin{figure}[t]
    \centering
    \includegraphics[width=\linewidth]{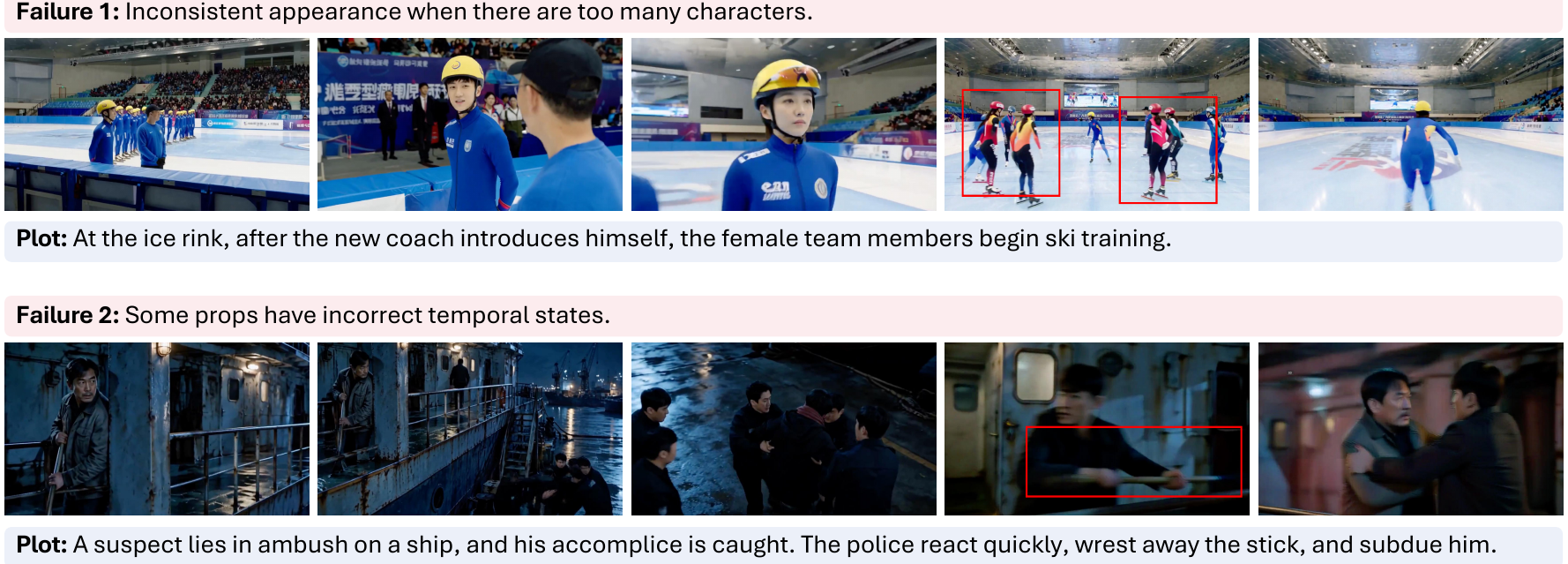}
    \caption{Representative failure cases. Top: character appearance changes
    in a scene with many subjects. Bottom: a stick's state is inconsistent
    with the action sequence. Red boxes highlight the errors.}
    \label{fig:failure_case}
\end{figure}

\paragraph{Failure case analysis.} We show representative failure cases. First, as shown in Fig.~\ref{fig:failure_case}, when the subjects in the frame are too complex, the video generation model may find it difficult to ensure consistency, and the VLM may find it difficult to detect errors. Second, in long videos, there are complex coupling relationships among the states of multiple subjects. This makes it difficult for the VLM to ensure the temporal consistency of the states of some props. Incorporating edge cases during the evolution process may be effective. Moreover, as foundation models advance, these problems will also be alleviated.

\subsection{Design of User Study}
\label{app:user-study}

\begin{figure}[t]
    \centering
    \includegraphics[width=\linewidth]{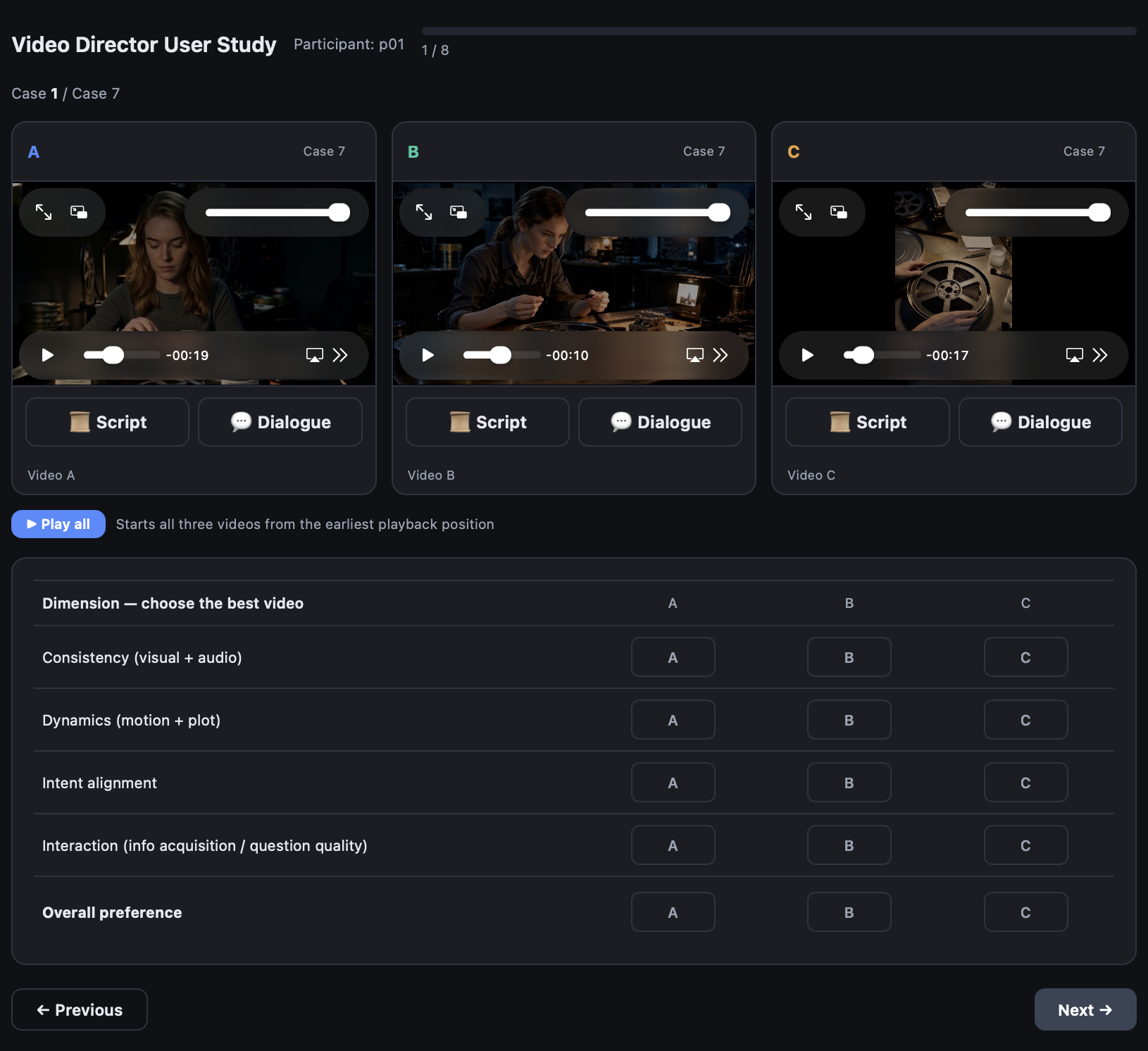}
    \caption{Visualization of user study website.}
    \label{fig:user-study-design}
\end{figure}

\paragraph{Materials and participants.}
The study contains 16 task groups and 48 videos in total.
Each group contains one video from ViMax, one from VideoWeaver, and one from LIT for the same task.
Fifteen participants each assess four groups, yielding 60 group assessments over this pool of videos.
Fig.~\ref{fig:user-study-design} illustrates the layout for comparing three videos within a group.

\paragraph{Evaluation dimensions and evidence.}
Participants rate the videos along Consistency, Dynamic, Interaction, and Intent, using the dimension definitions in Sec.~\ref{sec:metrics}.
They inspect the generated videos and audio to assess Consistency and Dynamic.
For Interaction and Intent, they also read the Agent dialogue records and the task's Oracle Script.
The Oracle Script provides the reference requirements, while the dialogue records show what information the Director requested and received.
These materials allow participants to assess both the interaction process and alignment with the intended task.

\paragraph{Reported comparisons.}
Tab.~\ref{tab:user-preference} reports human win counts aggregated over 60 group assessments and automatic evaluator win counts over the 16 task groups.
We report Kendall's $\tau$ between the human and automatic method rankings for each dimension.

